%% file: main.tex
\documentclass[10pt,conference]{IEEEtran}
\IEEEoverridecommandlockouts
\usepackage{cite}
\usepackage{amsmath,amssymb,amsfonts}
\usepackage{algorithmic}
\usepackage{graphicx}
\usepackage{textcomp}
\usepackage{xcolor}
\usepackage{xspace}
\usepackage{subfigure}
\usepackage{multirow}
\usepackage{multicol}
\usepackage{makecell}
\usepackage{stfloats}
\usepackage{flushend}
\usepackage{caption}
\usepackage{url}
\usepackage[skip=2pt]{caption}
\def\BibTeX{{\rm B\kern-.05em{\sc i\kern-.025em b}\kern-.08em
    T\kern-.1667em\lower.7ex\hbox{E}\kern-.125emX}}

\definecolor{third_s}{HTML}{008080}
\newcommand\fname{E-Navi}

\IEEEoverridecommandlockouts

\begin{document}




\title{\Huge \fname{}: Environmental-Adaptive Navigation\\for UAVs on Resource-Constrained Platforms}

\author{
\IEEEauthorblockN{Boyang Li\IEEEauthorrefmark{2}\IEEEauthorrefmark{3}, Zhongpeng Jin\IEEEauthorrefmark{2}, Shuai Zhao\IEEEauthorrefmark{2}, Jiahui Liao\IEEEauthorrefmark{2}, Tian Liu\IEEEauthorrefmark{2}, Han Liu\IEEEauthorrefmark{2}, Yuanhai Zhang\IEEEauthorrefmark{2}, Kai Huang\IEEEauthorrefmark{2}\IEEEauthorrefmark{3}
}
\IEEEauthorblockA{
\IEEEauthorrefmark{2}Sun Yat-sen University
\IEEEauthorrefmark{3}Key Laboratory of Machine Intelligence and Advanced Computing, Ministry of Education
}
}

\maketitle
\setcounter{equation}{0}
\begin{abstract}
\input{tex/abstract}

\end{abstract}

\setcounter{figure}{0}
\section{Introduction}
    \input{tex/introduction}
\section{Related Work and Motivation}
    \subsection{Related Work}
    \input{tex/relatedwork}
    \subsection{Motivation Example}
    \input{tex/motivation}
\vspace{-0.3em}
\section{Overall Framework}
    This section presents the framework of E-Navi, including environmental evaluation and dynamic task configuration to improve system performance. 
    \vspace{-0.3em}
    \subsection{System Model}
    \input{tex/systemmodel}
    \subsection{Module Interaction and Workflow}

\input{tex/framework}

\section{Environment Analysis via ECI Modeling}
    \input{tex/eci}
\section{Adaptive Configuration for Navigation System}
    \input{tex/computation}
\section{Evaluation}
    \input{tex/experiment}
\section{Conclusion}
    \input{tex/conclusion}

\bibliographystyle{IEEEtran}
\bibliography{ref}
\end{document}

%% file: tex/abstract.tex
The ability to adapt to changing environments is crucial for the autonomous navigation systems of Unmanned Aerial Vehicles (UAVs). However, existing navigation systems adopt fixed execution configurations without considering environmental dynamics based on available computing resources, e.g., with a high execution frequency and task workload. This static approach causes rigid flight strategies and excessive computations, ultimately degrading flight performance or even leading to failures in UAVs. Despite the necessity for an adaptive system, dynamically adjusting workloads remains challenging, due to difficulties in quantifying environmental complexity and modeling the relationship between environment and system configuration.
Aiming at adapting to dynamic environments, this paper proposes E-Navi, an environmental-adaptive navigation system for UAVs that dynamically adjusts task executions on the CPUs in response to environmental changes based on available computational resources. Specifically, the perception-planning pipeline of UAVs navigation system is redesigned through dynamic adaptation of mapping resolution and execution frequency, driven by the quantitative environmental complexity evaluations. In addition, E-Navi supports flexible deployment across hardware platforms with varying levels of computing capability. Extensive Hardware-In-the-Loop and real-world experiments demonstrate that the proposed system significantly outperforms the baseline method across various hardware platforms, achieving up to 53.9\% navigation task workload reduction, up to 63.8\% flight time savings, and delivering more stable velocity control.

%% file: tex/introduction.tex
\begin{figure}[t]
    \centering
    \includegraphics[width=\linewidth]{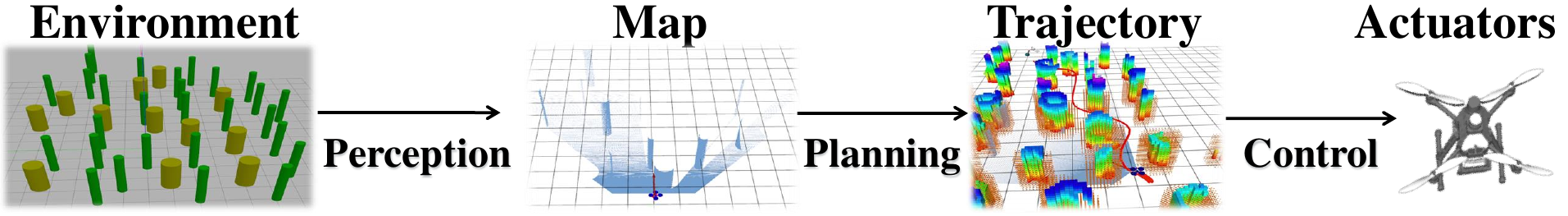}
    \caption{A typical task pipeline of autonomous UAV.}
    \label{intro-mavpipeline}
    \vspace{-10pt}
\end{figure}

Autonomous Micro Aerial Vehicles (UAVs) have been widely applied across various domains, such as transportation, industry, agriculture, and rescue \cite{8951121,alsamhi2022uav,hazmy2023potential}. 
In these application domains, UAVs often face dynamic environments characterized by complex structures and diverse obstacle distributions. Meanwhile, the computing capabilities are limited by resource-constrained onboard hardware platforms. Consequently, the ability to adapt to environmental changes under limited computational resources is crucial for UAV navigation systems, which are typically organized into task pipelines\cite{hsiao2023silent} consisting of perception, planning, and control stages, as shown in Fig. \ref{intro-mavpipeline}. Since the outputs of the perception and planning stages directly influence the control stage, whose computational load remains relatively stable\cite{burri2016euroc}, maintaining adaptive configurations in perception-planning is essential to ensure efficient flight.

Existing UAV navigation systems\cite{zhou2020ego,wang2022geometrically,chen2020computationally} typically adopt fixed execution configurations for perception and planning tasks. These systems adopt a best-effort strategy that applies high-frequency and high-load workloads. Specifically, a partitioned scheme is applied to schedule tasks to cores with static periods, assuming stable workloads. However, in practice, the computational demand of perception and planning tasks changes with scene complexity, primarily due to variations in the amount of data to be processed. As a result, a simple partitioned schedule with fixed task configurations cannot accommodate such workload fluctuations~\cite{9520259,9355507}.
Such systems lead to two major problems: Firstly, a rigid flight strategy that fails to adapt in changing environments, degrading navigation efficiency. Secondly, redundant high-frequency and high-resolution computations, which introduce excessive planning and frequent trajectory changes, ultimately harming flight performance. These problems are especially prominent in highly dynamic scenarios on resource-constrained hardware, resulting in frequent abrupt accelerations, discontinuous velocities, detours, or even system failures. On the one hand, applying fixed coarse-grained perception and planning in sparse and dense environments results in detours due to inadequate obstacle awareness \cite{boroujerdian2018mavbench}. On the other hand, sustaining unnecessarily high-precision computations in simple environments can lead to over-planning and unstable maneuvers, which further degrade overall system performance.

\textbf{Challenges.} Several challenges arise when dynamically adjusting perception-planning configurations in changing situations. Firstly, quantifying obstacle distributions as an environment measurement is inherently difficult, as it requires identifying characteristics that capture the impacts on UAV decision making and motion behavior\cite{hadidi2021quantifying}. Secondly, establishing the interrelationship between environments and task configurations demands a holistic model that considers various factors such as spatial distribution, directionality, and constraints~\cite{hadidi2023context}. Further, most works overlook the collective effects of dynamic environments and navigation task workload on flight performance \cite{quan2021eva,hsiao2022zhuyi,wang2022neither}. Therefore, how to appropriately adjust the workload of the UAVs navigation tasks under dynamic conditions remains an open question~\cite{liu2021pi}.

\textbf{Contributions.} This paper presents an environment-adaptive navigation system (namely E-Navi) designed for UAVs on resource-constrained platforms. Specifically, the constructed E-Navi system is adaptive to varying situations by modeling the interrelationships between obstacle distributions in the environments, navigation task workload, and available computing resources. The Environment-Adaptive Spatial Complexity Index (ECI) is proposed to enable the system to dynamically adjust perception and planning configurations in response to environmental changes. We adopt the global EDF (Earliest Deadline First) as the OS-level task scheduler, with task periods dynamically assigned by E-Navi and relative deadlines set equal to their periods. In addition, E-Navi supports flexible deployment across hardware platforms with varying computing capabilities. To achieve this, the following contributions are made in this paper.
\begin{itemize}
\item An obstacle distribution modeling is established through directional sector analysis to quantify environmental complexity for autonomous UAV navigation.
\item The interrelationships are formulated between environments and navigation workload.
\item A configuration adjustment framework is proposed to improve system performance by dynamically tuning the perception planning pipeline.
\item Extensive Hardware-In-the-Loop (HIL) and real-world experiments indicate that E-Navi outperforms the baseline method across a range of devices.
\end{itemize}


%% file: tex/relatedwork.tex
Considerable progress has been made in autonomous UAV navigation. However, existing systems designed with rigid and statically defined execution configurations typically disregard the dynamics of real-world environments and the navigation task workload. Given the challenges associated with dynamically adapting system configurations, we review prior work from two perspectives: methods for quantifying environmental complexity, and execution strategies for perception and planning pipelines in the navigation system.

\textbf{Environmental Complexity Evaluation.} Assessing obstacle distribution plays a crucial role in guiding adaptive execution of the perception-planning pipeline in navigation systems. A fine-grained understanding of spatial obstacle distributions enables the system to dynamically adjust mapping resolution and perception-planning execution frequency, thus improving responsiveness to environmental variation while maintaining computational efficiency. Existing methods, such as \cite{shao2024design,wu2025towards,zhang2023asap,huang2024safe}, typically quantify environmental complexity by evaluating obstacle density and spatial dispersion in a set of fixed-size spatial zones within the whole 3D space. These methods present notable limitations that restrict their adaptability in dynamic and resource-constrained scenarios.

\textbf{Limitation 1.} \textit{As observed in \cite{shao2024design,wu2025towards,zhang2023asap,huang2024safe}, existing methods neglect the directional distribution of obstacles relative to the UAV's heading and the urgency implied by relative distances within the field of view. Treating all obstacles as equally important regardless of direction fails to distinguish between those directly in the flight path and those in peripheral regions, reducing the accuracy of the assessment of environmental complexity. Moreover, ignoring obstacle proximity leads to unnecessary processing of distant or low-impact areas, resulting in redundant and inefficient computations.}

Beyond basic density-based metrics, several works investigate other types of environment representation to guide UAV navigation. For example, \cite{faessler2016autonomous} incorporates depth-aware segmentation, where the scene is divided into semantically meaningful regions based on depth discontinuities, to improve UAVs’ perception of navigable space. Similarly, \cite{loquercio2021learning} employs learning-based models to estimate how easily a UAV can traverse a region, called environmental navigability, which captures the difficulty of navigation based on learned priors. Although these methods offer richer environmental understanding, they typically rely on dense sensing updates and require substantial offline training datasets, which impose significant computational overhead and limit their applicability in complex scenarios\cite{boroujerdian2018mavbench,boroujerdian2021roborun}.

\textbf{Perception and Planning Strategies.} Some works explore adaptive perception and planning modules. For instance, \cite{hsiao2022zhuyi} proposes a perception-driven framework to adjust sensor update rates based on the complexity of the observed scenario, to balance perception accuracy and task execution efficiency. Ego-Planner \cite{zhou2020ego} introduces an environmental-adaptive planner that dynamically adjusts UAV flight paths under aerodynamic constraints (e.g., velocity and maneuverability). Shah et al. \cite{shah2023energy} focus on energy-efficient path planning by reducing redundant computations across pipeline stages, such as repeated map updates or re-planning under static conditions. At the system level, MavBench \cite{boroujerdian2018mavbench} and RoboRun \cite{boroujerdian2021roborun} evaluate how environmental changes affect system performance. In particular, RoboRun employs polynomial regression models with a constrained optimization solver to improve the performance. Although it provides an end-to-end navigation pipeline, it does not include OS-level scheduling adaptations. In addition, the implementation of RoboRun is not publicly available, which poses a barrier to direct experimental comparisons. MAVBench serves mainly as a benchmarking tool for UAV workloads. Ego-Planner emphasizes trajectory re-planning without adapting system-level parameters.

In conclusion, most existing methods focus on application adjustment in an isolated hardware device, without addressing the joint orchestration of the entire perception-planning pipeline under dynamic environments and various resource conditions \cite{xu2024automatic}, resulting in the following limitation. 

\textbf{Limitation 2.} \textit{Although some prior systems have explored dynamic adaptations, their scope is generally confined to the application layer. These mechanisms improve performance but remain limited to algorithmic tuning, without considering how tasks interact at the OS level, or how such interactions manifest across different hardware platforms. As a result, their effectiveness can diminish in complex environments, leading to execution delays or unstable performance.}

Although most UAV navigation methods adopt fixed execution configurations, several recent efforts have explored adaptive system behaviors under constrained computational budgets. For instance, the RED framework \cite{li2023red} introduces a real-time framework to improve the efficiency of multi-task deep neural networks (DNNs) in robotic systems with limited resources. Similarly, \cite{mcgowen2024scheduling} investigates the allocation of heterogeneous tasks in cyber-physical systems, showing that fine-grained execution control can significantly improve system responsiveness and throughput.

Unfortunately, these techniques focus on generic scheduling or neural network inference, rather than the UAV navigation, where perception and planning are tightly coupled with environmental dynamics. More importantly, they do not add environmental semantics, such as obstacle distribution into the execution logic, which is essential for UAV in dynamic spaces.

In summary, previous work highlights the value of adapting system behavior in complex environments. However, few efforts have considered the joint impact of environmental complexity and the task workload on the design of UAV navigation systems. Therefore, we propose an environmental-adaptive navigation system that dynamically configures both perception and planning pipelines by explicitly modeling spatial environment features and system states. The proposed system enables UAVs to maintain efficient behavior in a variety of resource-constrained and dynamically evolving scenarios.

%% file: tex/motivation.tex
\vspace{-1em}
\begin{figure}[!htbp]
    \centering{
        \subfigure[UAV velocities in simple scenario with sparse obstacle distribution.]{
            \includegraphics[width=\linewidth]{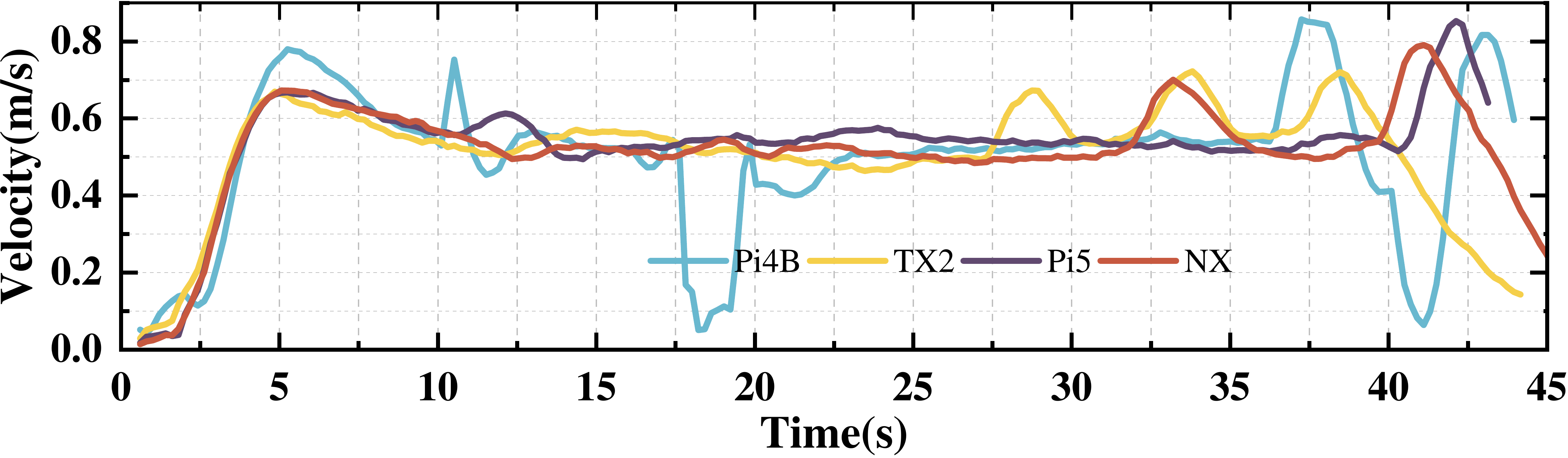}
            \label{motiva-1-simple}
        }
        \subfigure[UAV velocities in complex scenario with dense and irregular obstacles.]{
            \includegraphics[width=\linewidth]{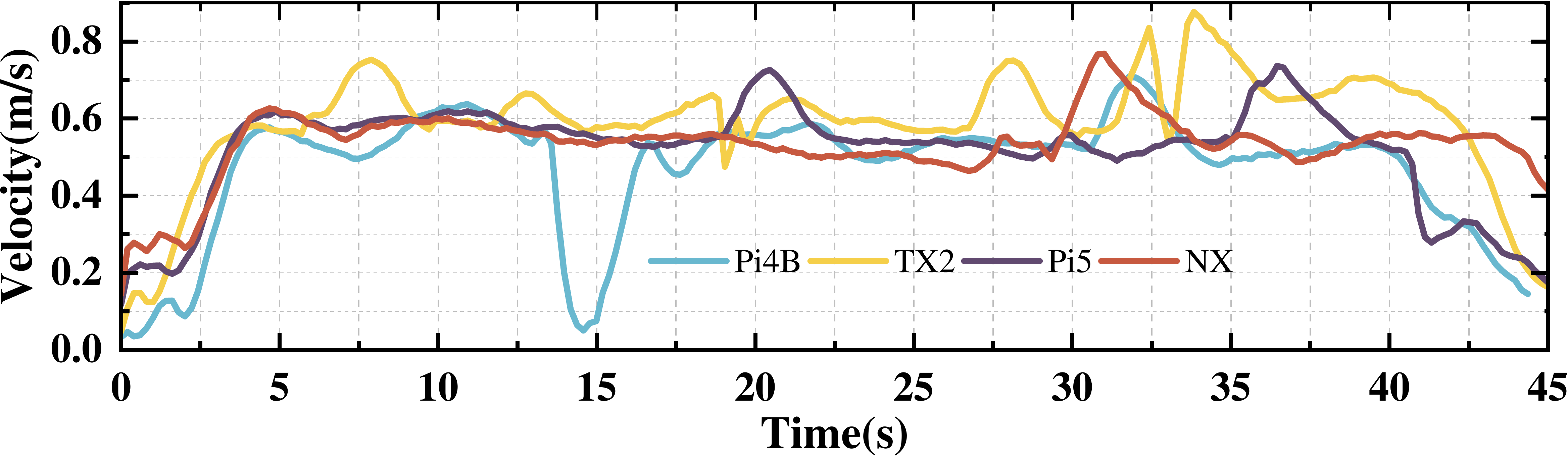}
            \label{motivate-1-complex}
        }
    }
    \caption{Velocities of UAV with fixed configurations.}
    \label{motiva-1}
\end{figure}

\vspace{-.7em}

Unlike existing UAV navigation systems that adopt fixed configurations, we argue that both the execution frequency and the task workload should be dynamically adapted to environmental conditions. This adaptability brings two key benefits: (i) it improves the performance of UAVs by enabling smoother navigation in dynamically changing environments, and (ii) it improves navigation efficiency and system performance.

\begin{figure*}[!hbp]
    \centering{
        \subfigure[Results with resolutions.]{
            \includegraphics[width=.20\linewidth]{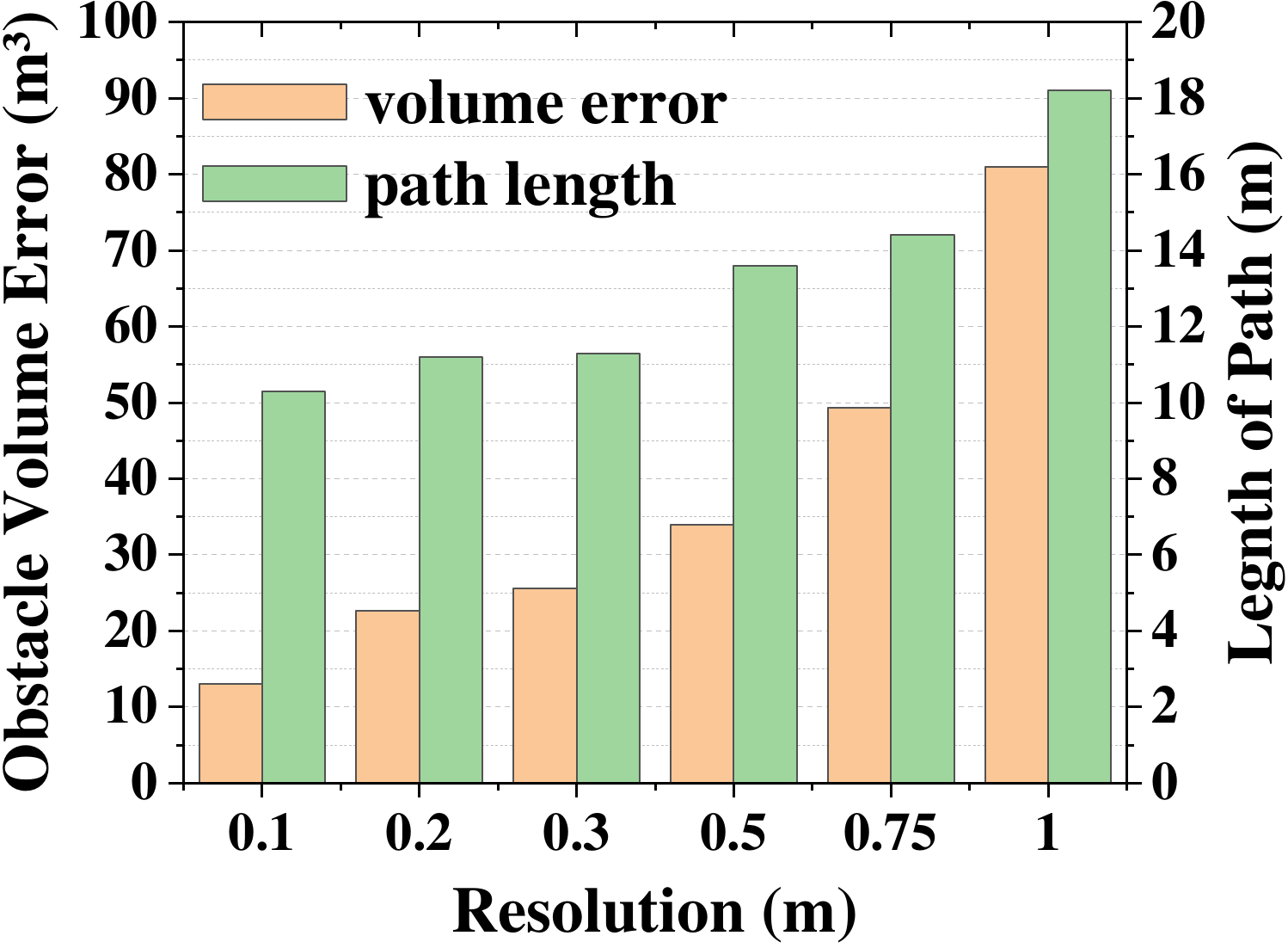}
            \label{res-volume-error}
        }
        \subfigure[Scenario]{
            \includegraphics[width=.16\linewidth]{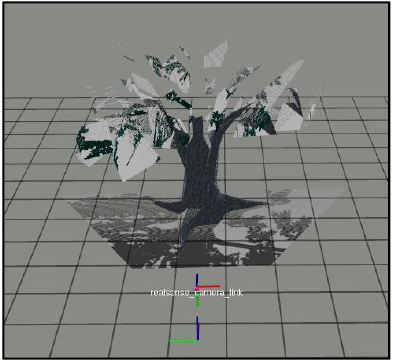}
            \label{res-sc}
        }
        \subfigure[Resolution=10cm.]{
            \includegraphics[width=.16\linewidth]{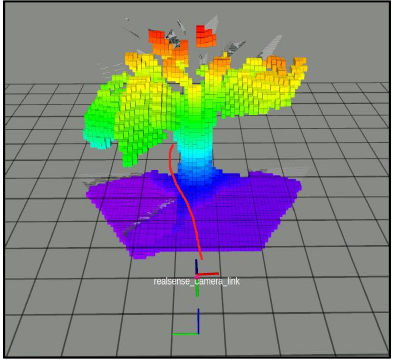}
            \label{res-10}
        }
        \subfigure[Resolution=50cm.]{
            \includegraphics[width=.16\linewidth]{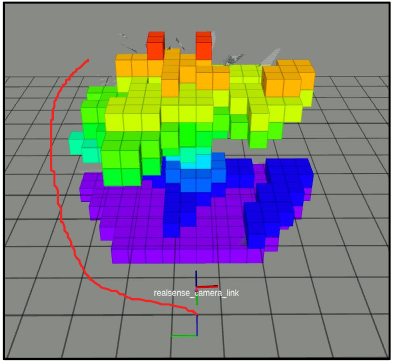}
            \label{res-50}
        }
        \subfigure[Resolution=100cm.]{
            \includegraphics[width=.16\linewidth]{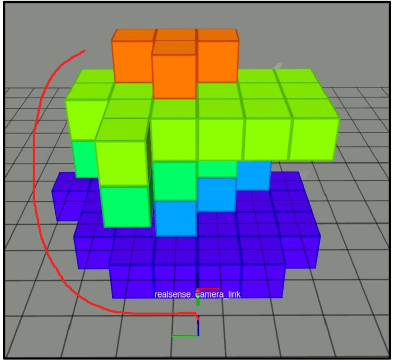}
            \label{res-100}
        }
    }
    \setlength{\abovecaptionskip}{-2pt}
    \setlength{\belowcaptionskip}{-5pt}
    \caption{The resolution of the map, the spatial error in estimating the volume of obstacles, and the length of the flight path.}
    \label{res-discussion}
\end{figure*}

\textbf{Motivation Example 1.} The ability of a UAV to perform stable and responsive flight maneuvers depends on the tight and efficient integration between perception and planning in autonomous navigation. In current systems, perception and planning modules are typically configured with fixed parameters, such as high-frequency updates and fine-grained maps, regardless of changes in environmental complexity. This configuration lacks adaptability, causing delayed reactions or redundant computation in dynamic environments, which in turn degrades flight stability with abrupt accelerations, decelerations, or intermittent halts.

To illustrate this, Fig. \ref{motiva-1} shows the velocities of UAV equipped with different platforms during navigation in both simple (sparse obstacle distribution) and complex (dense and irregular obstacle layout) environments. Across all configurations, we observe that traditional designs assume static workloads, using fixed task parameters, and thus fail to adapt to the varying environments. In the sparse scenario, obstacles are concentrated near the target region. The Raspberry Pi, due to its limited compute capability, responds less effectively to this complexity increase in the final flight stage, resulting in velocity fluctuations. In contrast, the dense environment triggers conservative behavior early on, supporting the motivation: dynamic reconfiguration is essential in varying conditions.

\textit{These observations highlight the limitations of static configurations, which ignore environmental variations, resulting in delayed responses and unstable velocities.}

\textbf{Motivation Example 2.} In dynamic environments, UAVs continuously update their flight trajectories through the perception-planning pipeline. For energy-constrained platforms, generating short and efficient trajectories is particularly important to the navigation system. However, when fixed perception and planning configurations are used, UAVs often suffer from incomplete environmental awareness or delayed responses to obstacle changes, leading to long detours.

Fig. \ref{res-discussion} illustrates the relationship between the resolution of the map, the spatial error in estimating the volume of obstacles, and the length of the flight path. As resolution decreases, the estimation error increases significantly from approximately $10\text{m}^3$ at 10 cm resolution to over $90\text{m}^3$ at 100 cm. Although high-resolution maps require more computation resources, they allow the navigation system to more accurately perceive the environment, resulting in more informed trajectory planning. In contrast, coarse-resolution maps reduce the computational load but tend to underestimate obstacle volumes, increasing the risk of trajectories that pass dangerously through the obstacles or require abrupt emergency stops.

\textit{This example shows that fixed configurations (either overly fine or coarse grained) can cause  redundant computation or unsafe trajectories, demonstrating the need for adaptive strategies with environmental complexity.}

In conclusion, the two examples show that static configurations are inadequate for dynamic environments, leaving UAVs prone to degraded performance manifested by unstable control and inefficient trajectories in complex scenarios.

\begin{figure*}[!ht]
    \centering
    \setlength{\belowcaptionskip}{-10pt}
    \includegraphics[width=.85\linewidth]{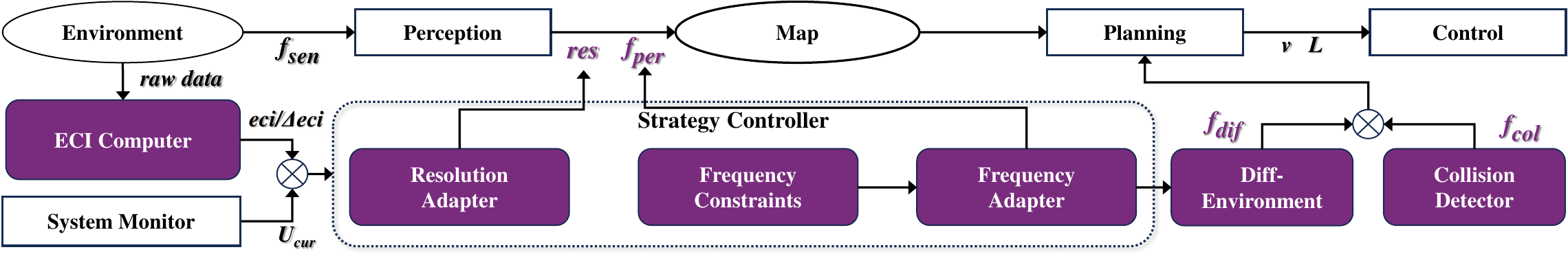}
    \caption{The framework of the proposed E-Navi.}
    \label{framework-framework}
    \vspace{-3pt}
\end{figure*}

%% file: tex/systemmodel.tex
The UAV onboard system is equipped with a collective of $N$ identical CPU processors, denoted $\mathcal{U}_{total}$. These processors execute a set of navigation tasks $\Gamma$, including perception and planning tasks.  Consequently, the total system utilization is $\mathcal{U}_{total} = N \times 100\%$. The computational cost associated with each task $\tau_i \in \Gamma$ is defined by a tuple $\tau_i = \{\mathcal{U}_{i}, \mathcal{C}_{i}\}$, where $\mathcal{U}_{i}$ denotes the required CPU utilization rate, and $\mathcal{C}_{i}$ represents the worst-case execution time for one execution.

Our proposed navigation system is modeled as follows. The sensor continuously outputs perceived environmental data at a frequency of $f_{sen}$. The perception task $\tau_{per}$ is executed to build and update the environmental $map$ with a resolution of $res$ at a frequency of $f_{per}$. The planning task $\tau_{plan}$ computes the trajectory $\mathcal{L}$ and velocity $v$ using the updated $map$, with collision and diff-environmental detections, executing at frequencies $f_{col}$ and $f_{dif}$, respectively, serving as triggers for the planning process. $\mathcal{C}_{\mathcal{L}}$ represents the flight time along $\mathcal{L}$. Upon completion of $\tau_{plan}$, the current $\mathcal{L}$ is terminated and transitions to the newly computed trajectory.

It is worth noting that the proposed E-Navi system does not directly deal with task scheduling problems. Rather, the global Earliest Deadline First (EDF) scheme \cite{lopez2004utilization} is utilized to schedule task instances in the navigation system based on the dynamic configurations generated by E-Navi. In such a setup, the proposed E-Navi can provide timing guarantee by enforcing a upper bound on the utilization of the tasks when configuring the task frequency (see details in Sec. V.D)\cite{liu1973scheduling}.

%% file: tex/framework.tex
As illustrated in Fig. \ref{framework-framework}, E-Navi consists of two modules: \textit{ECI Computer} for analyzing environmental complexity (Sec. IV) and the \textit{Strategy Controller} for adaptive task configuration (Sec. V). The raw sensor data at frequency $f_{sen}$ is processed to compute two indicators: the Environment-Adaptive Spatial Complexity Index $eci$ and its temporal variation $\Delta eci$. Meanwhile, the \textit{System Monitor} tracks CPU utilization $\mathcal{U}_{cur}$. These outputs jointly inform the \textit{Strategy Controller}, which adjusts the execution configurations based on environmental dynamics and resource availability. Specifically, \textit{Strategy Controller} updates the map resolution $res$ in \textit{Resolution Adapter} (Sec. V.A), and modifies the frequency of tasks, including mapping $f_{per}$, collision detection $f_{col}$, and environmental change detection $f_{dif}$ in \textit{Frequency Adapter} (Sec. V.C), under constraints in \textit{Frequency Constraints} (Sec. V.B).

Within the Strategy Controller, the \textit{Resolution Adapter} firstly calculates the appropriate map resolution $res$ and evaluates the feasibility of the current planned trajectory. Based on this, the \textit{Frequency Constraints} module formulates a feasible frequency space. These outputs are then sent into the \textit{Frequency Adapter}, which determines the execution frequencies: $f_{per}$ for perception updates, $f_{col}$ for obstacle re-evaluation, and $f_{dif}$ for detecting significant environmental transitions.

Unlike fixed-configuration systems, the proposed E-Navi dynamically adapts perception and planning modules. Both resolution and map update frequency are adjusted online, while planning tasks are triggered by significant environmental changes. After each planning task, the resulting trajectory $\mathcal{L}$ is followed until a new trigger occurs, forming an adaptive control loop that ensures efficient navigation under varying environmental and computational conditions.


%% file: tex/eci.tex
In UAV navigation systems, the ability to perceive and interpret environmental complexity is essential not only to avoid obstacles but also to enable adaptive task management under resource constraints. To support this, we propose a layered Environment-Adaptive Spatial Complexity Index model to quantify environmental complexity, referred to as $eci$. Unlike dense data, $eci$ provides a structured and computationally efficient abstraction of obstacle distributions, enabling a rapid assessment of navigation difficulty. Using $eci$ to evaluate the environments, the model can selectively adjust configurations, ensuring efficient navigation on constrained platforms.

To capture environmental complexity, $eci$ evaluates obstacles induced risks along two spatial dimensions: direction and depth upon each frame using angular and depth information from a depth camera or LiDAR sensor. Obstacles at the UAV’s forward direction pose the most immediate threats and warrant higher considerations in risk assessment, while lateral and vertical obstacles offer more maneuvering flexibility. Similarly, near-field obstacles require prompt evasive actions, whereas those in the far field allow for more strategic planning.

\begin{figure}[!htbp]
    \centering
    \includegraphics[width=.8\linewidth]{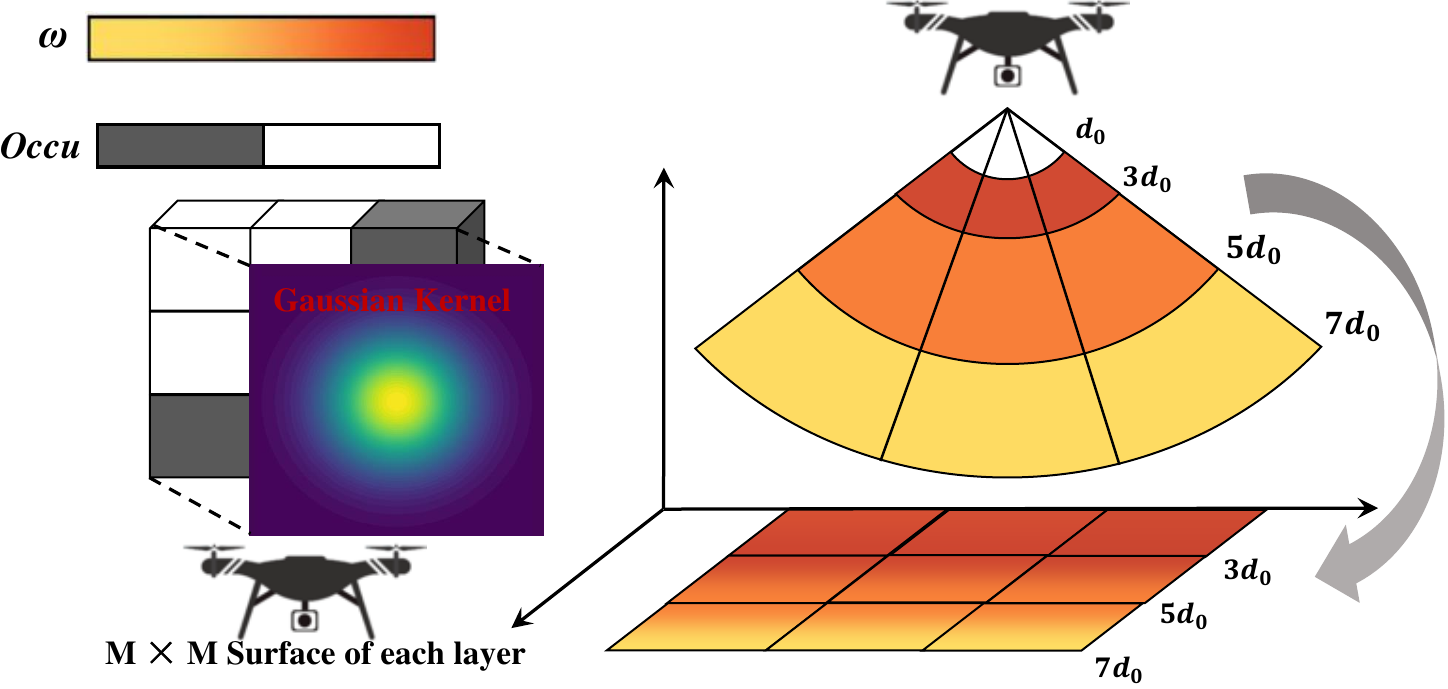}
    \setlength{\belowcaptionskip}{-5pt}
    \caption{$eci$ evaluation for UAVs.}
    \label{eci-demo}
    \vspace{-0.6em}
\end{figure}

To reflect this anisotropic structure, the surrounding space of the UAV is hierarchically partitioned along both the depth axis and the angular field of view, forming a structured set of directional zones $Z_{3d}$ in the 3D space displayed in Fig. \ref{eci-demo}. $Z_{3d}$ is composed of $Lay$ layers of 2D angular zones $Z_{2d}$. Each 3D zone is associated with an occupancy state $Occ$ and a weight $\omega$ that dynamically reflects its importance. These weighted observations serve as the basis for computing $eci$ and enable a spatially complexity evaluation. The key parameters involved in $eci$ computation are summarized in Tab. \ref{eci-para}.
\begin{table}[!ht]
    \centering
    \vspace{-2pt}
    \caption{Key Parameters for computing $eci$.}
    \label{eci-para}
    \begin{tabular}{c|c|c}
      \hline
      Parameter  & Descriptions & Type\\
      \hline
      $d_{per}$ & Maximum perception distance & Given\\
      \hline
      $v_{cur}$ & Current velocity & Measurable\\
      \hline
      $d_{0}$ & Ranging distance of the nearest obstacle & Measurable\\
      \hline
      $Z_{3d}$ & The generated 3D spatial zones& Computed\\
      \hline
      $Z_{2d}$ & The 2D angular zones in each layer& Computed\\
      \hline
      $Lay$ & Number of depth layers & Computed\\
      \hline
      $\omega\in(0,1)$ & Weight of each zone & Computed\\
      \hline
      $Occ=\{0,1\}$& \makecell{Occupancy state of each zone, \\1, 0 means occupied and unoccupied} & Measurable\\
      \hline
    \end{tabular}
    \vspace{-1em}
\end{table}

\subsection{Spatial Area Division}
\textbf{Division of Layer in Depth Direction.} The depth axis is discretized using the strategy that reflects the urgency of obstacle avoidance. Given the current velocity $v_{cur}$ and the perceived ranging distance to the nearest obstacle $d_0$, the first layer covers the interval $[0, d_0)$, representing the minimum stopping distance. Obstacles within this region demand immediate evasive action upon detection~\cite{liu2021real}. The corresponding reaction time is computed as:
\begin{equation}
    t_0 = \frac{2d_0}{v_{cur}}
    \label{eci-d0}
\end{equation}

Beyond the first layer, all subsequent layers have a fixed thickness of $2d_0$, forming intervals such as $[d_0, 3d_0)$, $[3d_0, 5d_0)$, etc. This structure aligns spatial segmentation with a constant-time traversal model, assuming uniform velocity $v_{cur}$ in the fixed time interval $t_0$, enabling a consistent temporal interpretation of obstacle proximity. Hence, this time-aware partitioning allows for fine granularity in the near field, where rapid response is essential, while reducing resolution in farther regions to maintain computational efficiency. The total number of layers is adaptively bounded by the maximum perception range of the sensor $d_{per}$, computed as:
\begin{equation}
    Lay = \left\lfloor \frac{d_{per} - d_0}{2d_0} \right\rfloor + 1
    \label{eci-d-per}
\end{equation}
Eq. \ref{eci-d-per} ensures that the layered spatial model remains consistent with the sensor's effective observation capabilities, preventing over-segmentation of undetectable space.

\textbf{Division of Angular Zones within Each Layer.} Within each depth layer, the perceptual range of the UAV is projected onto a spherical surface and partitioned into grids $M \times M$ of angular zones $Z_{2d}$, defined by the corresponding azimuth $\theta$ and pitch $\beta$ of each zone. This angular grid is aligned with the UAV's forward direction, ensuring consistent spatial interpretation across depth layers. The central zone, oriented along the UAV's velocity vector, represents the most threats region for potential collisions, while surrounding zones account for the threats from lateral and vertical directions. This direction effectively models the directional anisotropic of risks during flight, emphasizing the importance of obstacles directly ahead while allowing efficient representation of peripheral threats. 



\subsection{Weights Computation}
Each spatial zone is assigned a directional weight $\omega_{ak}$, which reflects its relative contribution to the evaluation of environments in our framework. This weight integrates two key factors: (1) the angular deviation of the zone from the UAV's forward direction, and (2) the depth distance of the obstacle relative to the UAV. Intuitively, zones located closer to the center of the field of view and at shorter distances are considered more important for timely avoidance.

Specifically, let $\theta_a$ and $\beta_a$ denote the azimuth and pitch angles corresponding to the angular zone $a \in Z_{2d}$ within layer $k$, and let $d_k$ denote the average depth value associated with this layer. A truncated Gaussian kernel is used to encode the angular importance centered at the UAV's heading, while a linear decay term models the urgency of obstacles at varying depths. The final weight for zone $Z_{3d}(a, k)$ is defined as:
\begin{equation}
    \omega_{ak} = \frac{d_{per} - d_k}{d_{per}} \cdot \frac{1}{\sigma \sqrt{2\pi}} 
    \cdot \exp\left( -\frac{(\theta_a - \mu)^2 + (\beta_a - \mu)^2}{4\sigma^2} \right),
    \label{framework-weights}
\end{equation}

where $\mu$ represents the direction of the UAV heading, $\sigma$ controls the angular sensitivity, and $d_{per}$ is the maximum perception range of the sensor. This formulation ensures that both temporal urgency and spatial orientation are jointly incorporated into the evaluation of environmental risk.

\subsection{$eci$ Computation}
Each spatial zone $Z_{3d}(a,k) \in Z_{3d}$ is associated with the corresponding occupancy status $Occ_{ak}$ and the weights $\omega_{ak}$, which reflects the local obstacle density and the influence within that segment. The overall $eci$ is calculated as the weighted sum of all zones in Eq. \ref{framework-eci}:
\begin{equation}
    eci = \sum_{k=1}^{Lay}\sum_{a=1}^{Z_{2d}}\omega_{ak} \cdot Occ_{ak}
    \label{framework-eci}
\end{equation}
The $eci$ is periodically computed within the framework as part of environment monitoring. Since $eci$ directly reflects the perceived complexity of obstacles in the surrounding environment, it serves as an interpretable measure of situational difficulty. When system is overloaded, a conservative mode is entered by keeping the UAV in a hovering state.

Notably, the proposed $eci$ computation remains computationally lightweight and suitable for deployment onboard. Spatial segmentation follows a fixed-size $M\times M\times Lay$ zones, which avoids dynamic map constructions or high-dimensional inference. All zone occupancies are obtained directly from raw depth-map generated by the sensor, while the weight assignments are derived via closed-form expressions using Gaussian kernels and linear decay, allowing for efficient computation. Moreover, $eci$ serves as an effective plug-in complexity quantifier that can be directly integrated into varying systems. $\sigma=1$ and $M=3$ are set as default parameters, which are empirically selected and kept fixed across all scenarios.

\textbf{Characteristic 1.} \textit{The ECI computer incorporates both obstacle proximity and directional weighting into a unified spatial complexity score. By emphasizing obstacles near the UAV's forward path and peripheral or distant elements, it captures the relative urgency of environmental features. 
This modeling directly addresses Limitation 1 by introducing directional and proximity-aware differentiation, thereby enhancing evaluation efficiency and reducing unnecessary computations.}

%% file: tex/computation.tex
Building upon the spatial complexity quantified by the ECI model, this section details the design of the \textit{Strategy Controller} for adaptively reconfiguring tasks in the navigation system. In response to dynamic environmental conditions, the \textit{Strategy Controller} integrates several submodules (i.e., \textit{Resolution Adapter}, \textit{Frequency Constraints}, and \textit{Frequency Adapter}) to dynamically reconfigure key navigation tasks.

\subsection{Resolution Adapter}

The \textit{Resolution Adapter} optimizes computational resources of the system by using a greedy algorithm to select the minimum resolution 
$res$ 
for feasible trajectory planning.
The algorithm progressively refines $res$ to ensure that the Manhattan distance between the start and end points is non-zero, guaranteeing enough discretization for path planning. In scenarios such as collision detection or frequent map updates, the system escalates to a higher resolution for more accurate path adjustments, ensuring the trajectory is both computationally feasible and responsive to unexpected environmental changes.

While the greedy-based algorithm ensures a feasible resolution for generating a valid path, it does not guarantee optimality in terms of length or smoothness. To assess trajectory quality, the system compares the actual path $\mathcal{L}$ with an estimated optimal reference $\mathcal{L^*}$, allowing the detection of suboptimal routes and triggering further refinement if necessary. The reference $\mathcal{L^*}$ is defined as follows:
\begin{equation}
    \label{design-optimal_path}
    \mathcal{L^*}=(1+\gamma\frac{\rho_c}{\rho^*_c})\mathcal{\bar{L}} 
\end{equation}

where $\rho_c$ denotes the number of occupied grids (obstacles), $\rho^*_c$ represents the total number of grids within the map, $\gamma \in \mathbb{R}^+$ is a scaling hyperparameter, and $\mathcal{\bar{L}}$ is the ideal straight-line path between the start and end points. Thus, $\mathcal{L}^*$ increases with obstacle density and approaches $\mathcal{\bar{L}}$ in obstacle-free environments. Consequently, the score $s \in (0-1]$ for the planned trajectory $\mathcal{L}$ is defined in Eq. \ref{design-s} to measure the efficiency of the trajectory, where a higher score indicates a more optimal path with less deviation from the ideal route.
\begin{equation}
    \label{design-s}
    s=e^{-\frac{(||\mathcal{L}||-||\mathcal{L^*}||)}{||\mathcal{\bar{L}}||}}, ||\mathcal{L}||\geq ||\mathcal{L^*}||
\end{equation}

\subsection{Frequency Constraints}
The \textit{Frequency Constraints} define conditions for task configuration and timing verification, considering limited computational resources and environmental conditions. 

\textbf{Constraint 1.} Perception Update Constraint. To avoid blind flight, the perception task $\tau_{per}$ must be executed at least once within the trajectory $\mathcal{L}$ of duration $\mathcal{C_{\mathcal{L}}}$. Furthermore, its frequency $f_{per}$ must not exceed the sensor output frequency $f_{sen}$, ensuring timely updates based on the latest sensor data.
\begin{equation}
    \frac{1}{\mathcal{C_{\mathcal{L}}}} \leq f_{per} \leq f_{sen}
    \label{framework-constraint1}
\end{equation}


\textbf{Constraint 2.} Task Cascade Constraint. As outlined in Eq. \ref{framework-constraint2}, the collision detection task must not operate more frequently than the perception task, as it depends on updated maps. In contrast, the environment difference detection processes raw sensor data and may execute up to the sensor frequency limit.
\vspace{-0.5em}
\begin{equation}
    0 \leq f_{col} \leq f_{per}, 0 \leq f_{dif} \leq f_{sen}
    \label{framework-constraint2}
\end{equation}

\textbf{Constraint 3.} Flight-Time Budget Constraint. To ensure the timely completion of tasks within the flight duration, the total execution time of perception, planning, collision detection, and environment-difference detection must not exceed the budget $\mathcal{C_{\mathcal{L}}} * N$, where $\mathcal{C_{\mathcal{L}}}$ is the duration of current trajectory and $N$ is the number of cores. For a task (say $\tau_{per}$), $f_{per} * \mathcal{C}_{per} * \mathcal{C_{\mathcal{L}}} $ provides its total execution time. Accordingly, we obtain the following constraint by dividing both sides by $\mathcal{C_{\mathcal{L}}}$. Note, the planning task executes once during the current flight, a new trajectory is generated after its execution (Sec. III.A). 
\begin{equation}
    f_{per}*\mathcal{C}_{per} + \frac{\mathcal{C}_{plan}}{\mathcal{C_{\mathcal{L}}}}  + f_{col}*\mathcal{C}_{col} + f_{dif}*\mathcal{C}_{dif} \leq N
    \label{framework-constraint3}
    \vspace{-0.2em}
\end{equation}

\textbf{Constraint 4.} CPU Utilization Constraint. Considering the limited computational capacity, the overall CPU utilization, including the current system utilization $\mathcal{U}_{cur}$ reported by the \textit{System Monitor} and the incremental utilization from task reconfigurations must remain within the total available resource $\mathcal{U}_{total}$. This is formalized in Eq.~\ref{framework-constraint4},
\vspace{-0.5em}
\begin{equation}
    \mathcal{U}_{cur} + \Delta\mathcal{U}_{per} + \Delta\mathcal{U}_{plan} + \Delta\mathcal{U}_{col} + \Delta\mathcal{U}_{dif} \leq \mathcal{U}_{total}
    \label{framework-constraint4}
\end{equation}

\textbf{Discussion on Dynamic Checking.}
Collectively, these constraints provide fundamental bounds for frequency adaptation. They ensure that task adjustments remain feasible within flight dynamics, perception capabilities, and resource limitations, laying the groundwork for the \textit{Frequency Adapter} to compute optimal execution parameters. The dynamic nature of UAV flight, such as sudden changes in environmental complexity or fluctuating computational loads, may pose risks of constraint violation. Therefore, dynamic monitoring of CPU utilization is incorporated to trigger online reconfiguration. This improves robustness against complex scenarios.

\subsection{Frequency Adapter}
\begin{figure}[!htbp]
    \centering
    \vspace{4pt}
    \includegraphics[width=\linewidth]{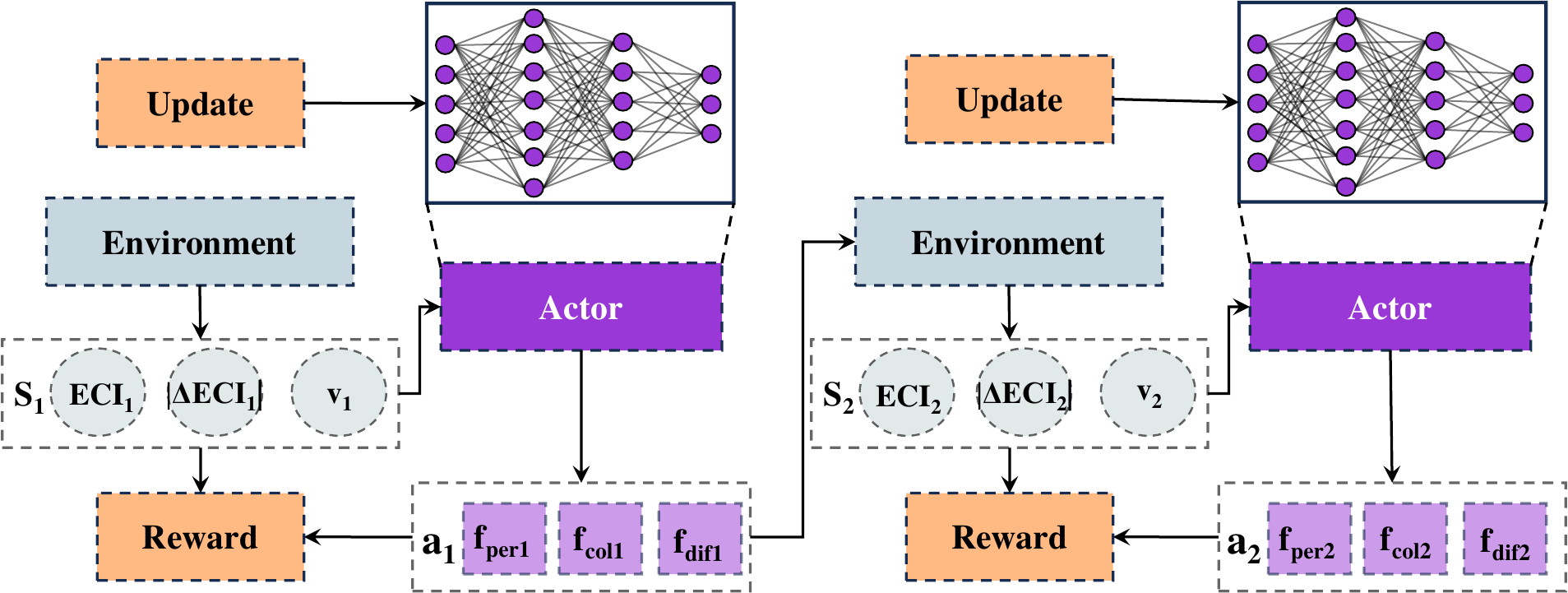}
    \caption{The designed Frequency Adapter network.}
    \vspace{-0.5em}
    \label{framework-action-marker}
\end{figure}

To dynamically adjust task frequencies under varying environmental conditions and resource constraints, the \textit{Frequency Adapter} employs a lightweight reinforcement learning (RL) network to model the relationship between environment dynamics and frequency adjustments, as illustrated in Fig. \ref{framework-action-marker}.

Each input state $s_i$ comprises three components: the current $eci_{i}$, its temporal variation $|\Delta eci_{i}|$, and a status vector $v_i$ encoding the satisfaction of the execution constraints. $eci_{i}$ reflects the spatial complexity of the surroundings, where a higher value indicates a more dense environment that requires a greater perception accuracy. Meanwhile, $|\Delta eci_{i}|$ captures abrupt environmental changes, such as entering or leaving confined spaces, necessitating planning responses. The output action $a_i$ is a tuple representing the execution frequencies $(f_{per}, f_{col}, f_{dif})$ for perception, collision detection, and environment difference detection tasks, respectively. The learning process is guided by two types of rewards.

\textbf{Computing Resource Reward $R_{c}$.} To satisfy onboard resource constraints, $R_{c}$ penalizes the excess of total CPU utilization over the available capacity, as in Eq. \ref{framework-r-comput},
\begin{equation}
        R_{c} = \mathcal{U}_{per}+\mathcal{U}_{plan}+\mathcal{U}_{col}+\mathcal{U}_{dif}- \mathcal{U}_{total}
    \label{framework-r-comput}
\end{equation}

\textbf{Flight Performance Reward $R_{f}$.} To encourage efficient flight, $R_{f}$ evaluates path quality, target reaching accuracy, and collision avoidance, displayed in Eq. \ref{framework-r-fli}. $\mathcal{\tilde{P}}_e$ and $\mathcal{\tilde{L}}_{e}$ denote the actual position and the flight path at epoch $e$, while $\mathcal{P}$ and $\mathcal{\tilde{L}}$ denote the intended target and the optimal path, respectively. $w_{a}, w_{s}$ and $w_{col}$ are reward weights. $\mathbb{I}_{Col}$ is an indicator function set to 1 if a collision occurs.
\begin{equation}
    \begin{split}
        &R_{f} = 
            w_{a}  ||\mathcal{P}-\mathcal{\tilde{P}}_{e}||_2+ w_{s}(1-\frac{||\mathcal{\tilde{L}}_{e}||}{||\mathcal{\tilde{L}}_{e-1}||}) + w_{col} \mathbb{I}_{col}
    \end{split}
    \label{framework-r-fli}
\end{equation}
The loss function to optimize the model is given as follows.
\begin{equation}
    L(\theta) = -\sum_{i=1}^{T} [\nabla_{\theta} \log \pi_{\theta}(a_i | s_i) \cdot (R_{c} + R_{f})]
    \label{framework-loss}
\end{equation}
\textbf{Characteristic 2.} \textit{To address the rigidity of fixed task configurations identified in Limitation 2, E-Navi introduces an adaptation mechanism that dynamically reconfigures the perception and planning pipeline. Using the Resolution Adapter and the Frequency Adapter, the system jointly considers environmental complexity metrics ($eci$ and $|\Delta eci|$) and platform-specific computational constraints to adjust the task configuration at runtime. This ensures more efficient resource utilization and robust flight performance under various conditions.}

\textbf{Implementation.} We adopt a lightweight RL model based on a two-layer MLP, where both the policy and value networks consist of two fully connected layers with 150 neurons and tanh activations. The reward function, jointly encourages smooth velocity, high mission success rates, and low CPU utilization. The state space incorporates a compact of the system’s status, including $eci$, $|\Delta eci|$, frequencies of perception and planning, UAV motion states, workload utilization, and target waypoints. The action space is discretized with sufficient granularity to approximate continuous space, facilitating stable learning while maintaining efficiencies. This design enables policy deployed across different platforms by explicitly incorporating computational status into the state representation and encouraging resource efficient behavior through the reward function. The Proximal Policy Optimization (PPO) algorithm \cite{son2023gradient} is employed with Actor-Critic architecture.

\vspace{-0.5em}
\subsection{Summary and Discussion}
This section summarizes the system from two aspects: timing guarantee and computational load.

\textbf{Timing Guarantees.}
To ensure the timing guarantees of the UAV system scheduled under the EDF policy \cite{liu1973scheduling}, we enforce that the total system utilization does not exceed the $U_{\text{total}}$. At run-time, E-Navi computes the execution budgets for perception and planning within the trajectory duration $\mathcal{C}_{\mathcal{L}}$, in compliance with this constraint while considering the current utilization $\mathcal{U}_{cur}$ obtained by the \textit{System Monitor}. In rare cases where the system utilization is higher than $U_{total}$, the system scales down task periods proportionally to enforce the constraint. This adjustment is lightweight and has negligible impact on UAV behavior. Furthermore, if the \textit{System Monitor} detects persistent overload during execution, a fail-safe mechanism is triggered, causing the UAV to enter a hovering mode to prevent unsafe maneuvers and avoid potential harm to the surrounding environment. In Sec. VI.B, we reported the deadline miss ratio measured from different flight scenarios, justifying the effectiveness of the proposed framework.


\textbf{Computation Load Discussion.}
The use of $eci$ and its temporal gradient $\Delta eci$ provides a compact metric for environmental risk and volatility. Higher $eci$ prompts higher perceptual resolution and frequency, while lower values enable task down-scaling. By linking $eci$ to system configurations, E-Navi aligns computational effort with task demands, enabling efficient resource utilization.


%% file: tex/experiment.tex
To evaluate the performance of E-Navi, extensive experiments are conducted on both Hardware-In-the-Loop (HIL) platforms and real-world scenarios. 
Due to the unavailability of RoboRun's implementation (see Sec. II.A), representative test cases are designed to benchmark the system against the method \cite{wang2022geometrically} with Ego-Planner \cite{zhou2020ego} applied as the trajectory generation module. This method serves as the baseline due to its open-source, full-stack design and widespread adoption in autonomous UAV systems. Specifically, it follows a modular design in a perception–planning–control pipeline: A depth sensor captures raw environmental data, which is processed into a local occupancy map for obstacle representation (perception). Based on this map, the system generates feasible and collision-free trajectories in 3D space (planning) using the optimization method. A controller tracks and executes the trajectory (control). For the configurations, the baseline software stack operates with fixed perception and planning frequencies of 30 Hz and 20 Hz, respectively, and uses a fixed map resolution of 0.1 meter~\cite{wang2022geometrically}. 


\begin{figure}[!htbp]
    \centering
    \includegraphics[width=.95\linewidth]{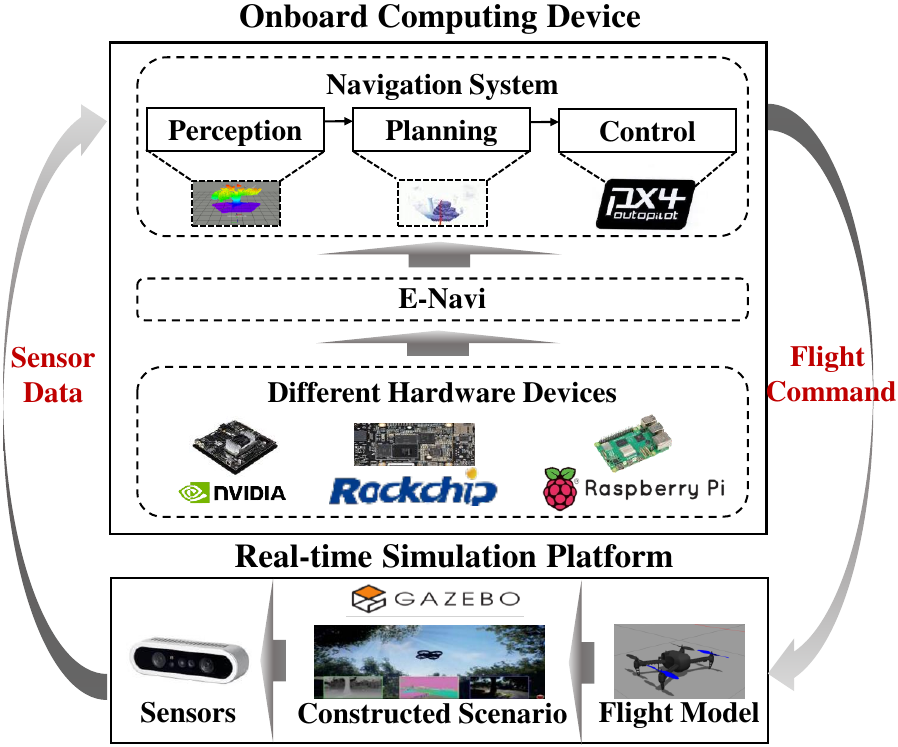}
    \caption{Our designed UAV-oriented HIL platform.}
    \setlength{\abovecaptionskip}{-2pt}
    \setlength{\belowcaptionskip}{-5pt}
    \vspace{-0.1em}
    \label{HIL}
\end{figure}

The evaluations are carried out across a variety of computing devices in commercial UAVs including NVIDIA Jetson Xavier NX, Jetson TX2, OrangePi 5, Raspberry Pi 4B (based on Arm), and an x86-based board. These devices span a wide spectrum of computational capabilities. Notably, the Jetson series is widely used in UAV by companies\cite{nvidiaJetson}. The OrangePi 5, powered by the Rockchip RK3588S SoC is gaining popularity in open-source UAV and robotics projects \cite{orangepi5}. The Raspberry Pi 4B remains a widely adopted choice in low-cost UAV designs and academic research \cite{raspberryPi}. x86-based platforms are commonly used in robotic systems \cite{x86}. 

In our experiments, 
The computational specifications of the boards are summarized in Tab. \ref{exp-devices}. For OrangePi 5 and TX2, the 4 performance cores are used. Since all tasks are executed on the CPU, Tab. \ref{exp-devices} reports the theoretical GFLOPs to represent CPU computing capabilities, estimated by multiplying the number of cores, clock frequency, and FLOPs per cycle. The DVFS are disabled with the performance governor. 

\begin{table}[!ht]
    \centering
    \caption{Platforms with varying CPU capabilities.}
    \begin{tabular}{cccc}
      \hline
      \textbf{Device}  & \textbf{CPU} & \textbf{Memory} & \textbf{GFLOPS} \\
      \hline
      OrangePi 5 & 4 $\times$ Cortex-A76 @2.4GHz & 16 GB & 19.2\\
      \hline
      Xavier NX & 6 $\times$ Carmel @1.4GHz & 8.0 GB & 16.8\\
      \hline
      x86 Board & 2 $\times$ i7-6500U @2.5GHz & 8.0 GB & 16.6\\
      \hline
      NVIDIA TX2  & 4 $\times$ Cortex-A57 @2.0GHz & 8.0 GB & 16.0\\
      \hline
      RaspberryPi 4B & 4 $\times$ Cortex-A72 @1.8GHz & 8.0 GB & 14.4\\
      \hline
    \end{tabular}
    \label{exp-devices}
\end{table}

For deployment, training is performed on the HIL platform, where real-world environments are abstracted into structured features to generate a set of editable common simulation scenes. These scenes, covering diverse settings such as urban streets, parks, indoor halls, are built in Gazebo and integrated with real onboard hardware. This setup enables training under realistic conditions. Therefore, the trained model demonstrates similar performance in both simulation and real-world experiments without further fine-tuning.
In addition, the SCHED\_DEADLINE policy is applied during execution to schedule the perception and planning tasks. The periods of tasks are configured via \textit{attr.sched\_period} in \textit{sched setattr()} based on the values produced by E-Navi, whereas the deadlines of tasks are set equal to their periods by updating the \textit{attr.sched\_deadline}. The \textit{attr.sched\_runtime} (i.e., the Worst-Case Execution Time) of a task is measured offline and remains fixed during execution.

\subsection{Experimental Setup}
The HIL simulation environment is hosted on a high-performance desktop equipped with an AMD Ryzen5 3600 CPU and an NVIDIA TITAN-V GPU. For onboard execution, the devices in Tab. \ref{exp-devices} are used to perform the E-Navi system.

\begin{table}[!htbp]
    \centering
    \caption{Descriptions of the five experimental scenarios.}
    \begin{tabular}{ccc}
    \hline
      \textbf{Scenario}  & \textbf{Description} & \textbf{$eci$ Range} \\
      \hline
      S1 & City with varying height buildings & [0.06-0.87]\\
      \hline
      S2 & Park with tress and a pavilion & [0.08-0.97]\\
      \hline
      S3 & Indoor hall with varying furniture & [0.19-0.83]\\
      \hline
      S4 & City with similar height buildings & [0.26-0.86]\\
      \hline
      S5 & Village with house, bridge, and river & [0.25-0.79]\\
      \hline
    \end{tabular}
    \label{exp-scenario-dis}
\end{table}

\begin{figure*}[!htbp]
    \centering{
        \subfigure[varying-height buildings]{
            \includegraphics[width=3.5cm, height=4cm]{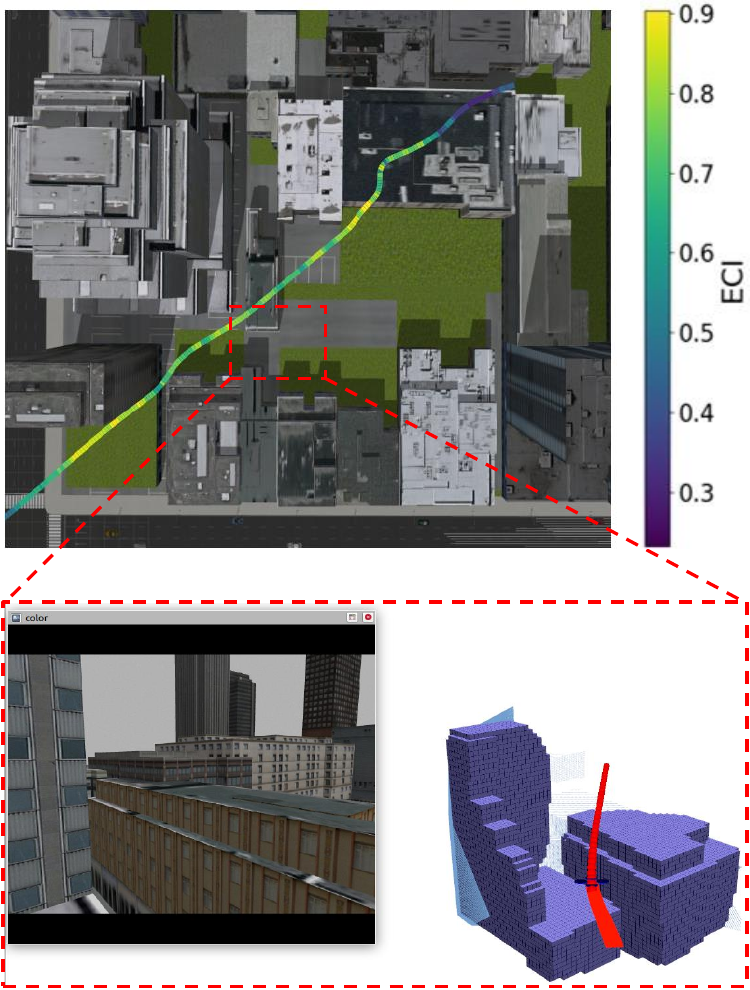}
            \label{exp-buildings-different}
        }\hspace{-0.3cm}
        \subfigure[park]{
            \includegraphics[width=3.5cm, height=4cm]{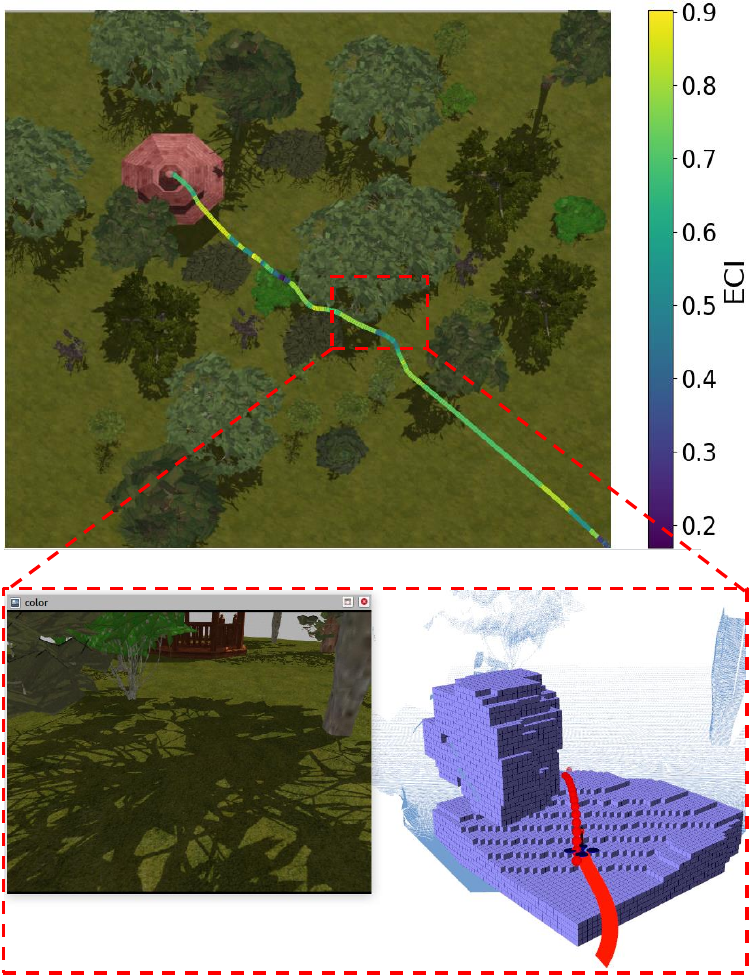}
            \label{exp-park}
        }\hspace{-0.3cm}
        \subfigure[indoor hall]{
            \includegraphics[width=3.5cm, height=4cm]{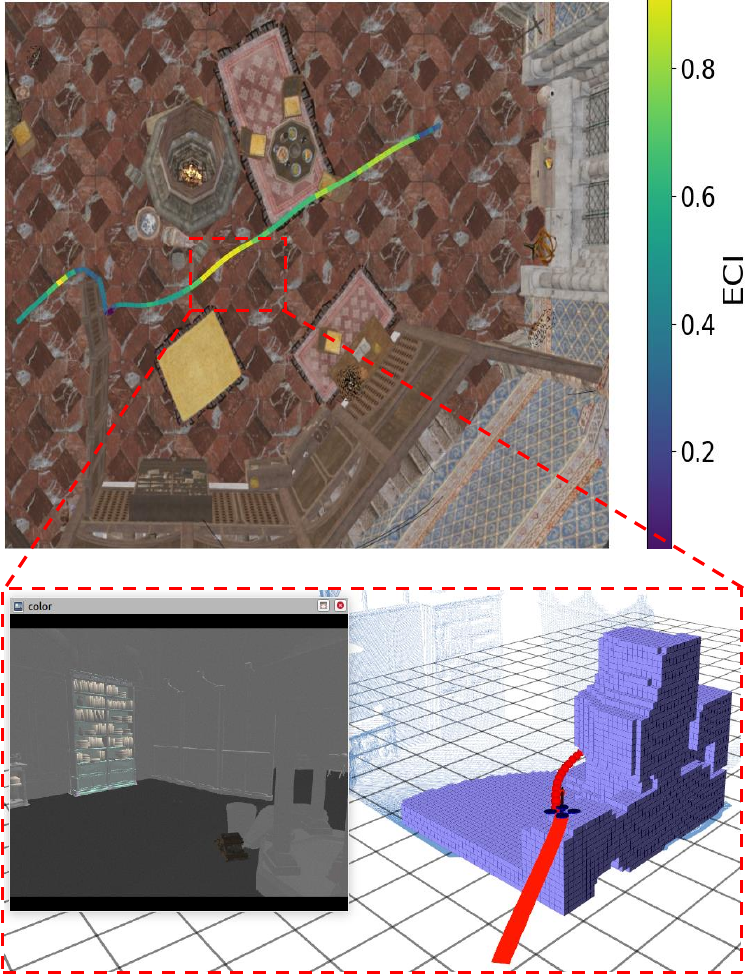}
            \label{exp-indoor}
        }\hspace{-0.3cm}
        \subfigure[similar-height buildings]{
            \includegraphics[width=3.5cm, height=4cm]{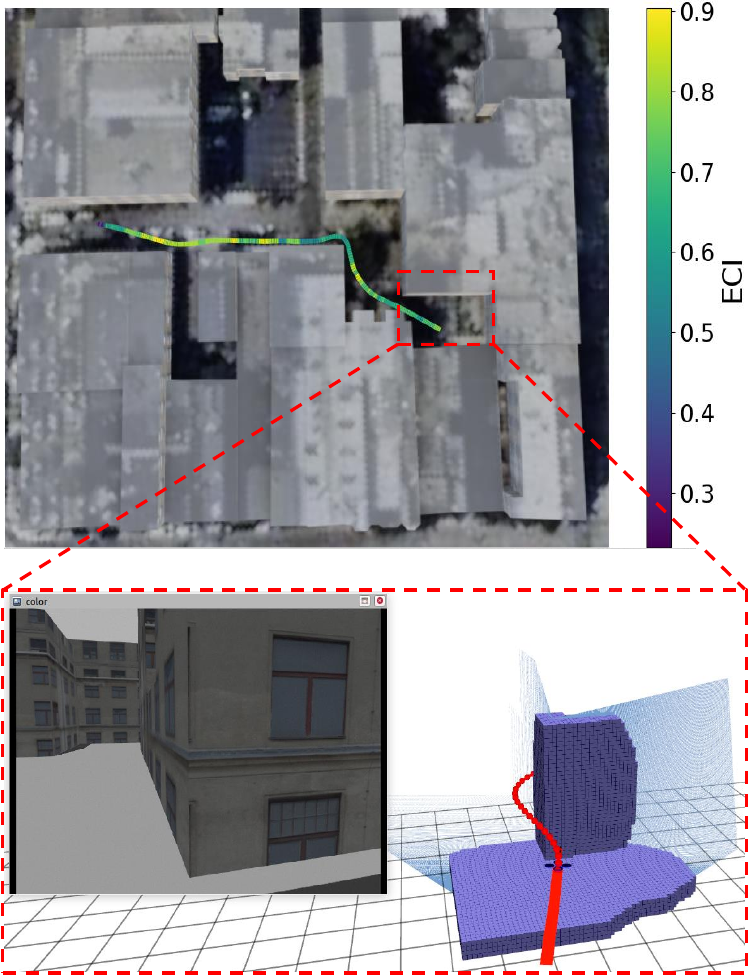}
            \label{exp-buildings-similar}
        }\hspace{-0.3cm}
        \subfigure[village]{
            \includegraphics[width=3.5cm, height=4cm]{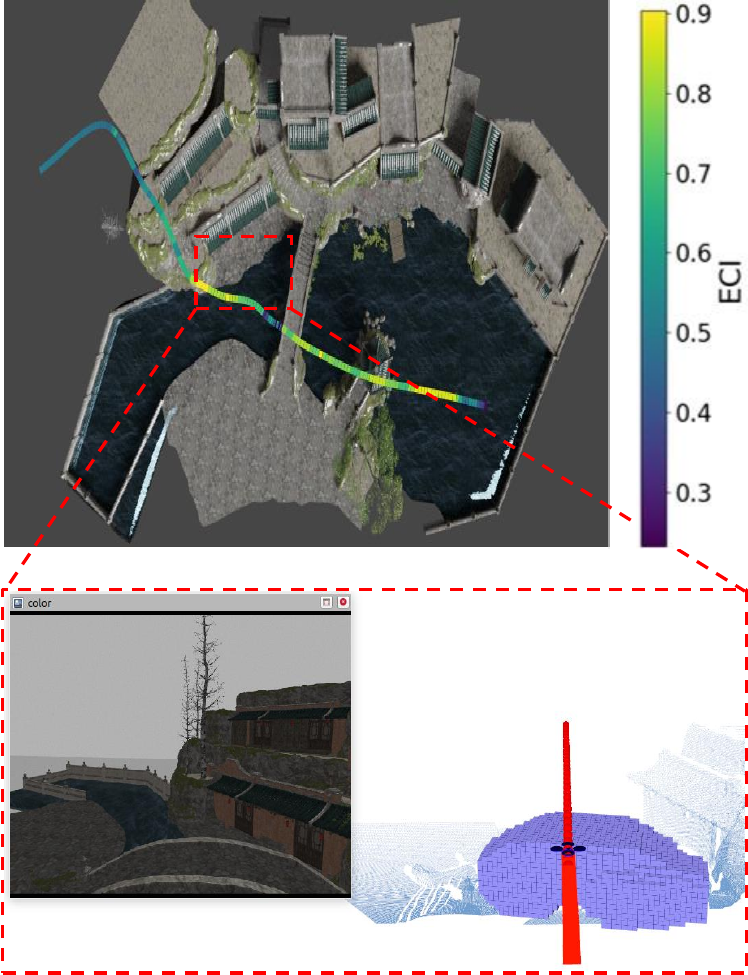}
            \label{exp-country}
        }
    }
    \setlength{\abovecaptionskip}{-2pt}
    \setlength{\belowcaptionskip}{-5pt}
    \caption{The five representative scenarios for evaluation. In each scenario, the first view images, the constructed maps, and the bird's-eye view images with corresponding trajectories where different colors indicate the $eci$.}
    \label{exp-hil-path}
\end{figure*}

\textbf{HIL Platform.}
Hardware-in-the-loop (HIL) simulation has become an indispensable methodology for verifying UAV systems, as it allows performance evaluation through hardware-software co-simulation before real-world deployment. As illustrated in Fig. \ref{HIL}, our UAV-oriented HIL platform consists of two main components: (1) a simulation environment that generates dynamic scenarios and corresponding sensor data and (2) embedded onboard computing devices that perform navigation system computations. This closed-loop system continuously feeds synthetic sensor inputs into the onboard systems, which control commands that are in turn applied to a virtual UAV model, providing high-fidelity, repeatable evaluation results.

\begin{figure}[!htbp]
    \centering
    \includegraphics[width=\linewidth]{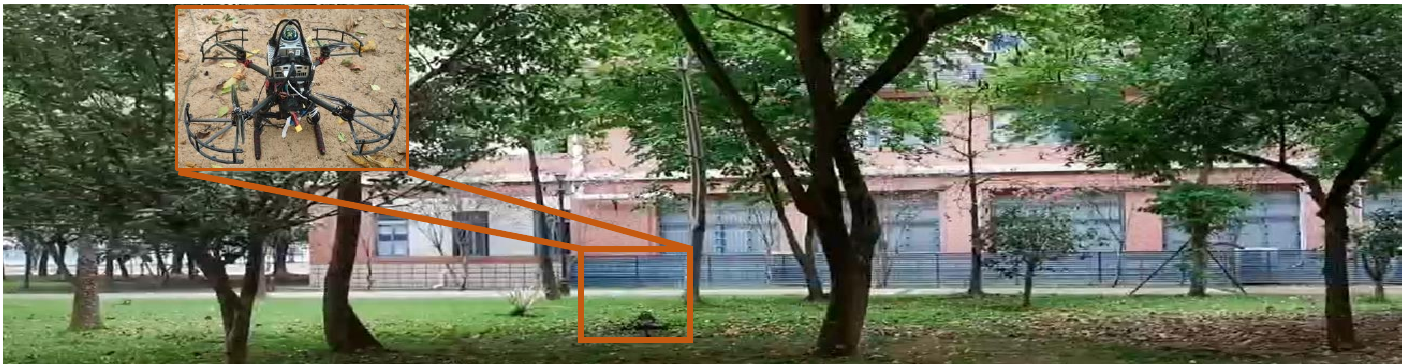}
    \caption{Our UAV platform and real-world scenarios.}
    \setlength{\abovecaptionskip}{-2pt}
    \setlength{\belowcaptionskip}{-5pt}
    \label{rw-d}
    \vspace{-1em}
\end{figure}

\textbf{Real-world Case Study.} To further validate E-Navi in practical settings, we conduct real-world experiments in an open campus environment with scattered trees. As shown in Fig. \ref{rw-d}, our UAV is based on an X500 frame and equipped with a PX4 flight controller, an Intel-Realsense depth camera, and an embedded x86 board.

\subsection{Navigation Experiments on HIL Platform}
To evaluate the E-Navi, a set of experiments is conducted in five representative scenarios with varied obstacle distributions and ECI ranges described in Tab. \ref{exp-scenario-dis}. The flight trajectories in each environment are visualized in Fig. \ref{exp-hil-path}. These scenarios are specifically designed to capture a diverse range of environmental complexities and structural patterns, enabling systematic assessment under realistic UAV navigation challenges.

The evaluation is organized as follows. We first describe the characteristics of the experimental scenarios. Next, we present and analyze the overall performance metrics, including the utilization of E-Navi navigation tasks (perception, planning, collision detection, difference detection) measured by the \texttt{psutil.Process(pid).cpu\_percent()} API, flight time, and flight path. Finally, we examine the timing costs, including $eci$ computation and adapter execution, under varying $eci$ conditions. 

In addition, we report the utilization of the navigation tasks while excluding the ECI computation and background activities such as OS kernel threads, device drivers, and middleware services. 
This enables a clearer comparison of how different navigation task configurations affect flight performance under the E-Navi and the baseline method, which also demonstrates the adaptive capability of E-Navi.
As shown later in this section, the costs of ECI computation and the background activities would not affect the effectiveness and the applicability of E-Navi on resource-constrained UAV platforms. 

\subsubsection{Scenario Description}
Fig. \ref{exp-hil-path} presents a bird's eye view of the UAV flight trajectories overlaid on $eci$ heatmaps, along with front-view snapshots of each environment. The color distribution indicates the computed $eci$, where warmer colors represent a higher environmental complexity.

The five scenarios span diverse environments: irregular urban layouts (S1), tree-filled parks with occlusions (S2), cluttered indoor halls (S3), visually uniform urban grids (S4), and rural areas with mixed elevations, and structures (S5), each posing distinct navigation challenges.

\begin{figure*}[!htbp]
    \centering{
        \subfigure[Pi5 in S1]{
            \includegraphics[width=.193\linewidth]{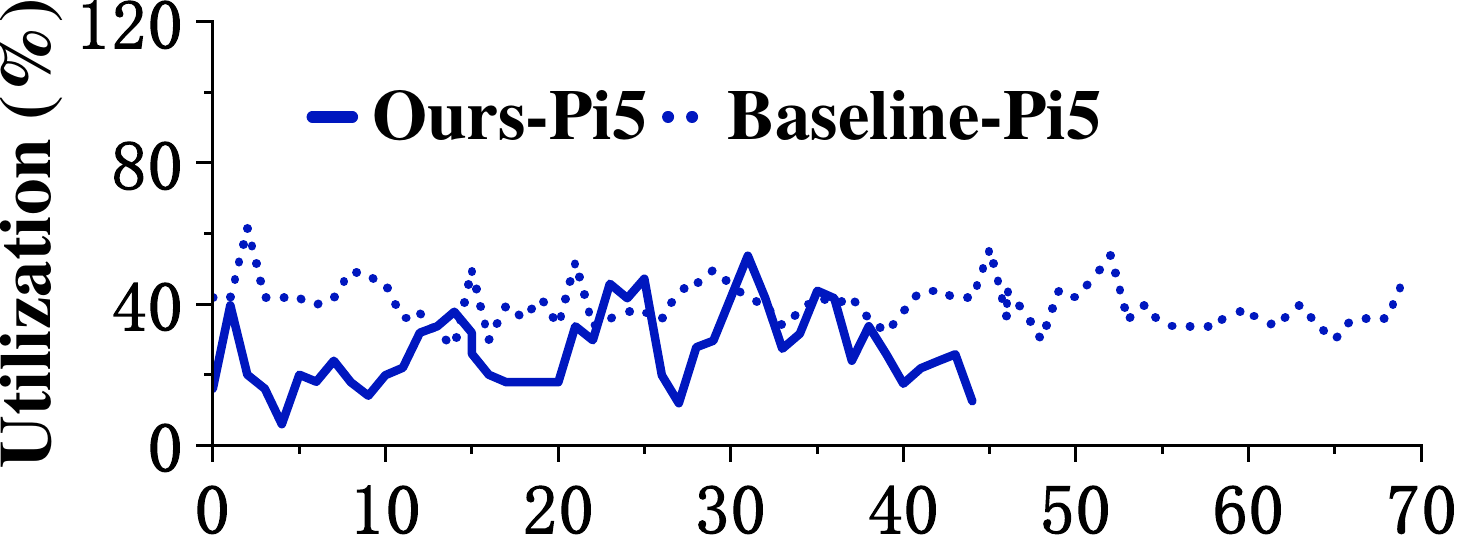}
            \label{pi5-c-s1}
        }
        \hspace{-4mm}
        \subfigure[Pi5 in S2]{
            \includegraphics[width=.193\linewidth]{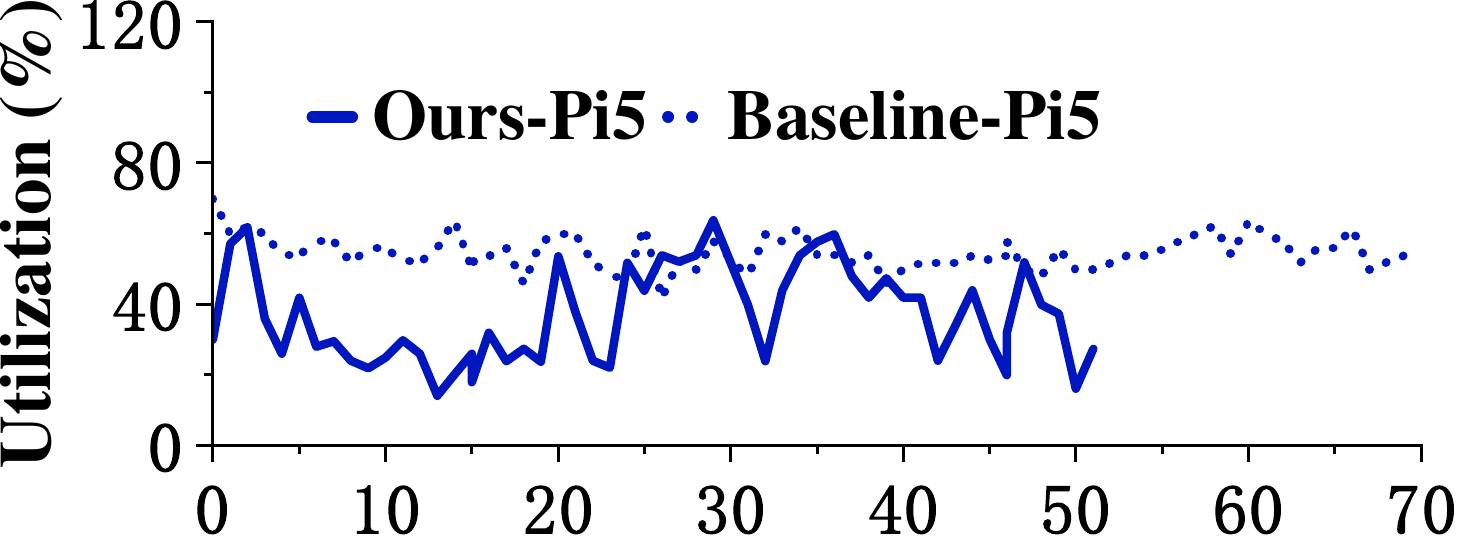}
            \label{pi5-c-s2}
        }
        \hspace{-4mm}
        \subfigure[Pi5 in S3]{
            \includegraphics[width=.193\linewidth]{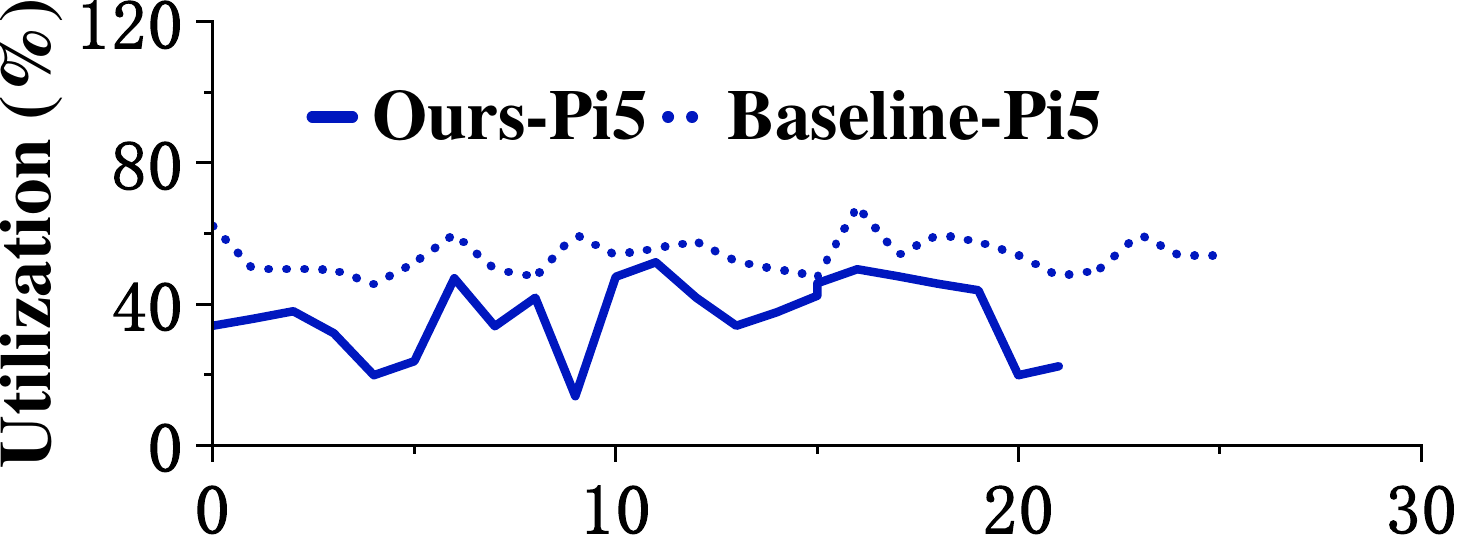}
            \label{pi5-c-s3}
        }
        \hspace{-4mm}
        \subfigure[Pi5 in S4]{
            \includegraphics[width=.193\linewidth]{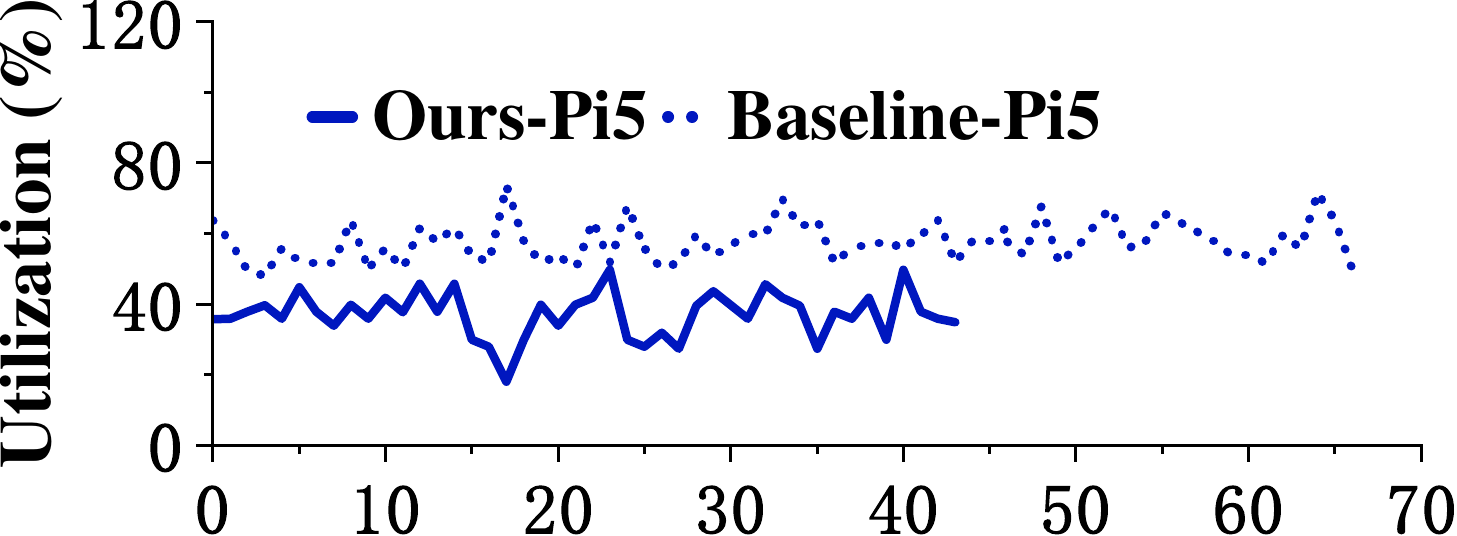}
            \label{pi5-c-s4}
        }
        \hspace{-4mm}
        \subfigure[Pi5 in S5]{
            \includegraphics[width=.193\linewidth]{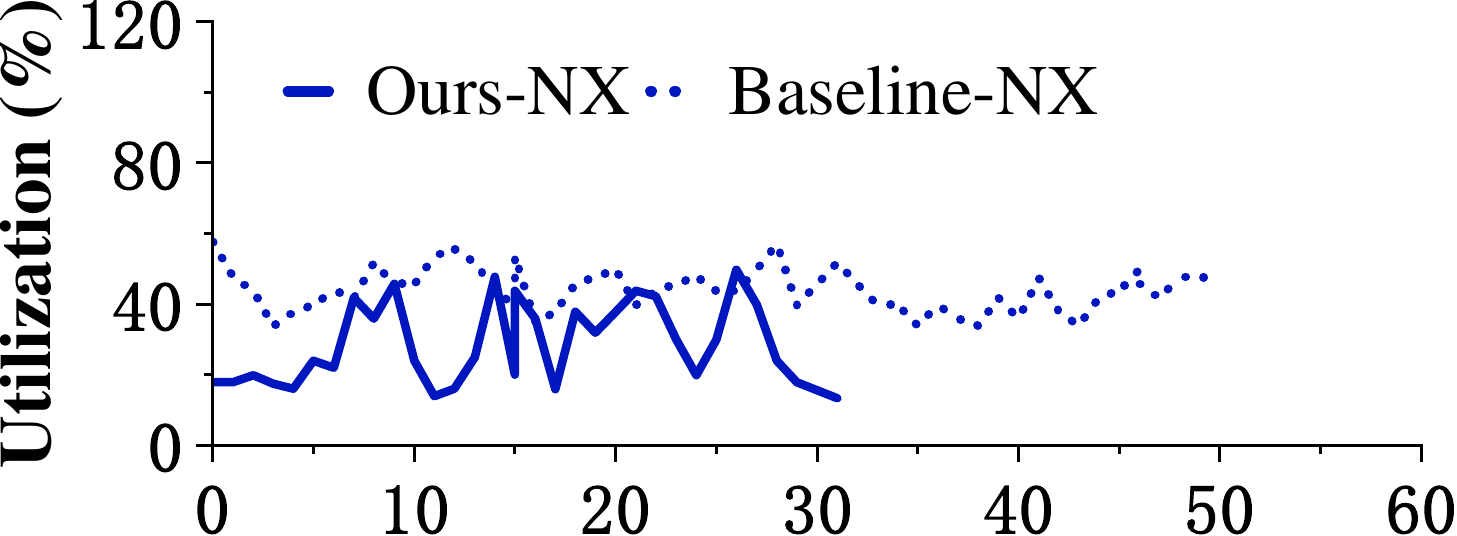}
            \label{pi5-c-s5}
        }
                \subfigure[NX in S1]{
        	\includegraphics[width=.193\linewidth]{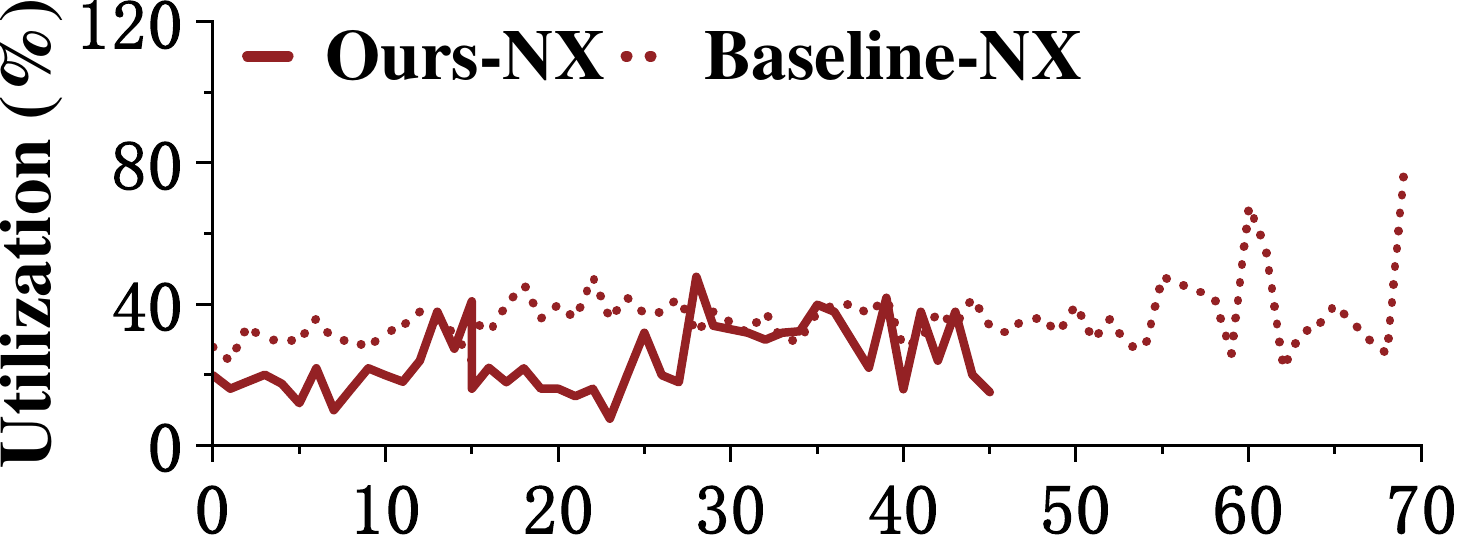}
        	\label{nx-c-s1}
        }
        \hspace{-4mm}
        \subfigure[NX in S2]{
        	\includegraphics[width=.193\linewidth]{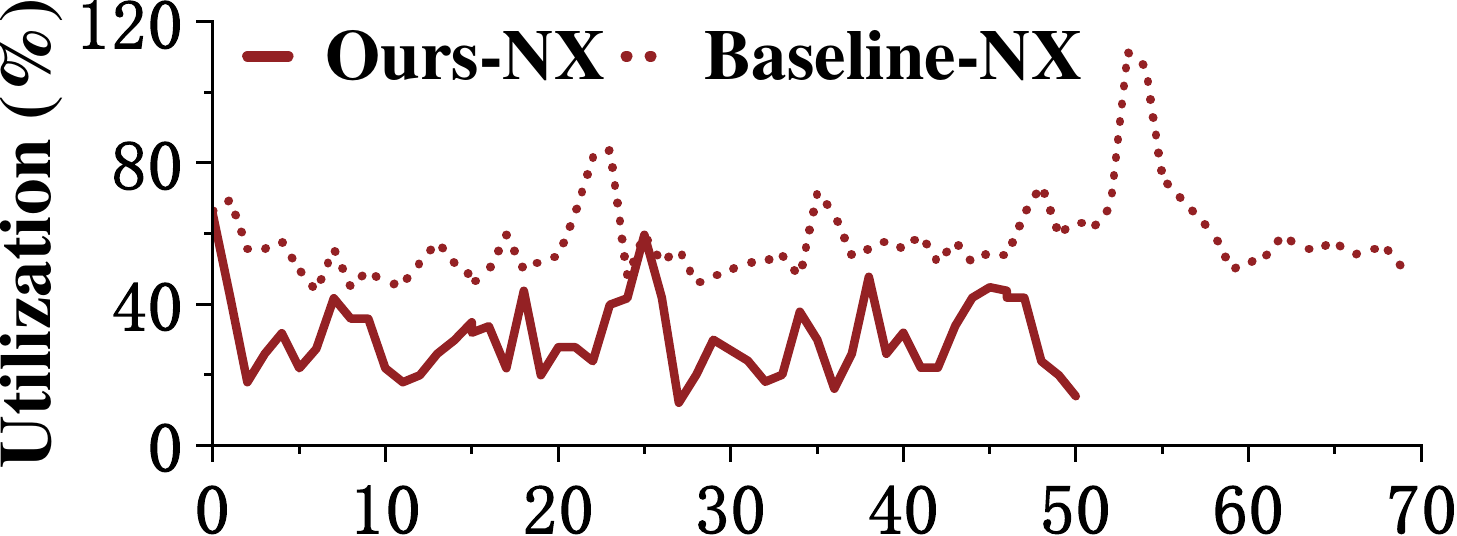}
        	\label{nx-c-s2}
        }
        \hspace{-4mm}
        \subfigure[NX in S3]{
        	\includegraphics[width=.193\linewidth]{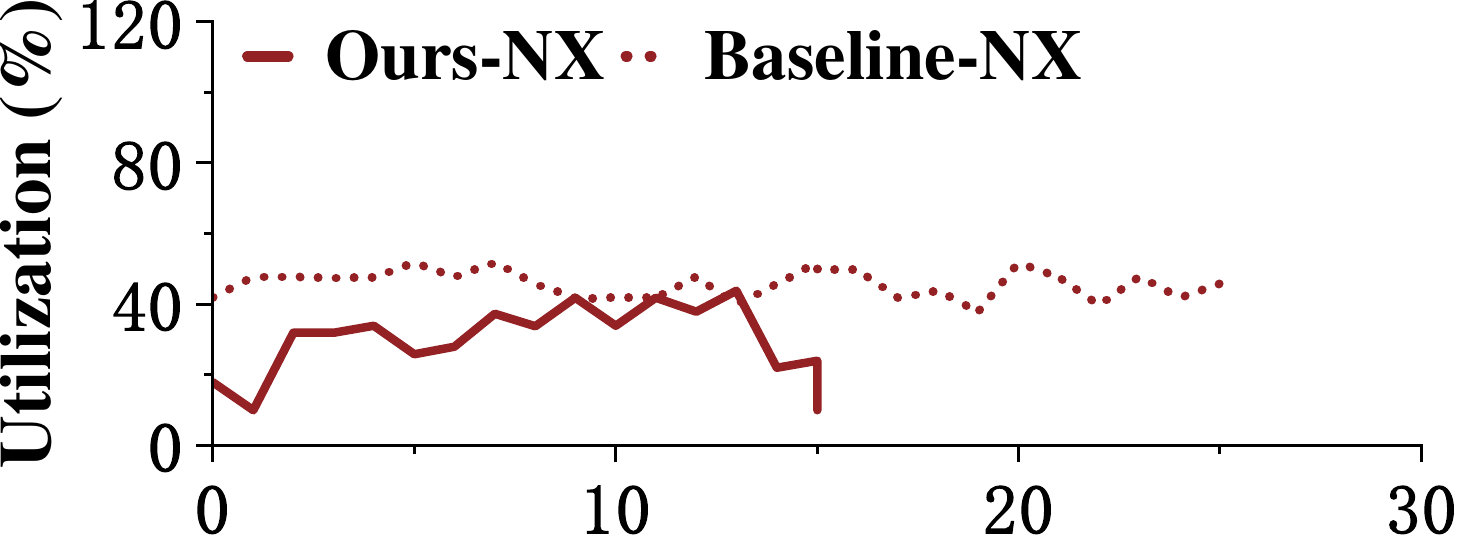}
        	\label{nx-c-s3}
        }
        \hspace{-4mm}
        \subfigure[NX in S4]{
        	\includegraphics[width=.193\linewidth]{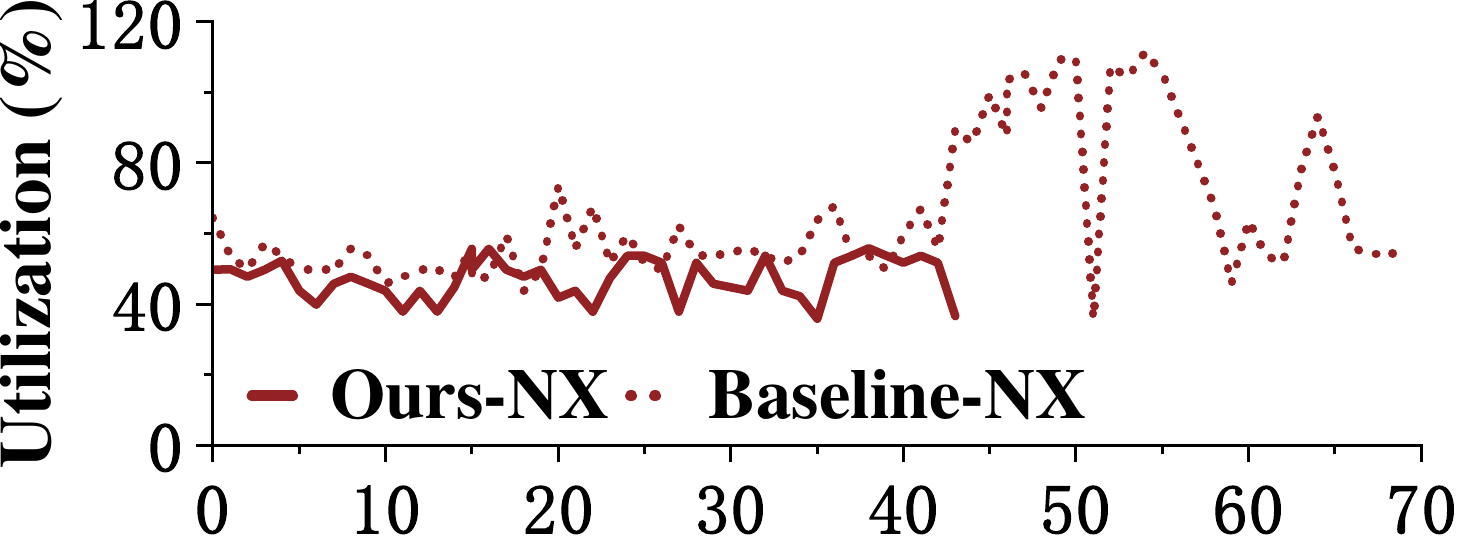}
        	\label{nx-c-s4}
        }
        \hspace{-4mm}
        \subfigure[NX in S5]{
        	\includegraphics[width=.193\linewidth]{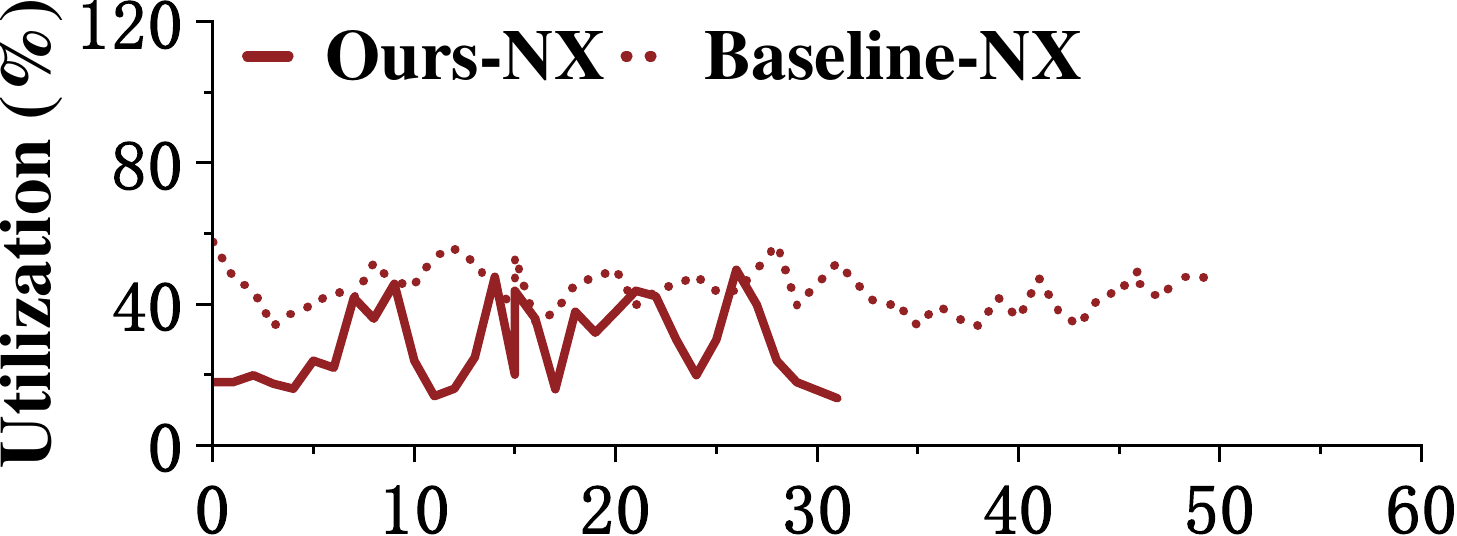}
        	\label{nx-c-s5}
        }
        \subfigure[x86 in S1]{
        	\includegraphics[width=.193\linewidth]{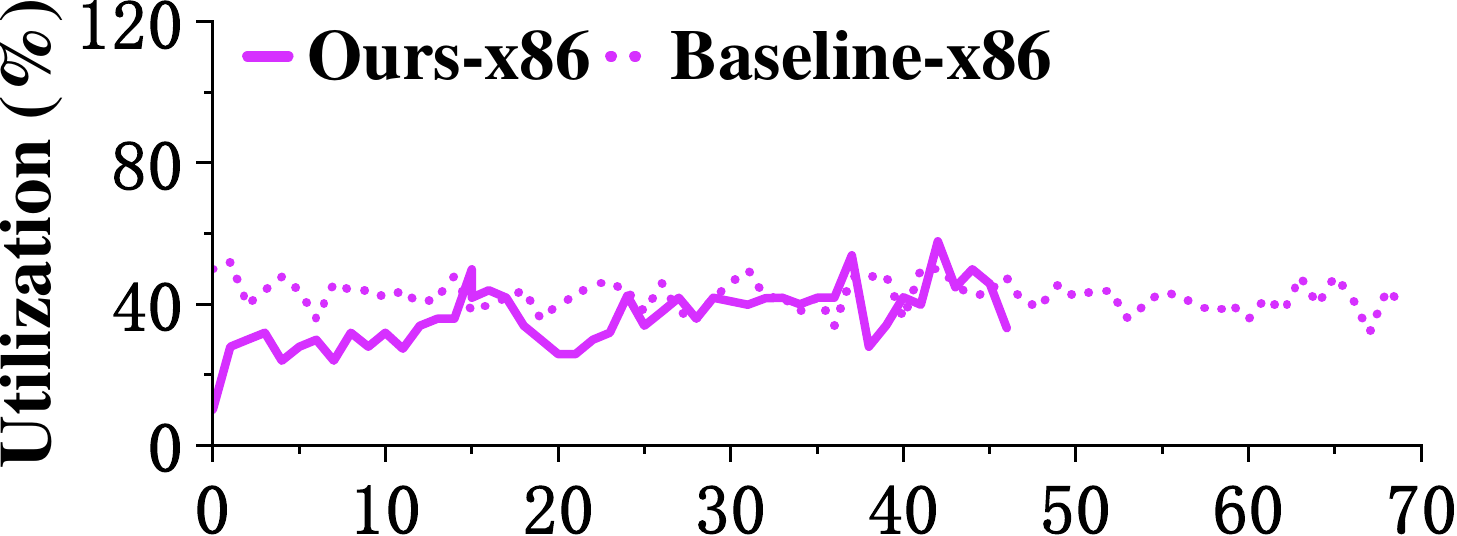}
        	\label{x86-c-s1}
        }
        \hspace{-4mm}
        \subfigure[x86 in S2]{
        	\includegraphics[width=.193\linewidth]{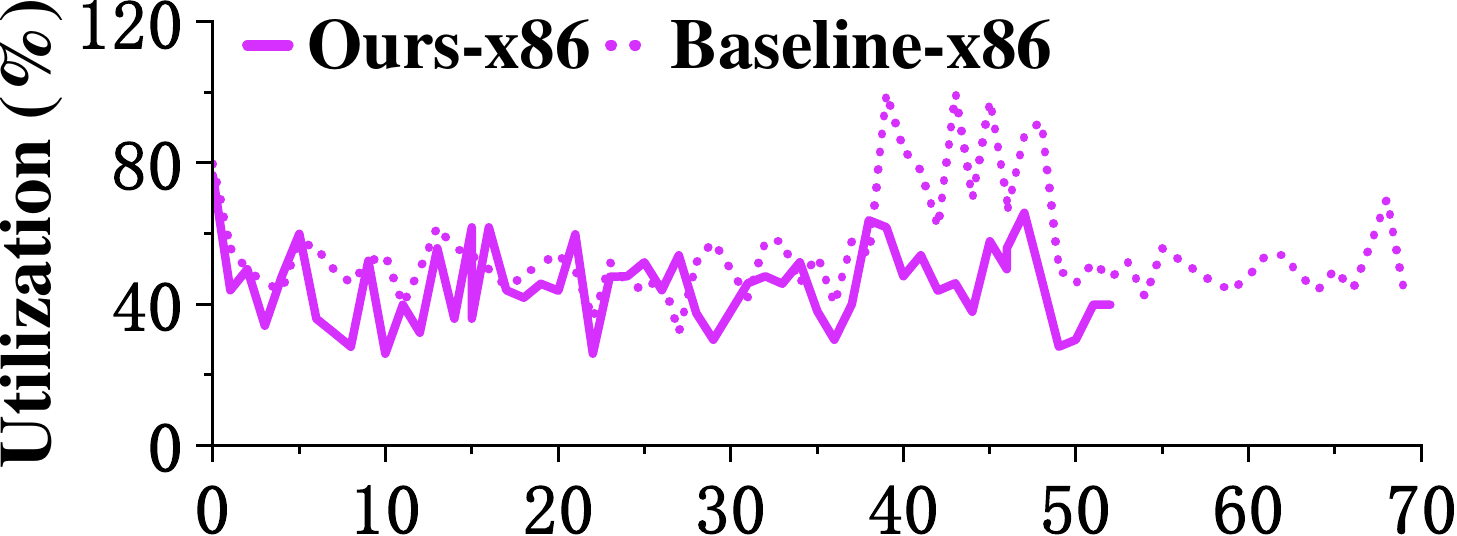}
        	\label{x86-c-s2}
        }
        \hspace{-4mm}
        \subfigure[x86 in S3]{
        	\includegraphics[width=.193\linewidth]{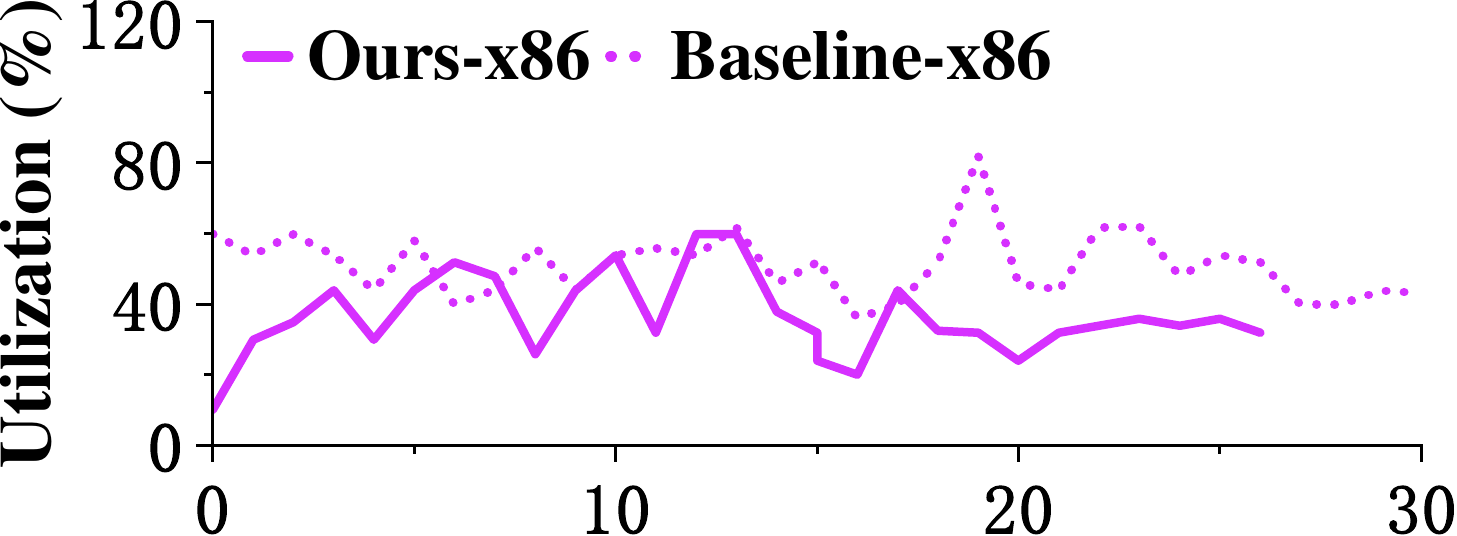}
        	\label{x86-c-s3}
        }
        \hspace{-4mm}
        \subfigure[x86 in S4]{
        	\includegraphics[width=.193\linewidth]{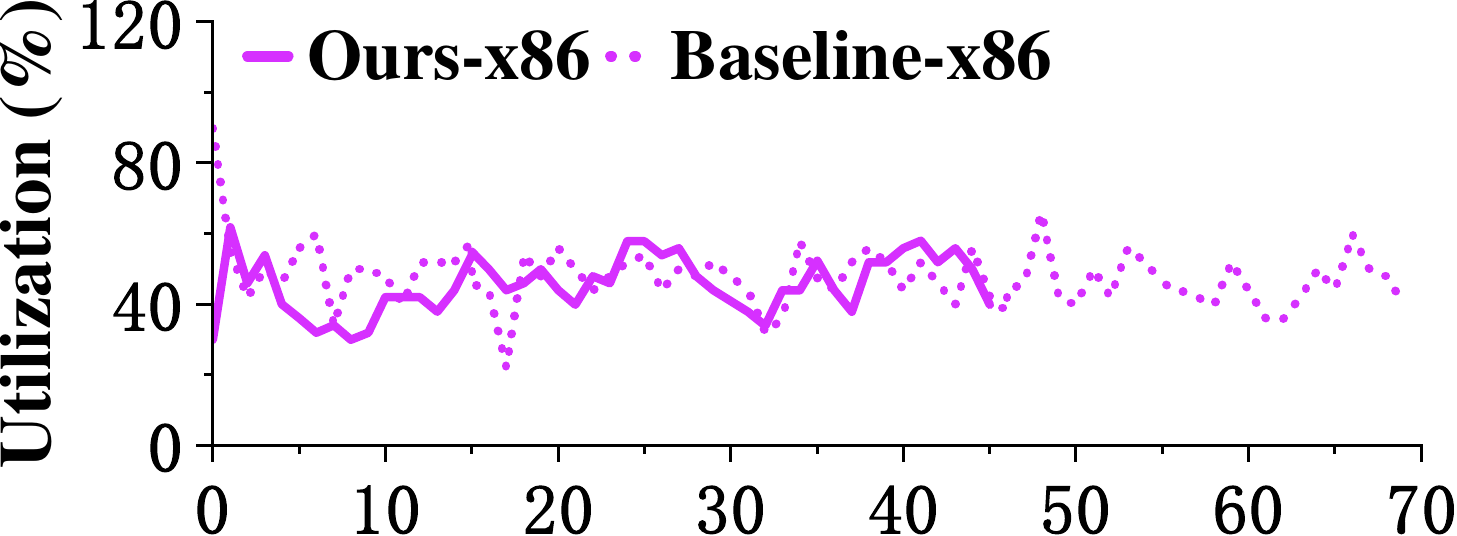}
        	\label{x86-c-s4}
        }
        \hspace{-4mm}
        \subfigure[x86 in S5]{
        	\includegraphics[width=.193\linewidth]{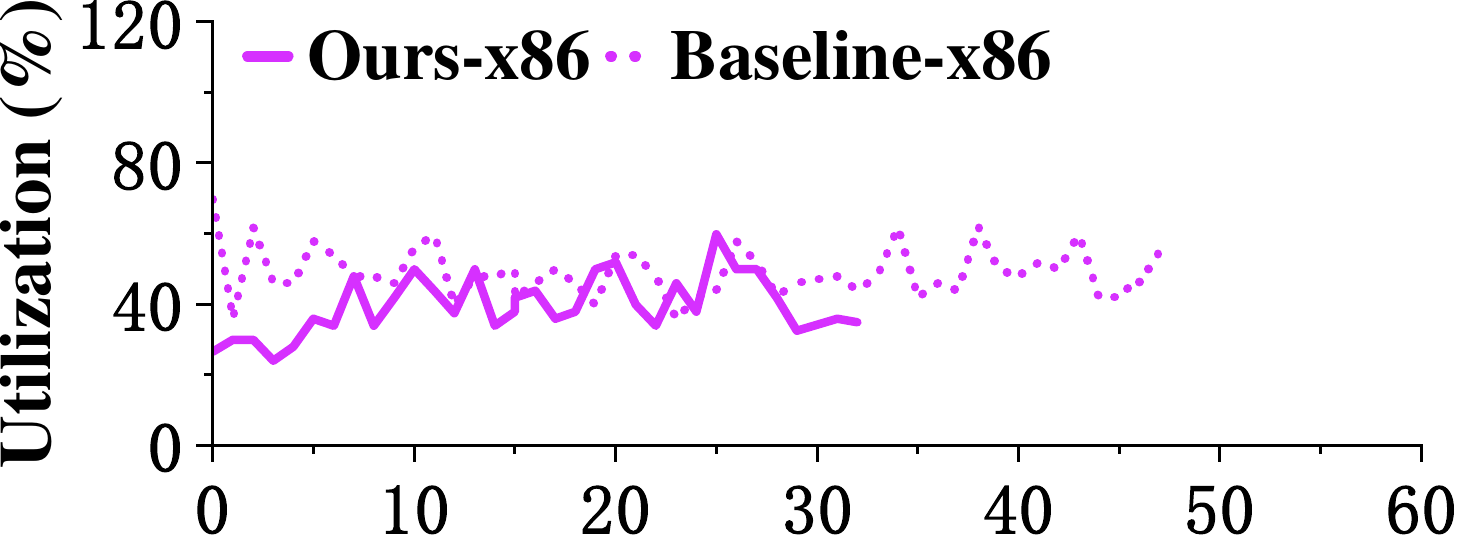}
        	\label{x86-c-s5}
        }
        \subfigure[TX2 in S1]{
            \includegraphics[width=.193\linewidth]{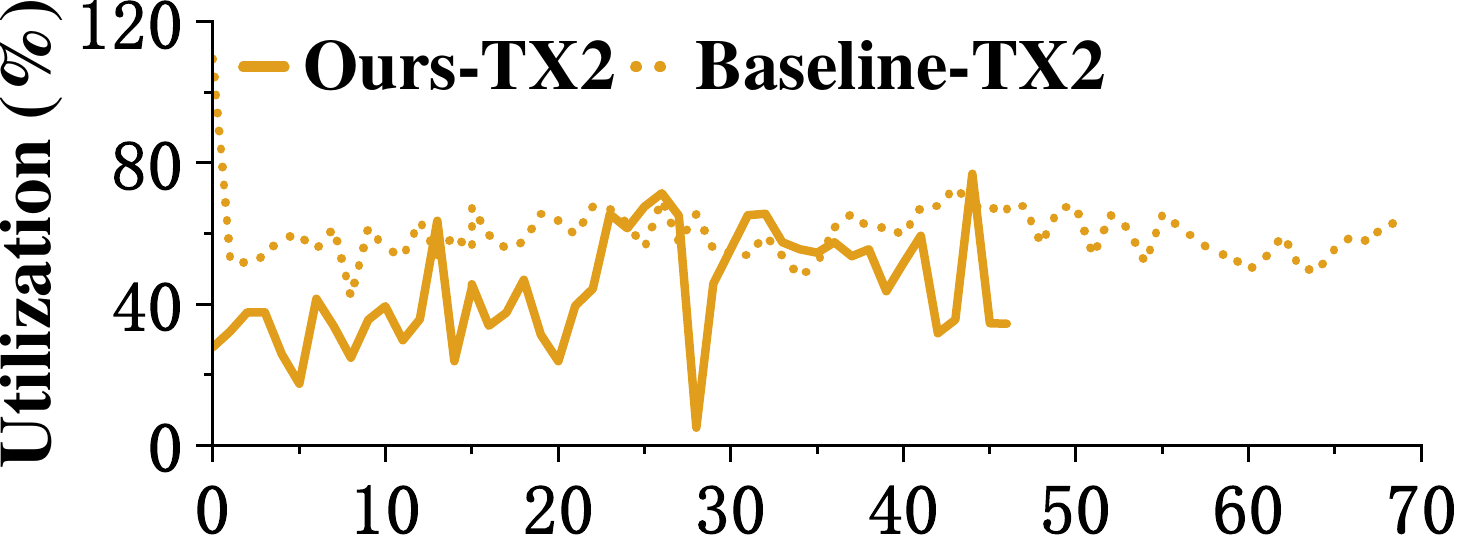}
            \label{tx2-c-s1}
        }
        \hspace{-4mm}
        \subfigure[TX2 in S2]{
            \includegraphics[width=.193\linewidth]{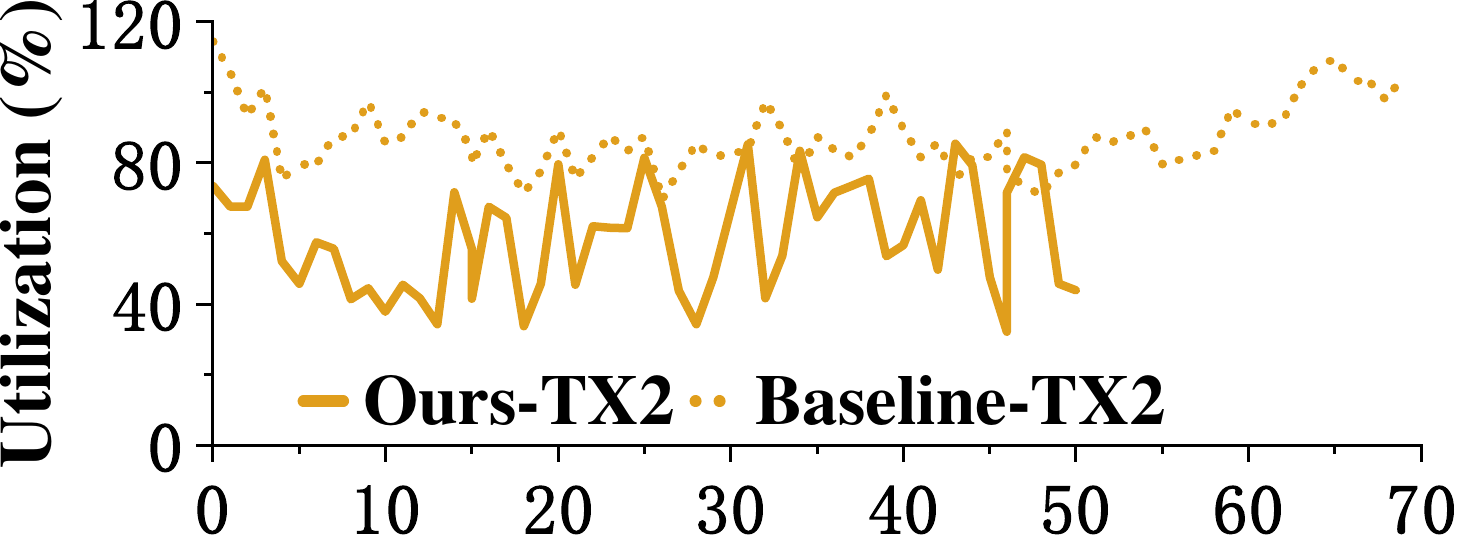}
            \label{tx2-c-s2}
        }
        \hspace{-4mm}
        \subfigure[TX2 in S3]{
            \includegraphics[width=.193\linewidth]{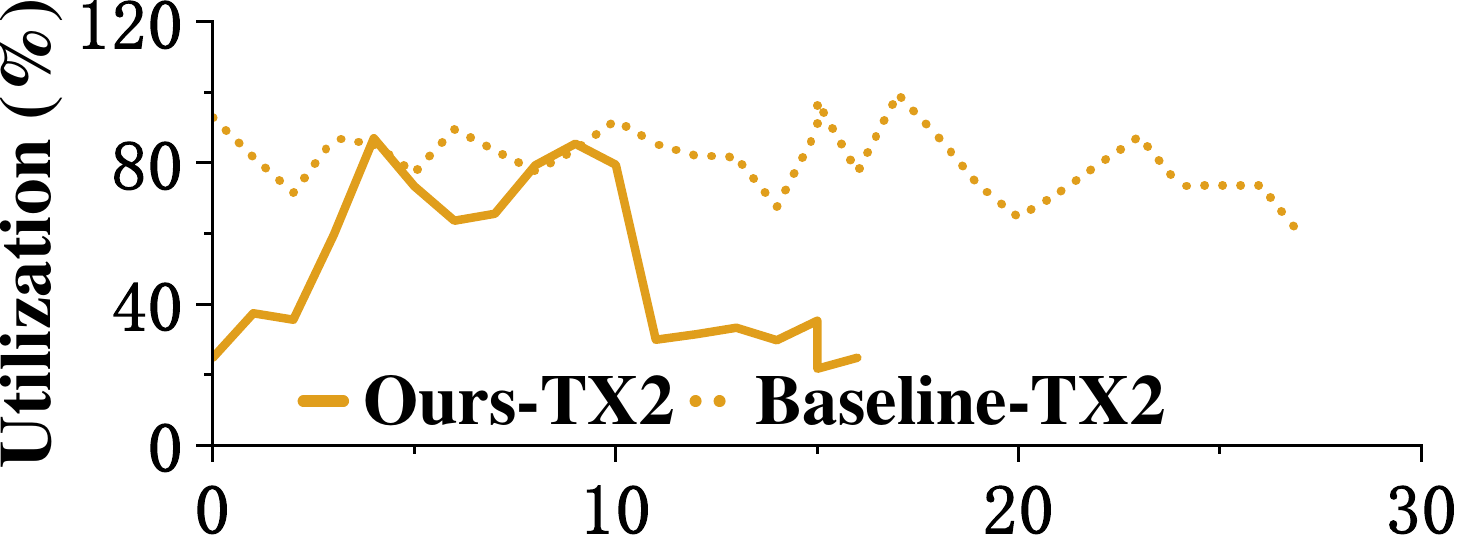}
            \label{tx2-c-s3}
        }
        \hspace{-4mm}
        \subfigure[TX2 in S4]{
            \includegraphics[width=.193\linewidth]{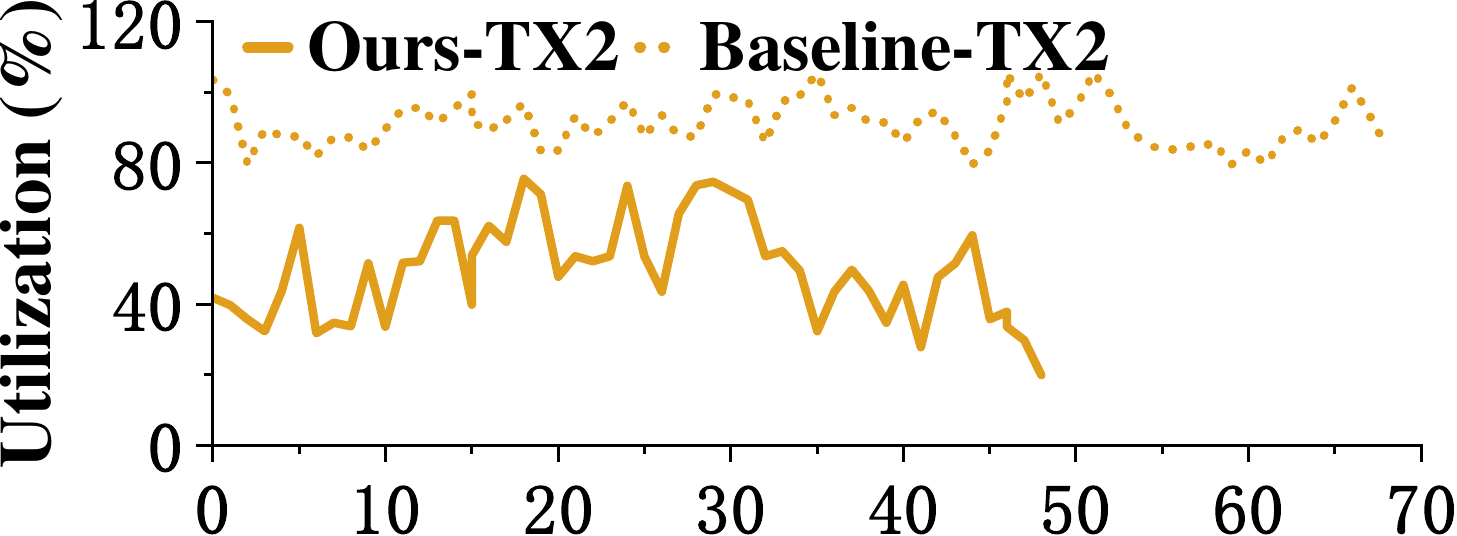}
            \label{tx2-c-s4}
        }
        \hspace{-4mm}
        \subfigure[TX2 in S5]{
            \includegraphics[width=.193\linewidth]{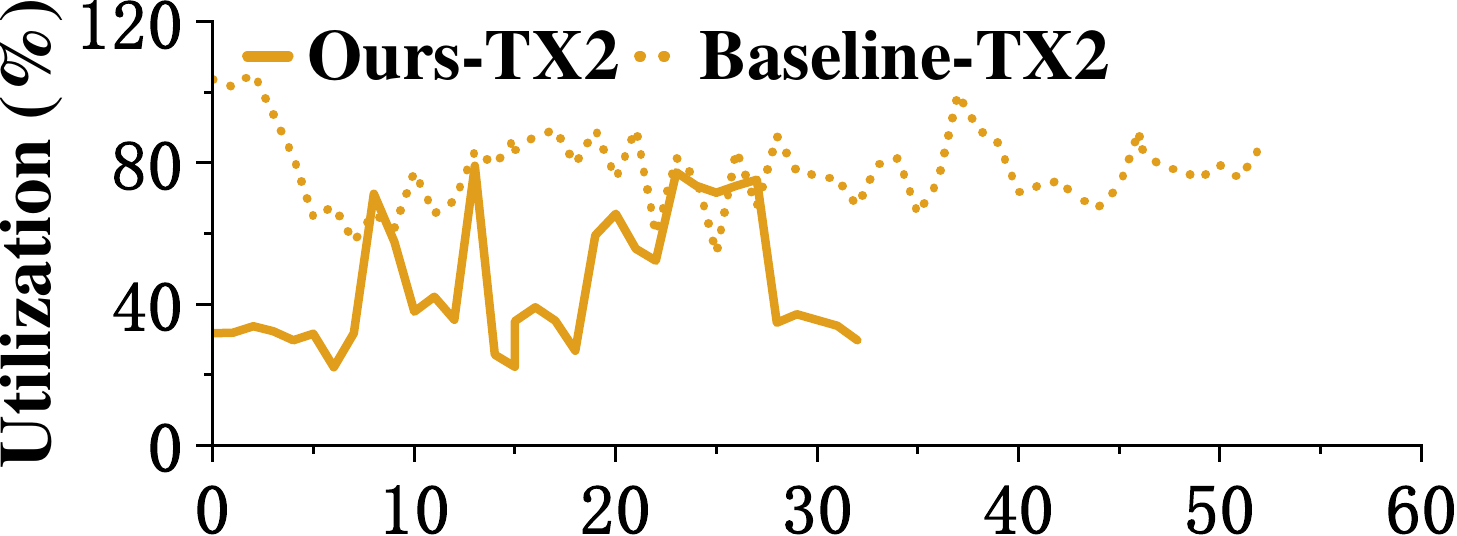}
            \label{tx2-c-s5}
        }
        \subfigure[Pi4B in S1]{
            \includegraphics[width=.193\linewidth]{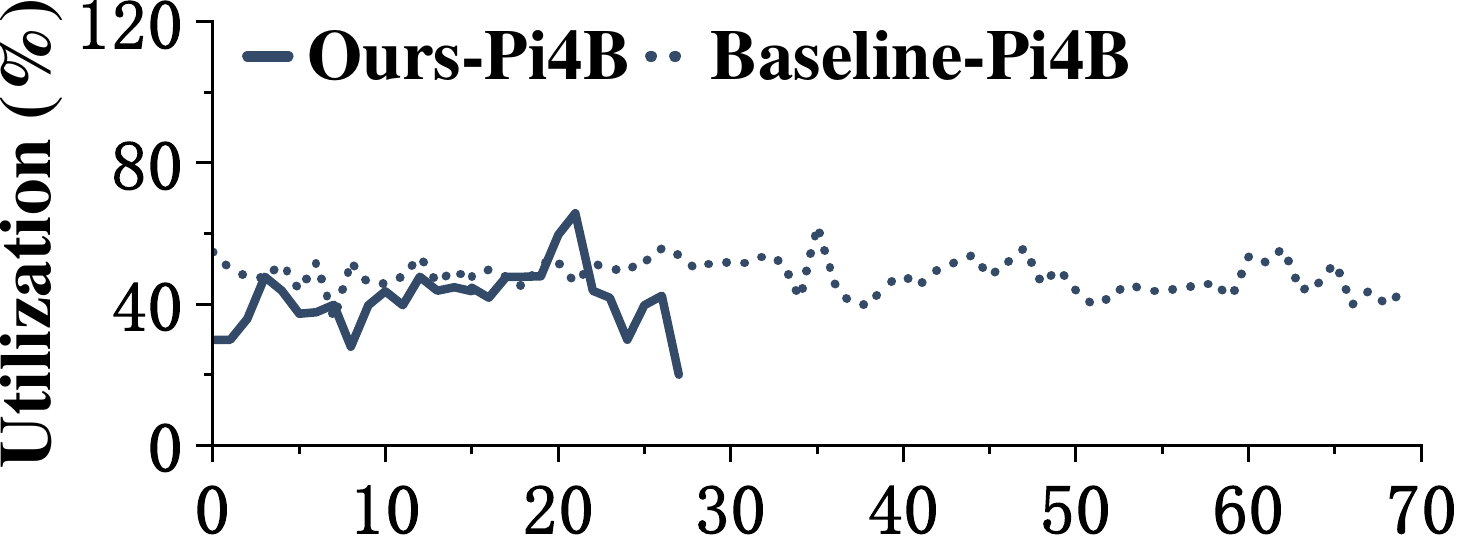}
            \label{pi4b-c-s1}
        }
        \hspace{-4mm}
        \subfigure[Pi4B in S2]{
            \includegraphics[width=.193\linewidth]{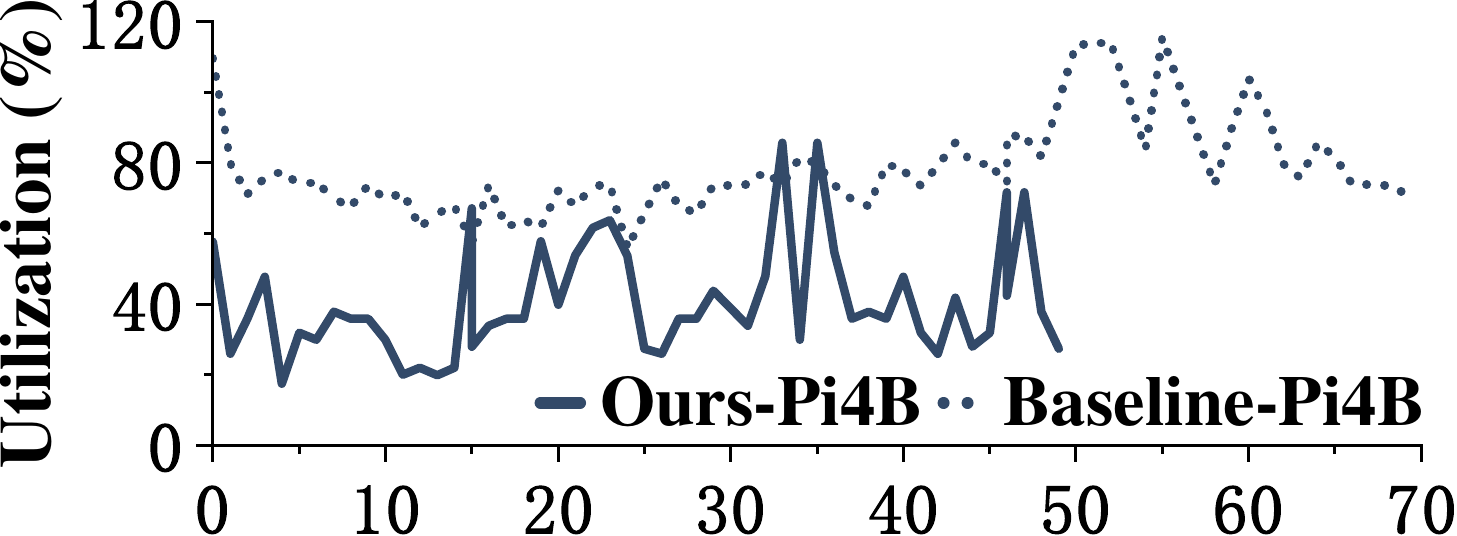}
            \label{pi4b-c-s2}
        }
        \hspace{-4mm}
        \subfigure[Pi4B in S3]{
            \includegraphics[width=.193\linewidth]{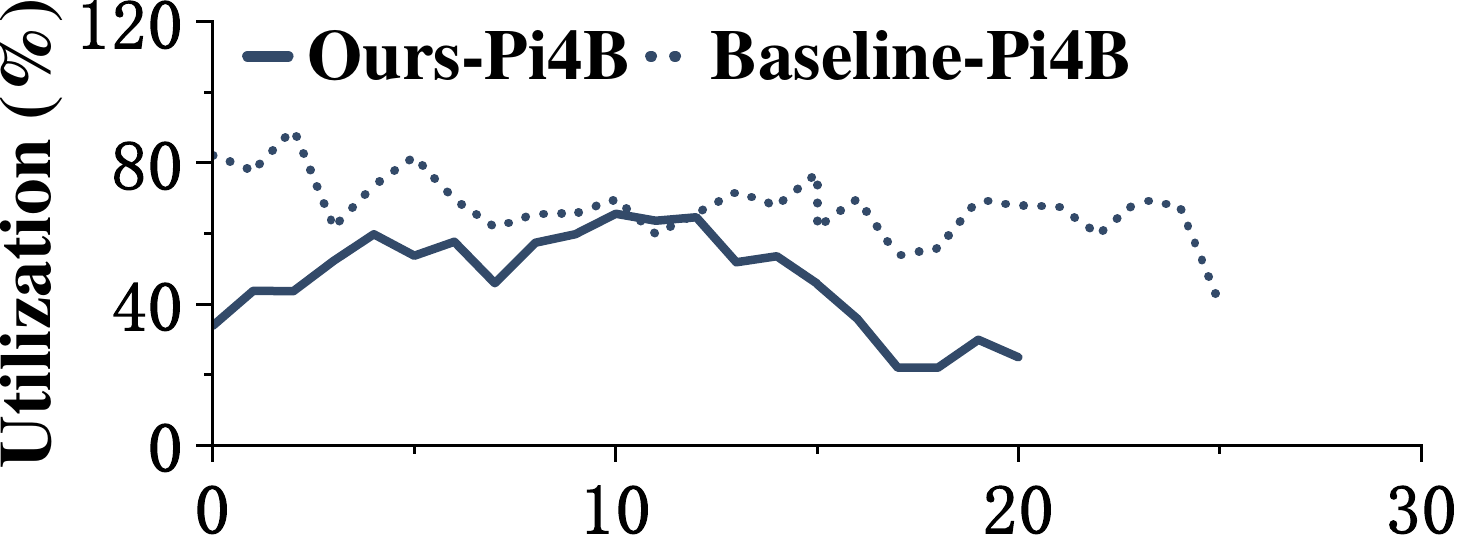}
            \label{pi4b-c-s3}
        }
        \hspace{-4mm}
        \subfigure[Pi4B in S4]{
            \includegraphics[width=.193\linewidth]{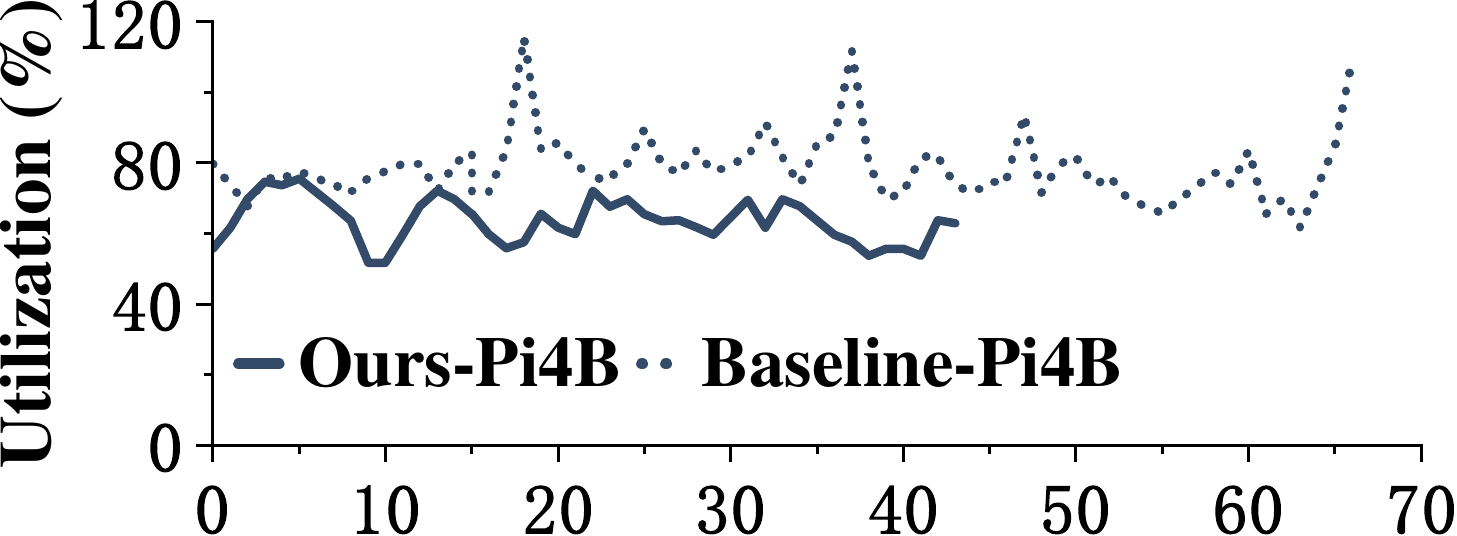}
            \label{pi4b-c-s4}
        }
        \hspace{-4mm}
        \subfigure[Pi4B in S5]{
            \includegraphics[width=.193\linewidth]{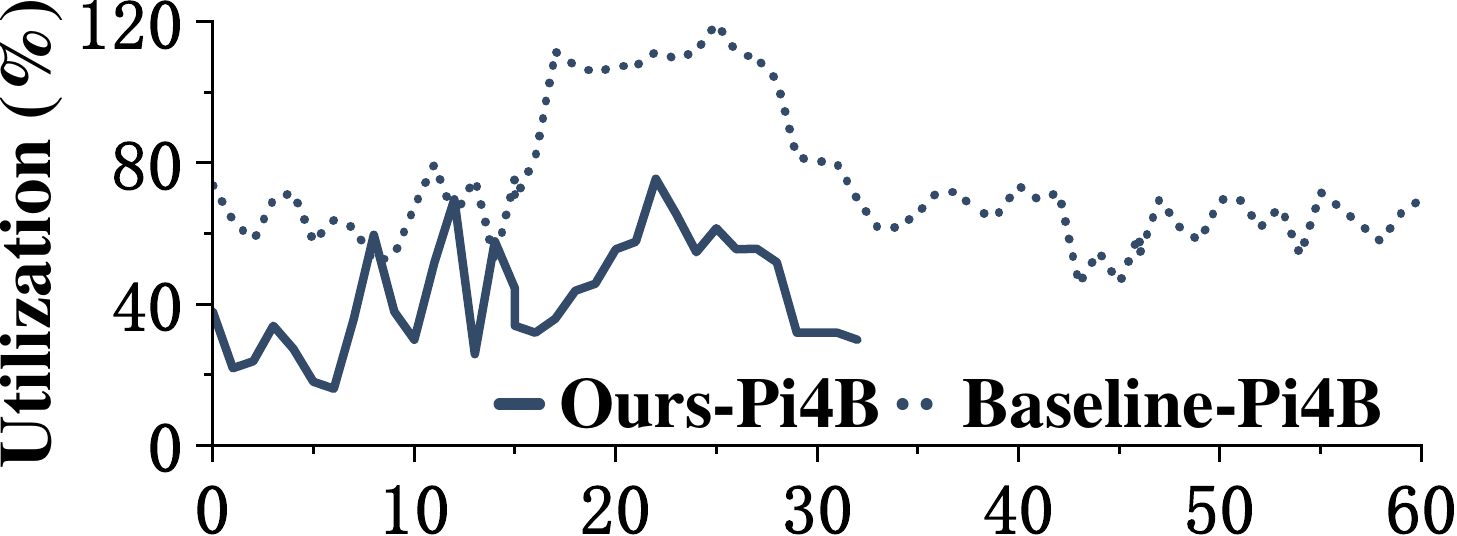}
            \label{pi4b-c-s5}
        }
    }
    \setlength{\abovecaptionskip}{-2pt}
    \setlength{\belowcaptionskip}{-5pt}
    \caption{Utilization of the navigation tasks on varying computing devices in different scenarios. (x-axis: time (s))}
    \vspace{-0.2em}
    \label{exp-hil-process}
\end{figure*}
\subsubsection{Overall Results}
\begin{figure*}[!htbp]
    \centering{
        \subfigure[Pi5 in S1]{
            \includegraphics[width=.193\linewidth]{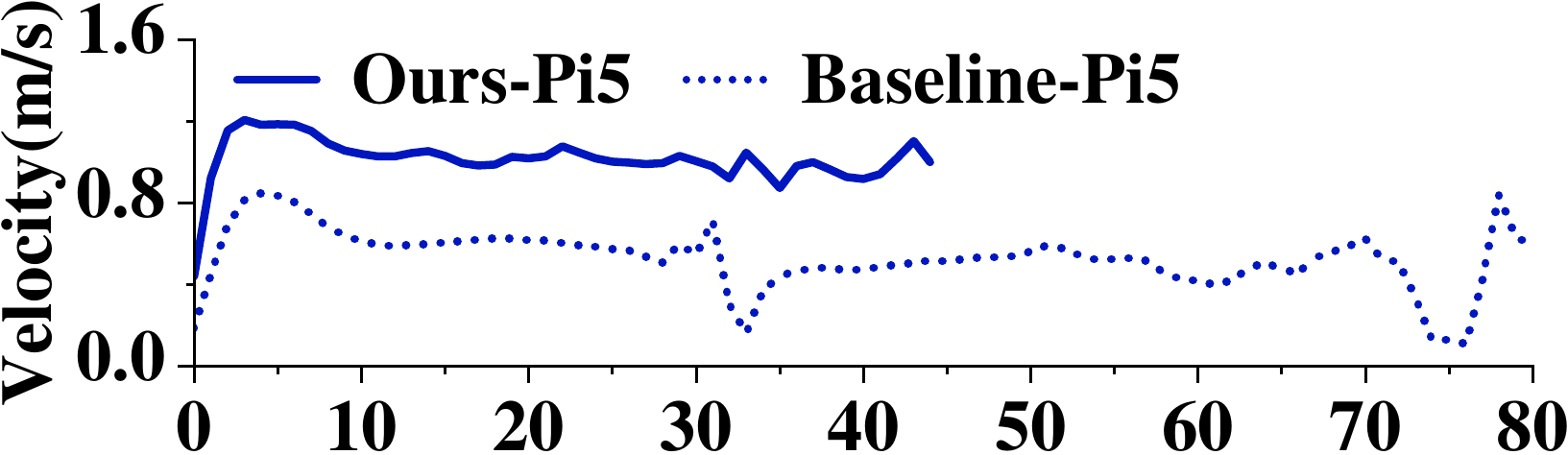}
            \label{pi5-v-s1}
        }
        \hspace{-4mm}
        \subfigure[Pi5 in S2]{
            \includegraphics[width=.193\linewidth]{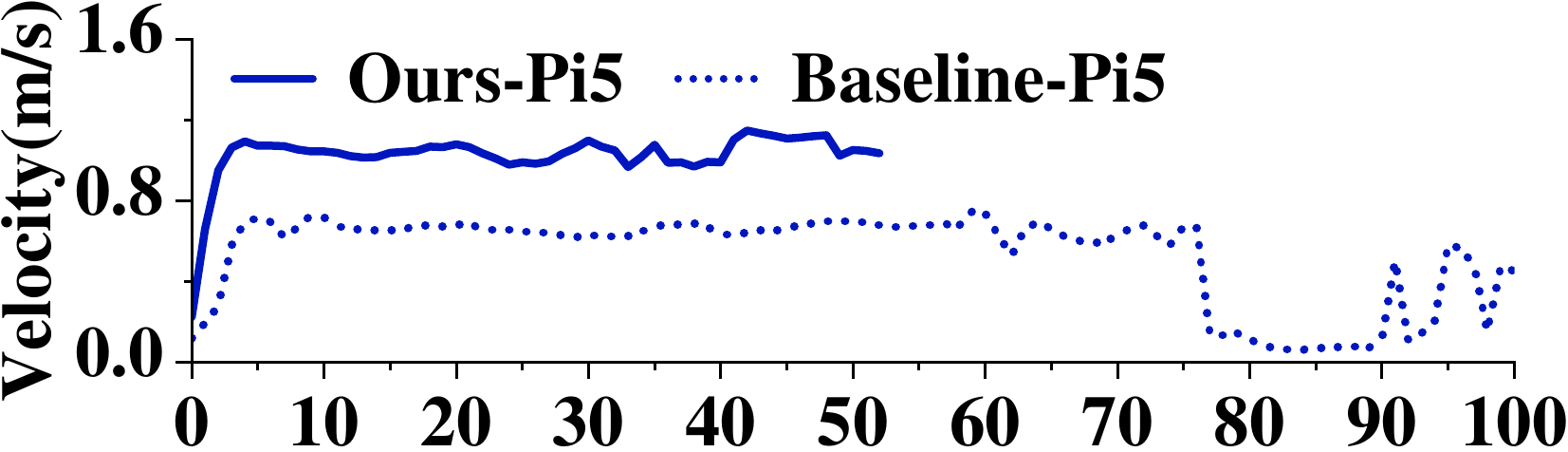}
            \label{pi5-v-s2}
        }
        \hspace{-4mm}
        \subfigure[Pi5 in S3]{
            \includegraphics[width=.193\linewidth]{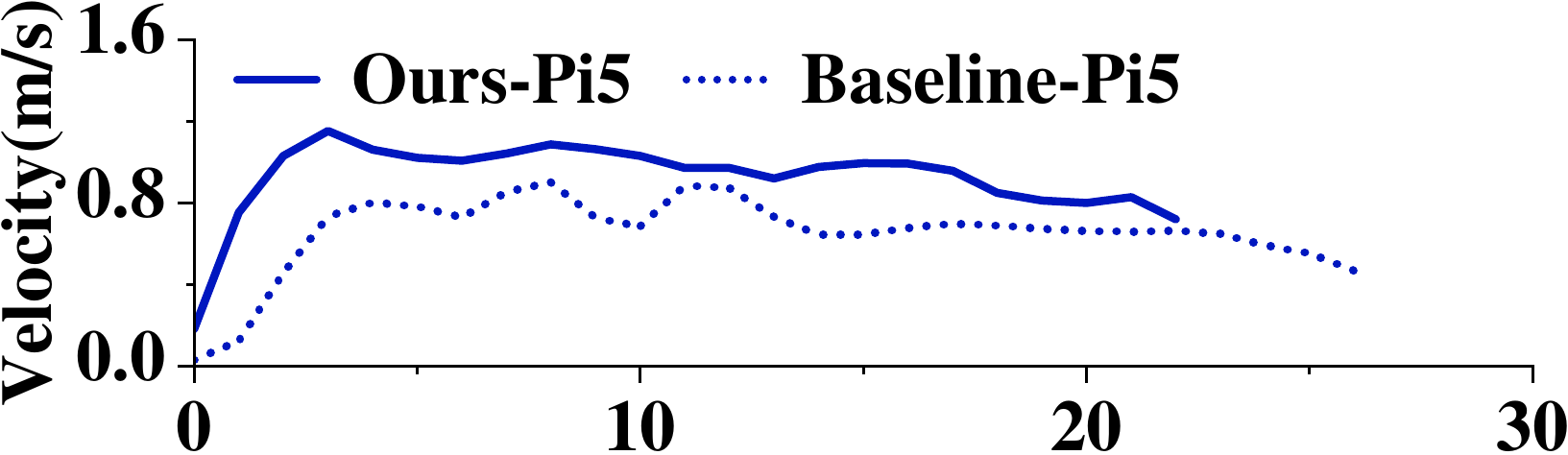}
            \label{pi5-v-s3}
        }
        \hspace{-4mm}
        \subfigure[Pi5 in S4]{
            \includegraphics[width=.193\linewidth]{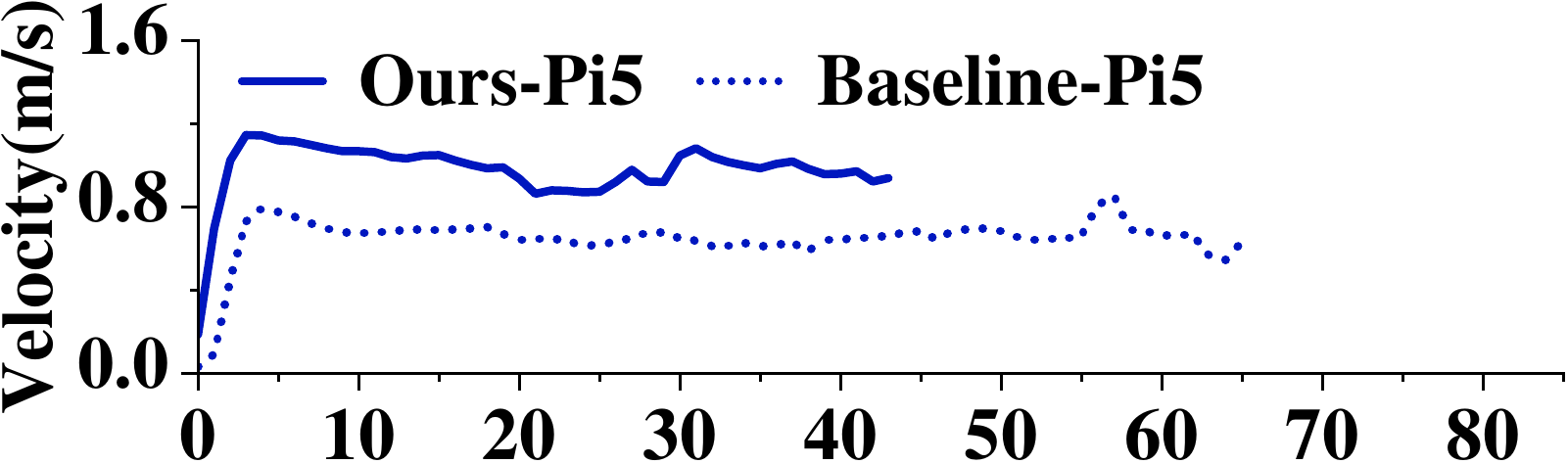}
            \label{pi5-v-s4}
        }
        \hspace{-4mm}
        \subfigure[Pi5 in S5]{
            \includegraphics[width=.193\linewidth]{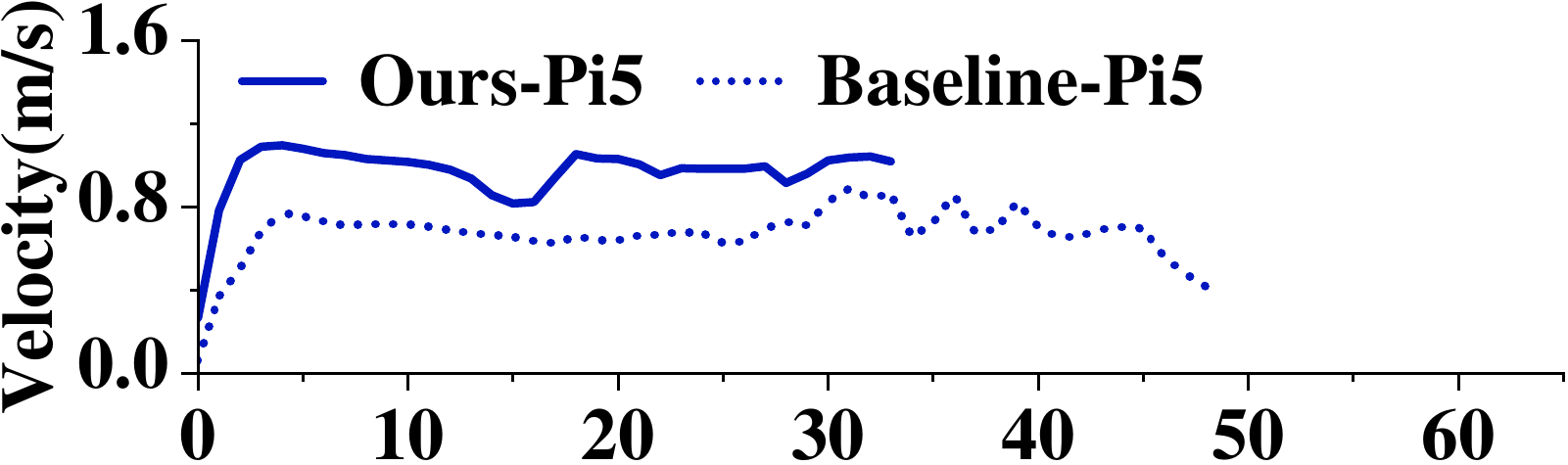}
            \label{pi5-v-s5}
        }
                \subfigure[NX in S1]{
        	\includegraphics[width=.193\linewidth]{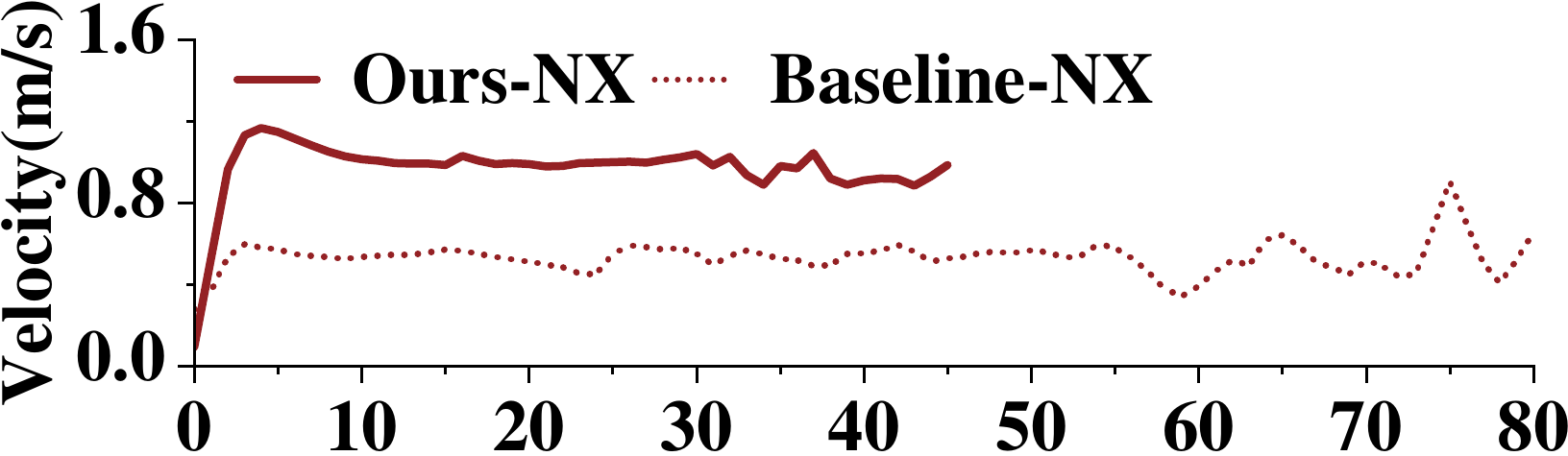}
        	\label{nx-v-s1}
        }
        \hspace{-4mm}
        \subfigure[NX in S2]{
        	\includegraphics[width=.193\linewidth]{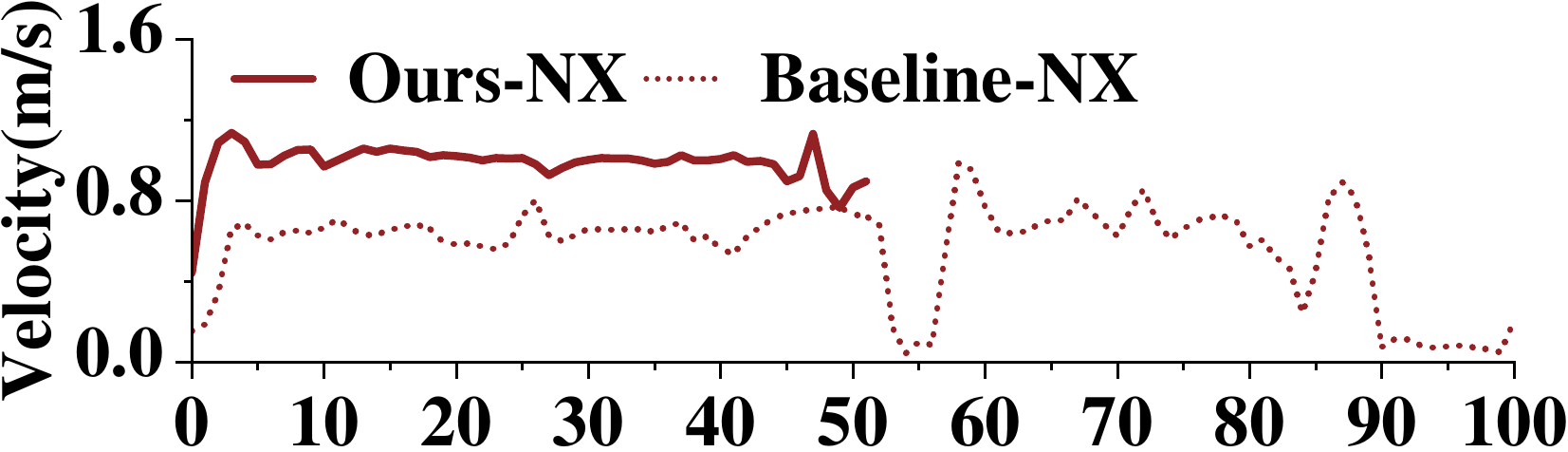}
        	\label{nx-v-s2}
        }
        \hspace{-4mm}
        \subfigure[NX in S3]{
        	\includegraphics[width=.193\linewidth]{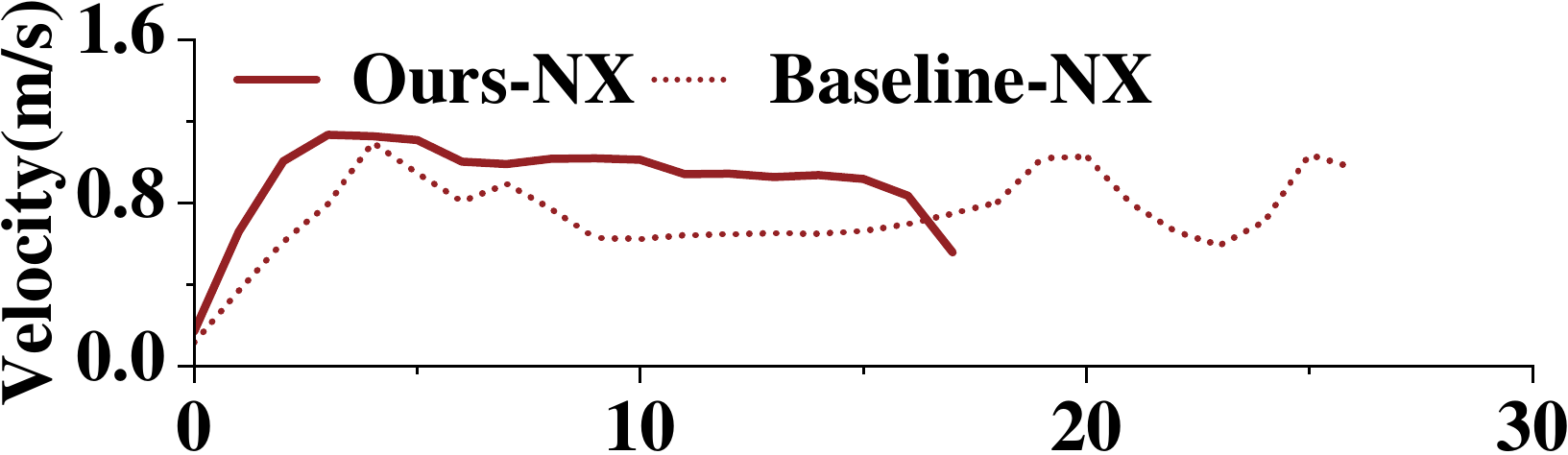}
        	\label{nx-v-s3}
        }
        \hspace{-4mm}
        \subfigure[NX in S4]{
        	\includegraphics[width=.193\linewidth]{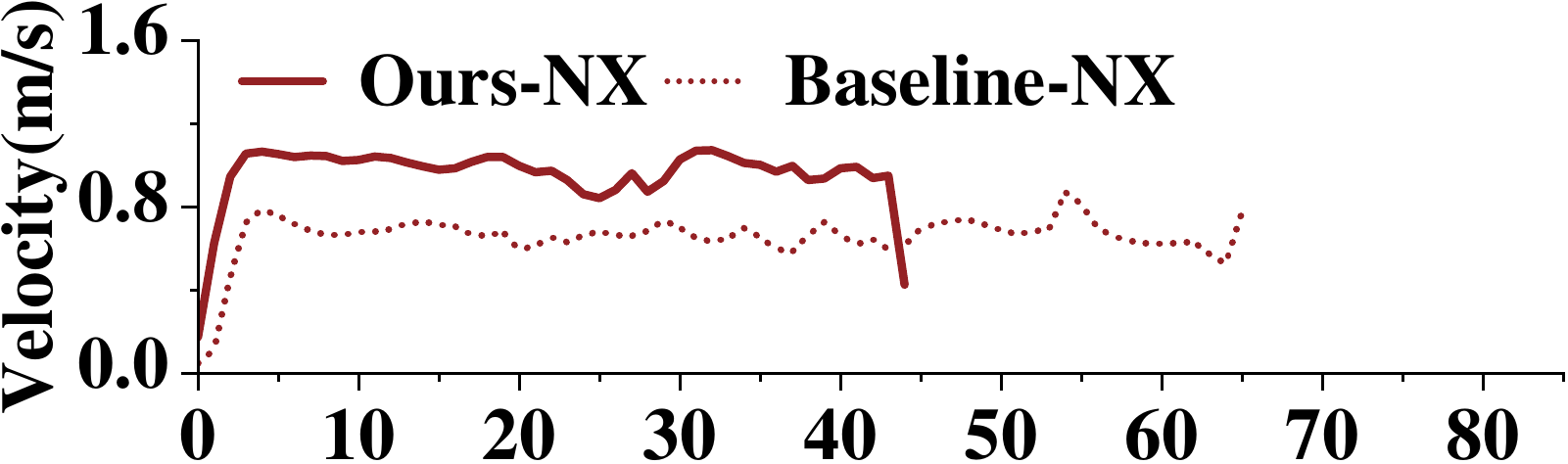}
        	\label{nx-v-s4}
        }
        \hspace{-4mm}
        \subfigure[NX in S5]{
        	\includegraphics[width=.193\linewidth]{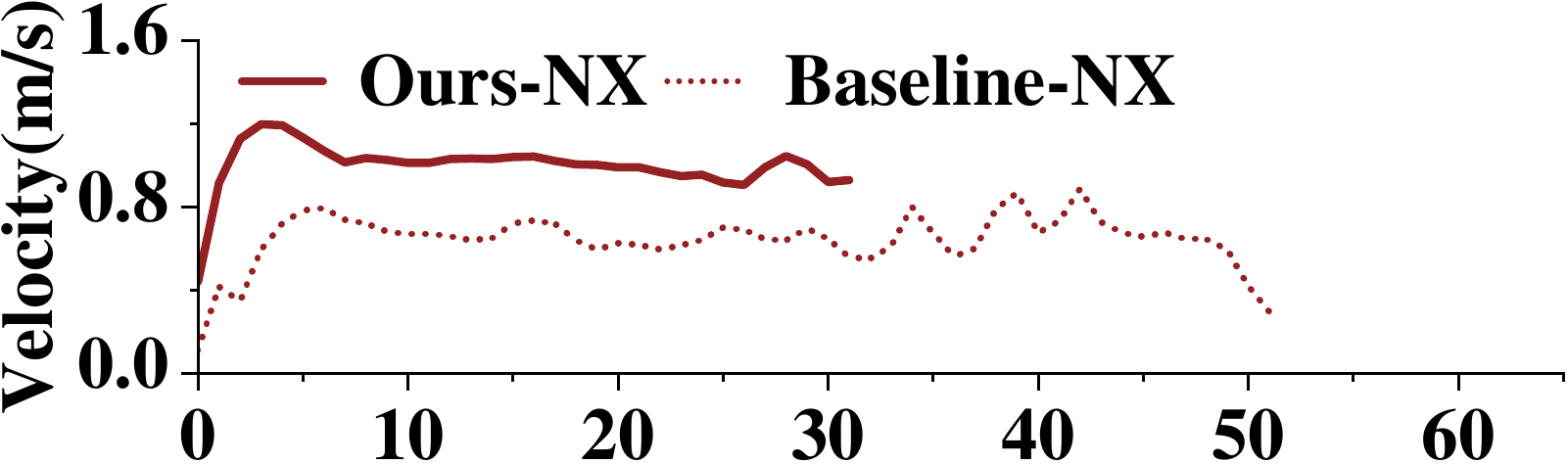}
        	\label{nx-v-s5}
        }
                \subfigure[x86 in S1]{
        	\includegraphics[width=.193\linewidth]{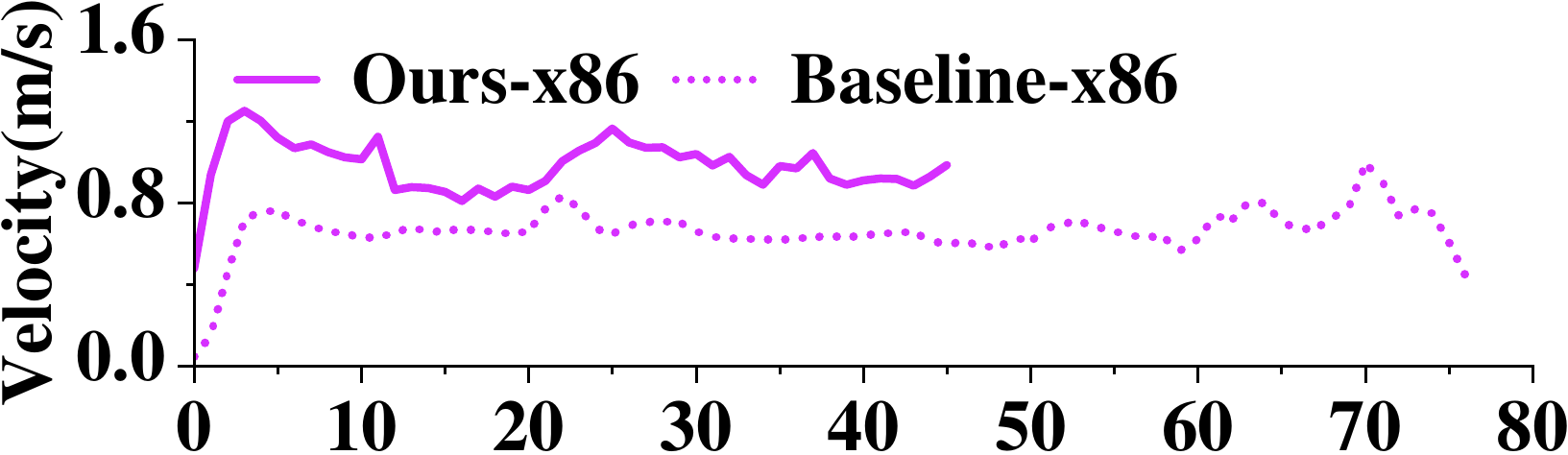}
        	\label{x86-v-s1}
        }
        \hspace{-4mm}
        \subfigure[x86 in S2]{
        	\includegraphics[width=.193\linewidth]{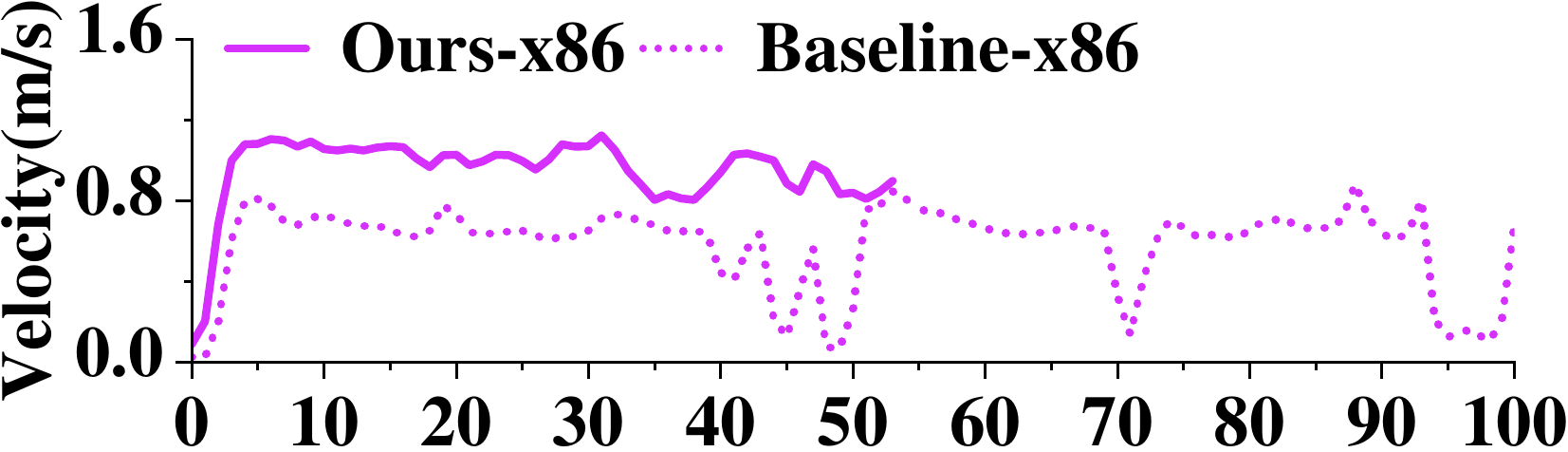}
        	\label{x86-v-s2}
        }
        \hspace{-4mm}
        \subfigure[x86 in S3]{
        	\includegraphics[width=.193\linewidth]{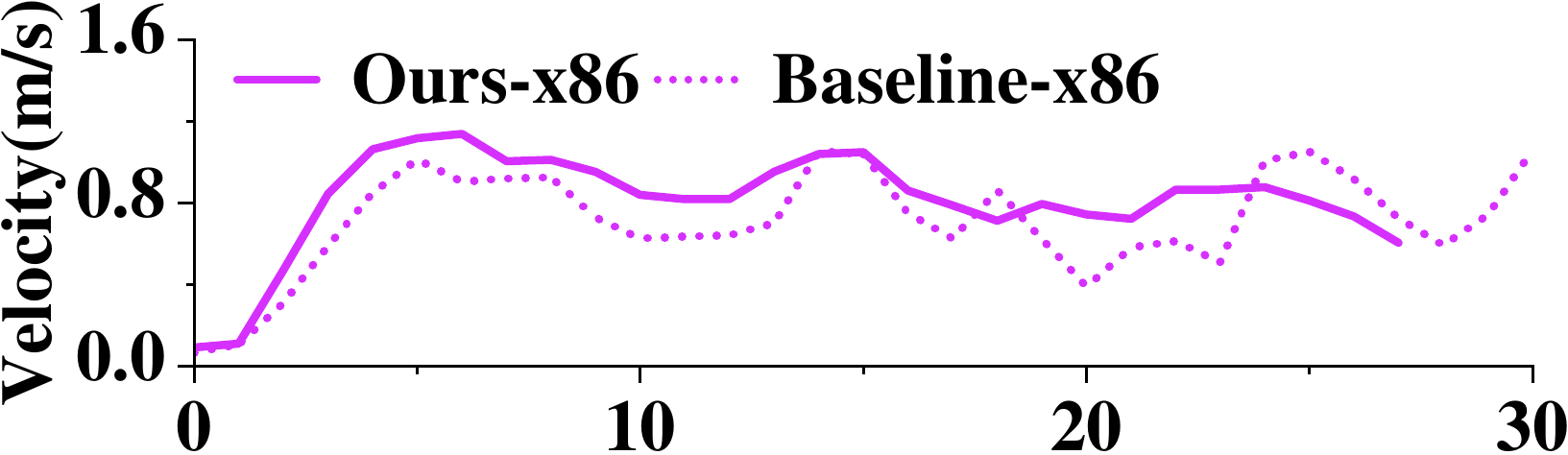}
        	\label{x86-v-s3}
        }
        \hspace{-4mm}
        \subfigure[x86 in S4]{
        	\includegraphics[width=.193\linewidth]{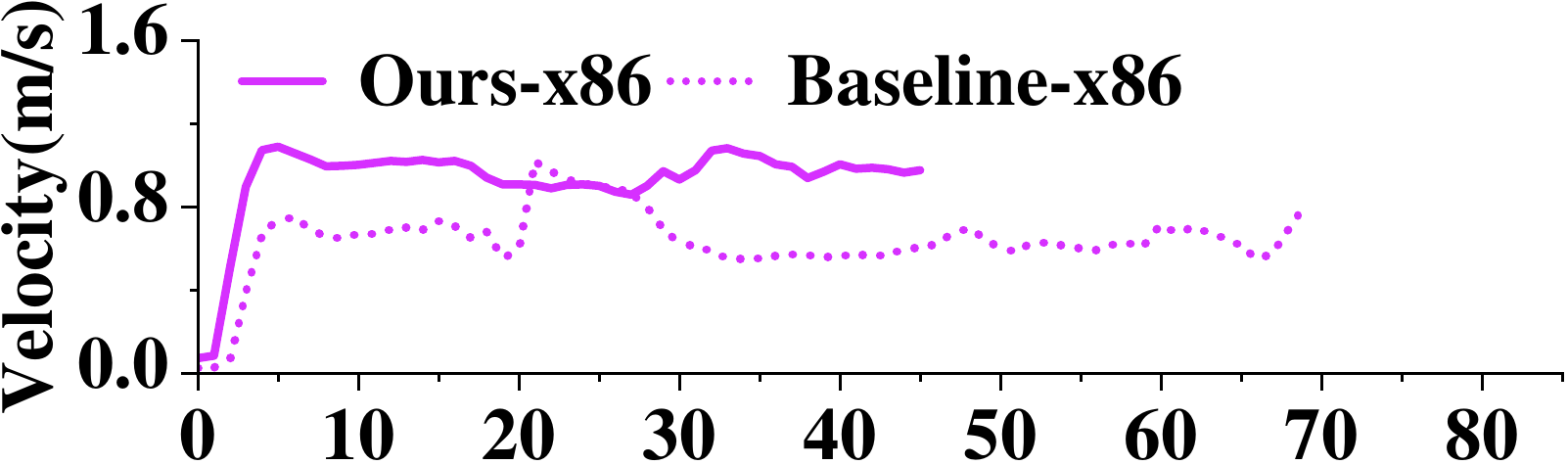}
        	\label{x86-v-s4}
        }
        \hspace{-4mm}
        \subfigure[x86 in S5]{
        	\includegraphics[width=.193\linewidth]{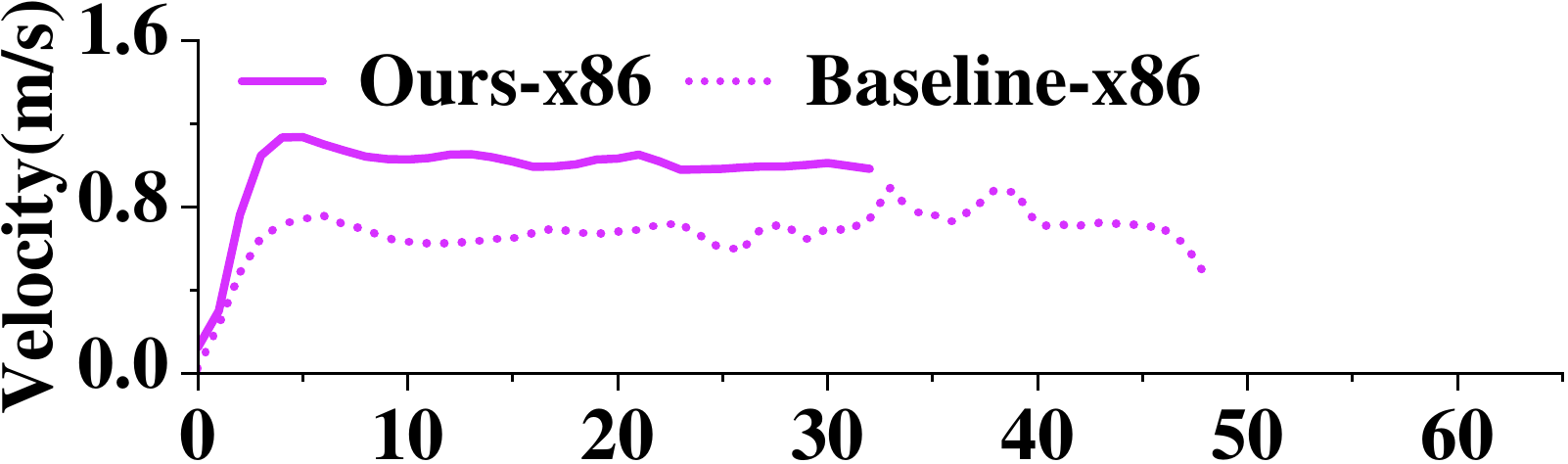}
        	\label{x86-v-s5}
        }
        \subfigure[TX2 in S1]{
            \includegraphics[width=.193\linewidth]{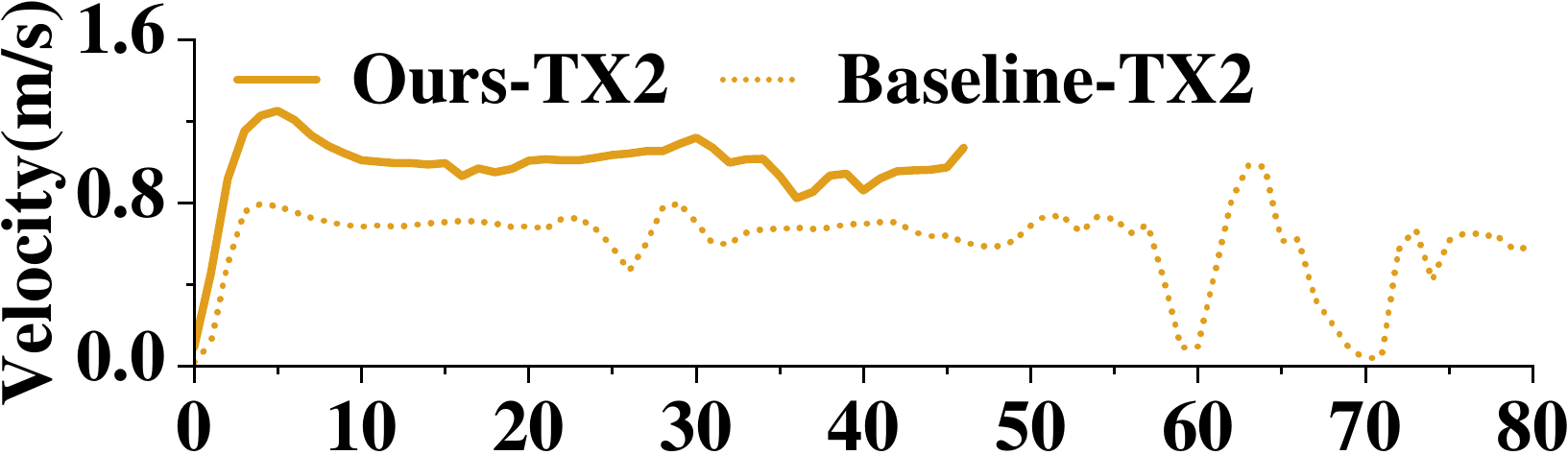}
            \label{tx2-v-s1}
        }
        \hspace{-4mm}
        \subfigure[TX2 in S2]{
            \includegraphics[width=.193\linewidth]{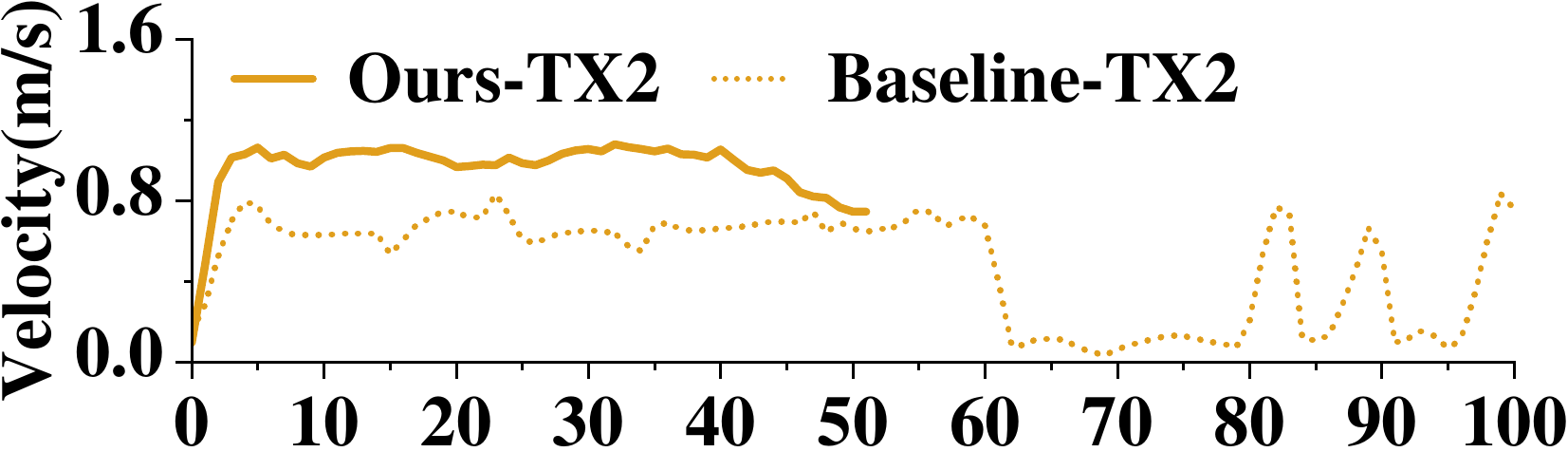}
            \label{tx2-v-s2}
        }
        \hspace{-4mm}
        \subfigure[TX2 in S3]{
            \includegraphics[width=.193\linewidth]{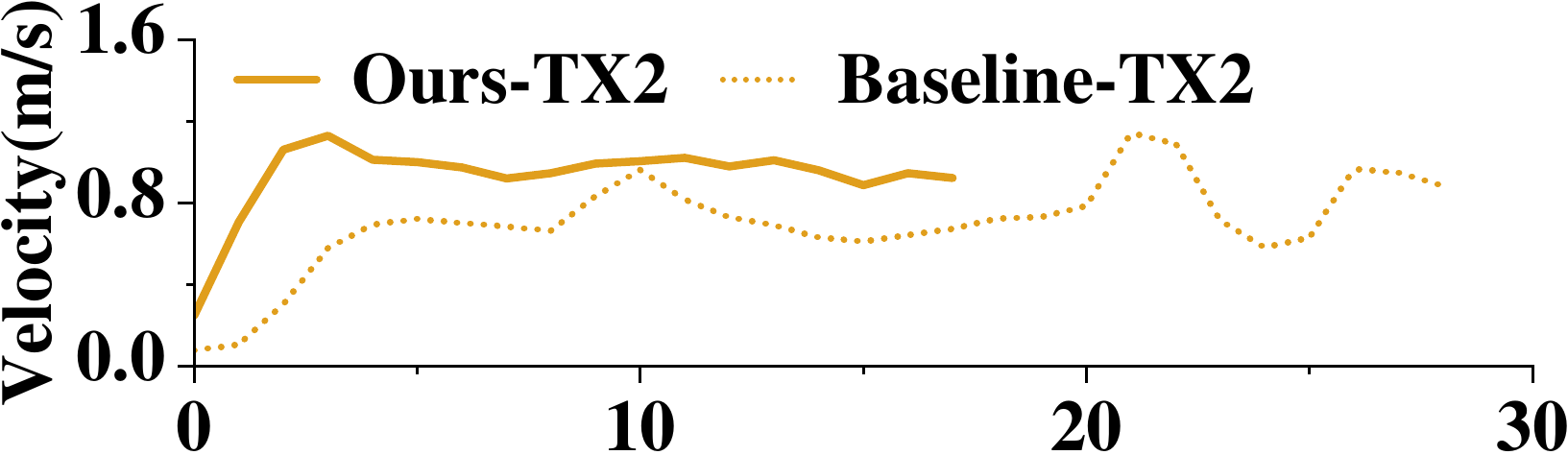}
            \label{tx2-v-s3}
        }
        \hspace{-4mm}
        \subfigure[TX2 in S4]{
            \includegraphics[width=.193\linewidth]{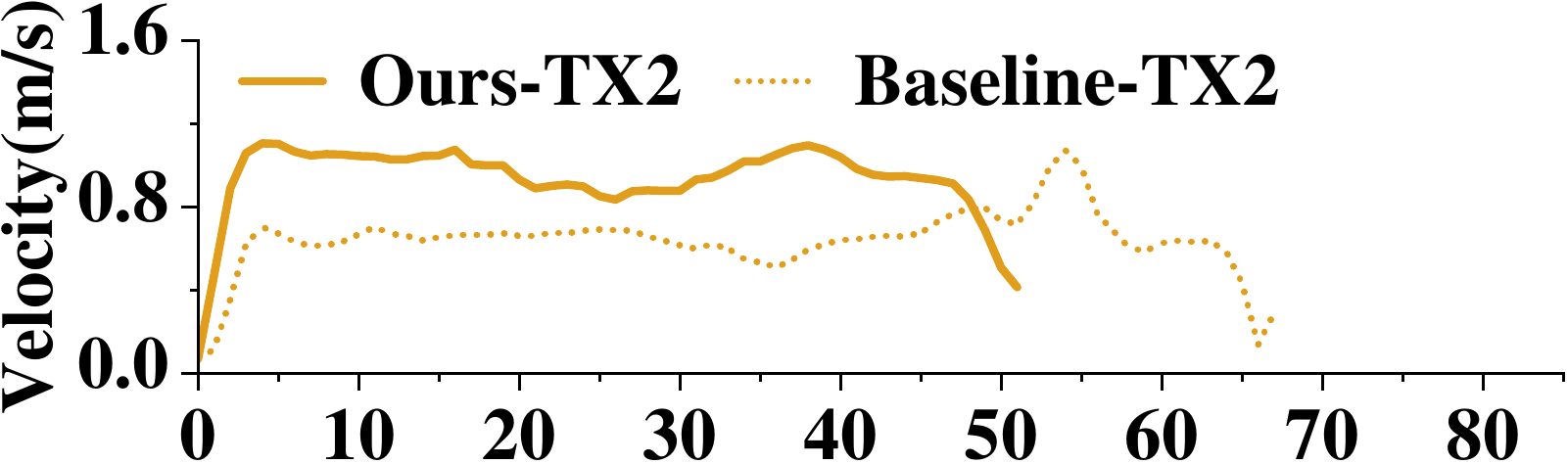}
            \label{tx2-v-s4}
        }
        \hspace{-4mm}
        \subfigure[TX2 in S5]{
            \includegraphics[width=.193\linewidth]{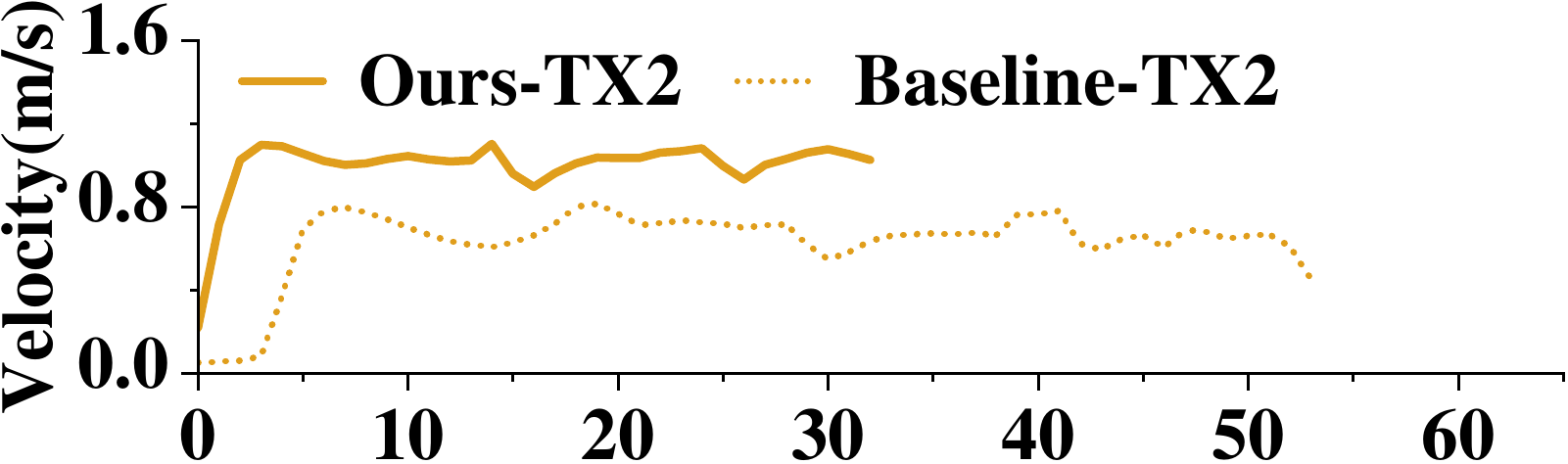}
            \label{tx2-v-s5}
        }
        \subfigure[Pi4B in S1]{
            \includegraphics[width=.193\linewidth]{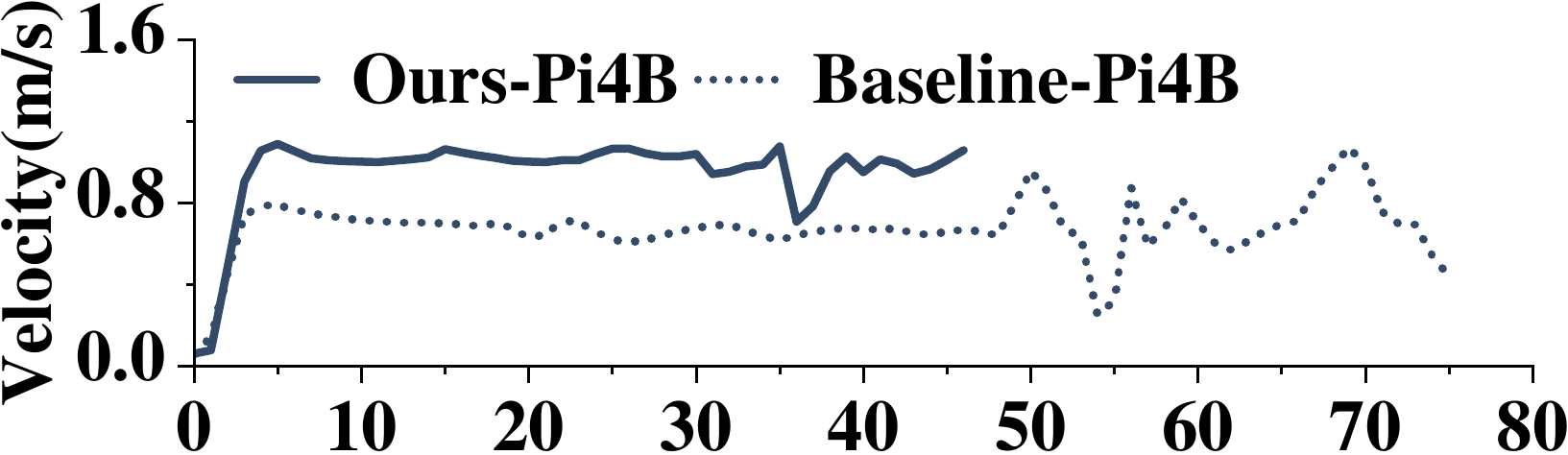}
            \label{pi4b-v-s1}
        }
        \hspace{-4mm}
        \subfigure[Pi4B in S2]{
            \includegraphics[width=.193\linewidth]{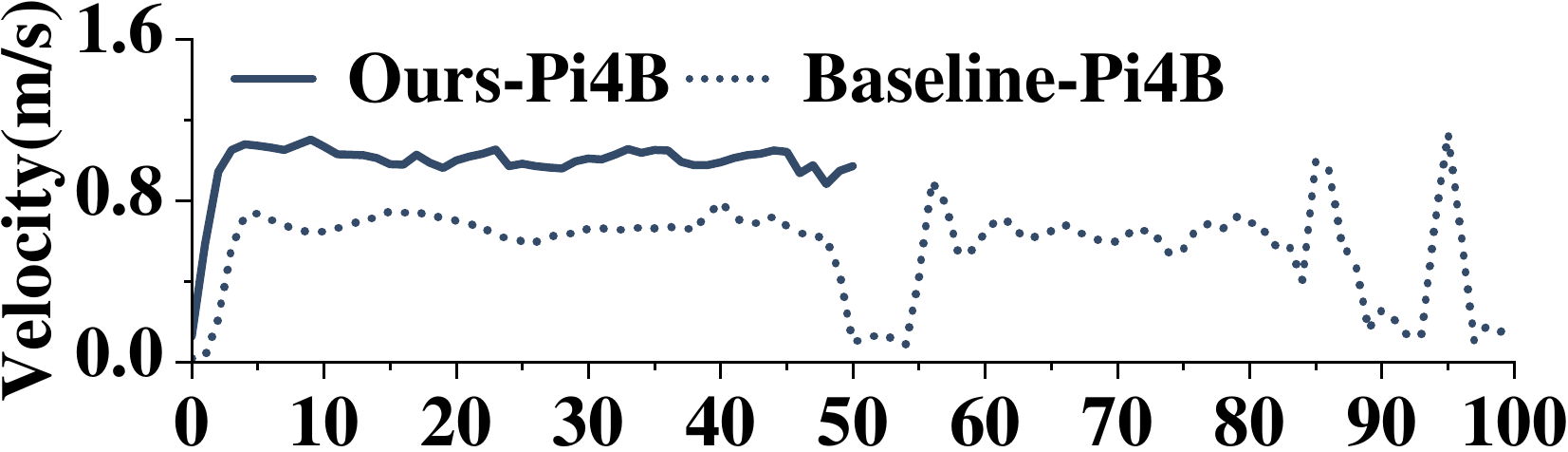}
            \label{pi4b-v-s2}
        }
        \hspace{-4mm}
        \subfigure[Pi4B in S3]{
            \includegraphics[width=.193\linewidth]{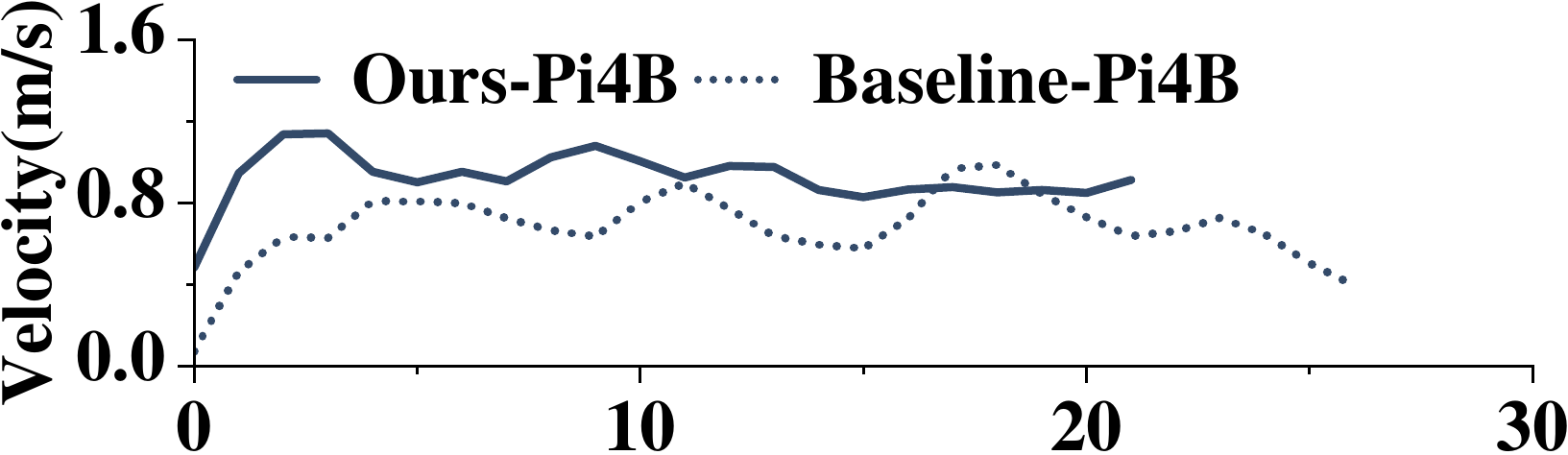}
            \label{pi4b-v-s3}
        }
        \hspace{-4mm}
        \subfigure[Pi4B in S4]{
            \includegraphics[width=.193\linewidth]{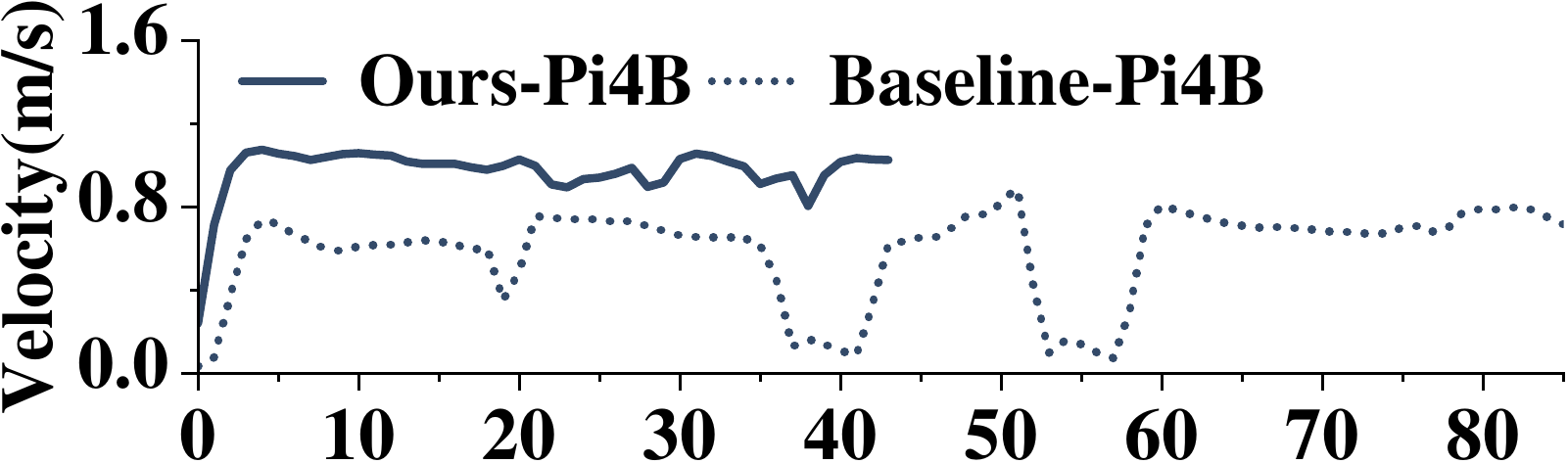}
            \label{pi4b-v-s4}
        }
        \hspace{-4mm}
        \subfigure[Pi4B in S5]{
            \includegraphics[width=.193\linewidth]{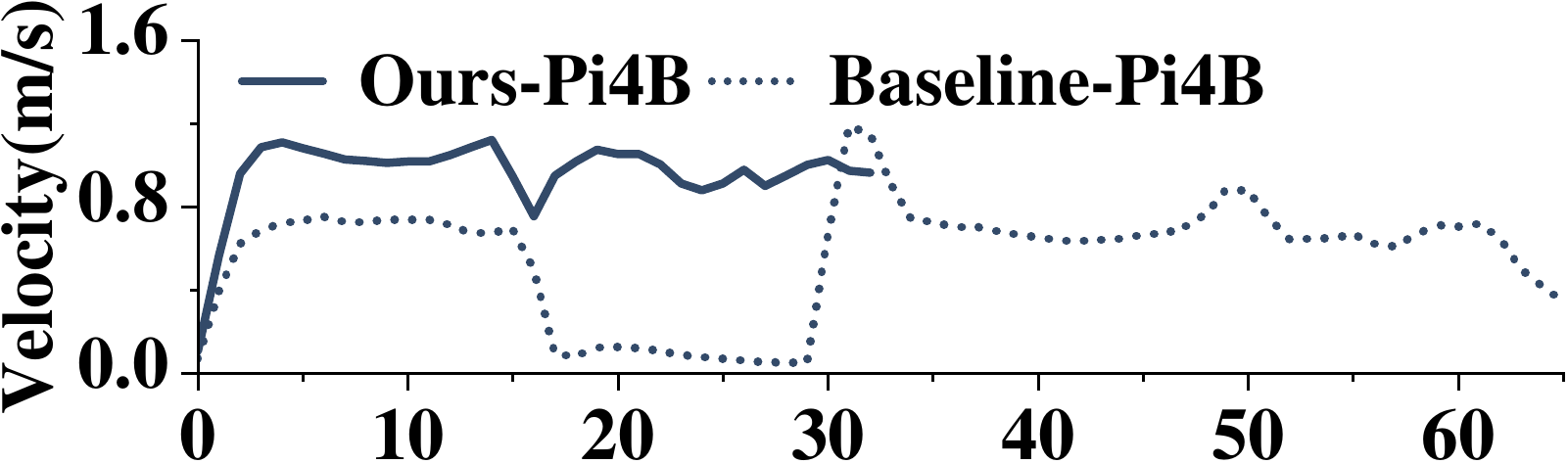}
            \label{pi4b-v-s5}
        }
    }
    \setlength{\abovecaptionskip}{-2pt}
    \setlength{\belowcaptionskip}{-5pt}
    \caption{Velocity on varying computing devices in different scenarios. (x-axis: time (s))}
    \label{exp-hil-velocity}
\end{figure*}

\begin{figure*}[!htbp]
    \centering{
        \subfigure[Frequency variation in S1]{
            \includegraphics[width=.193\linewidth]{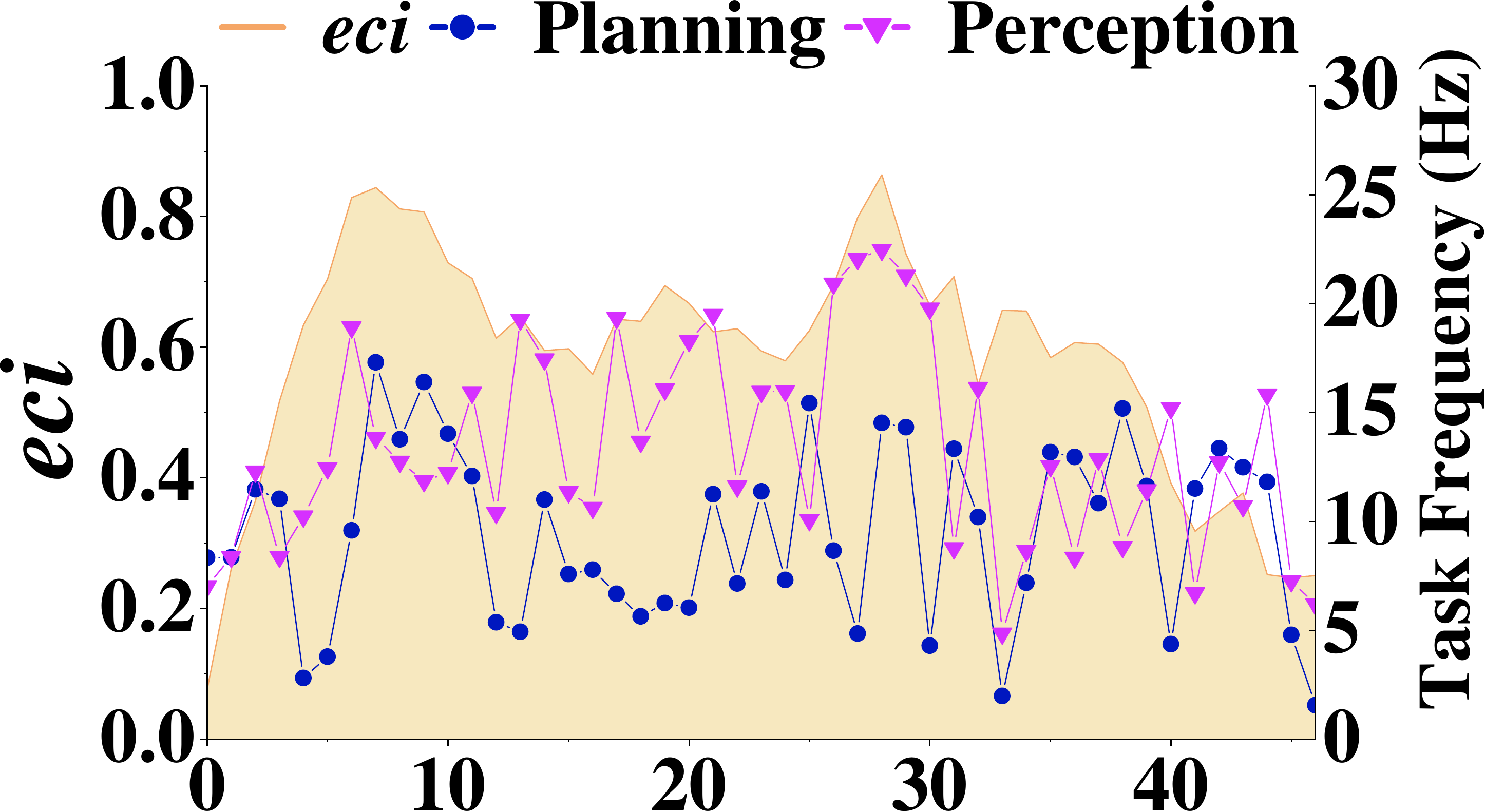}
            \label{pi5-a-s1}
        }
        \hspace{-4mm}
        \subfigure[Frequency variation in S2]{
            \includegraphics[width=.193\linewidth]{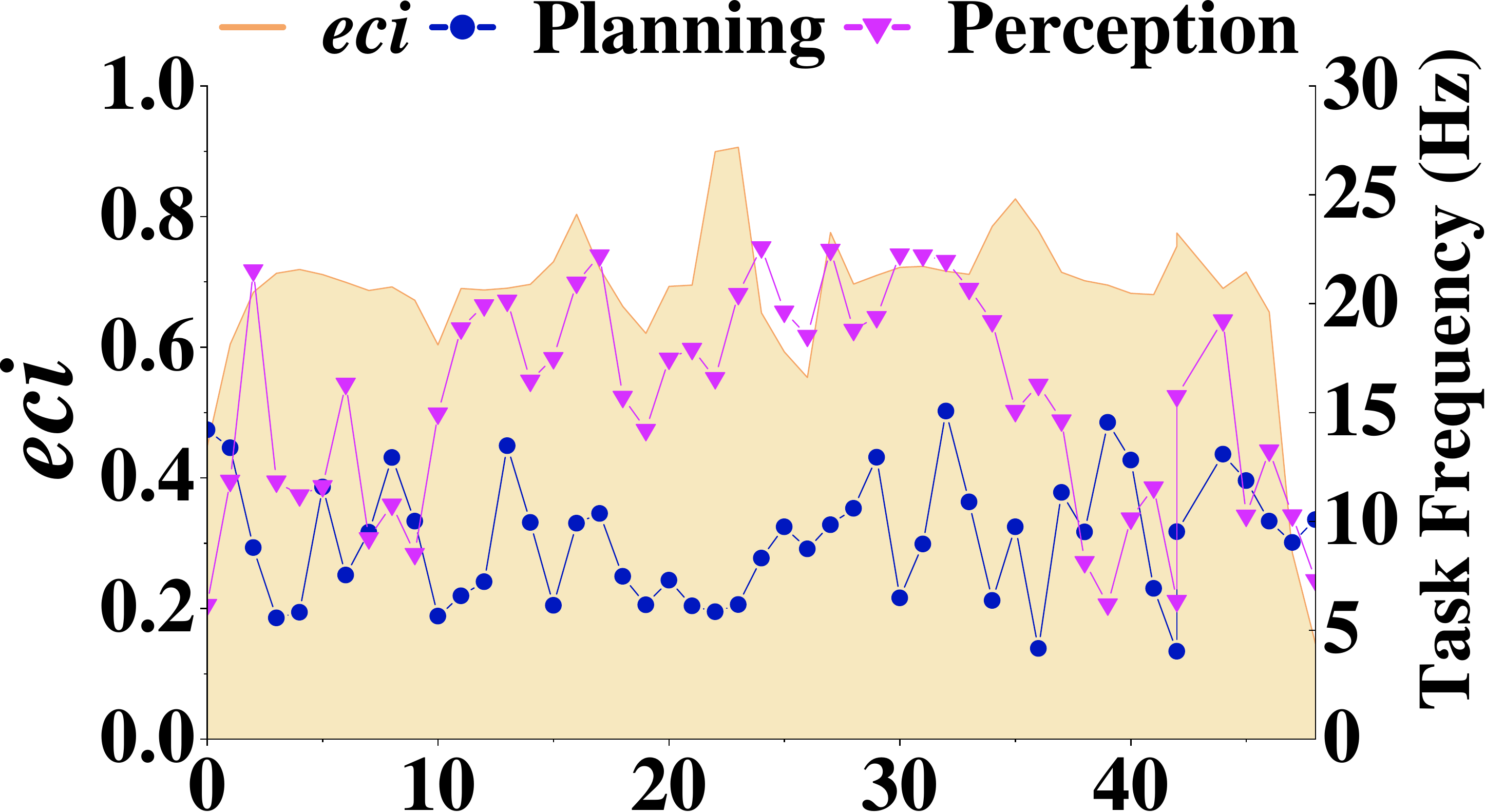}
            \label{pi5-a-s2}
        }
        \hspace{-4mm}
        \subfigure[Frequency variation in S3]{
            \includegraphics[width=.193\linewidth]{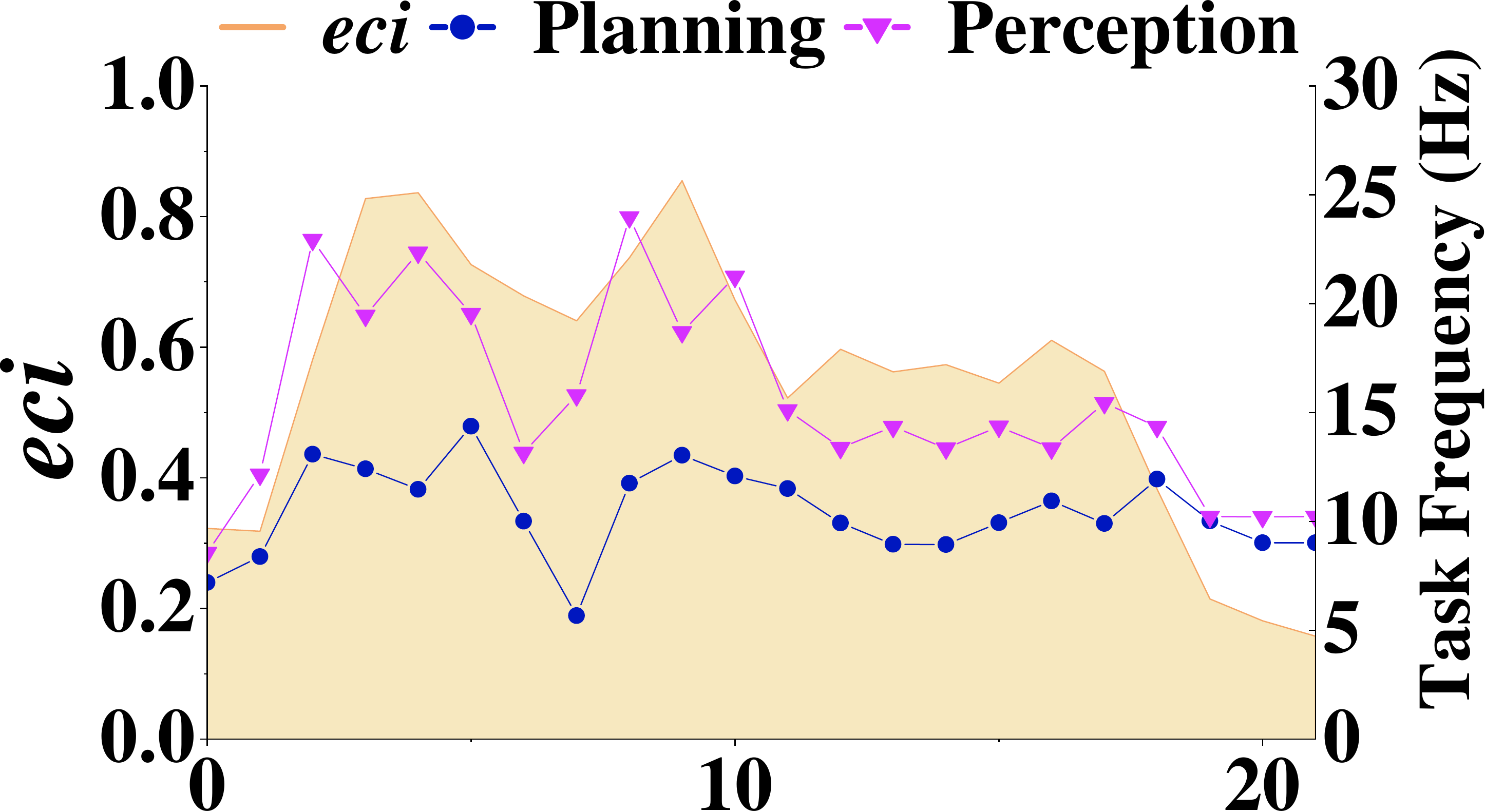}
            \label{pi5-a-s3}
        }
        \hspace{-4mm}
        \subfigure[Frequency variation in S4]{
            \includegraphics[width=.193\linewidth]{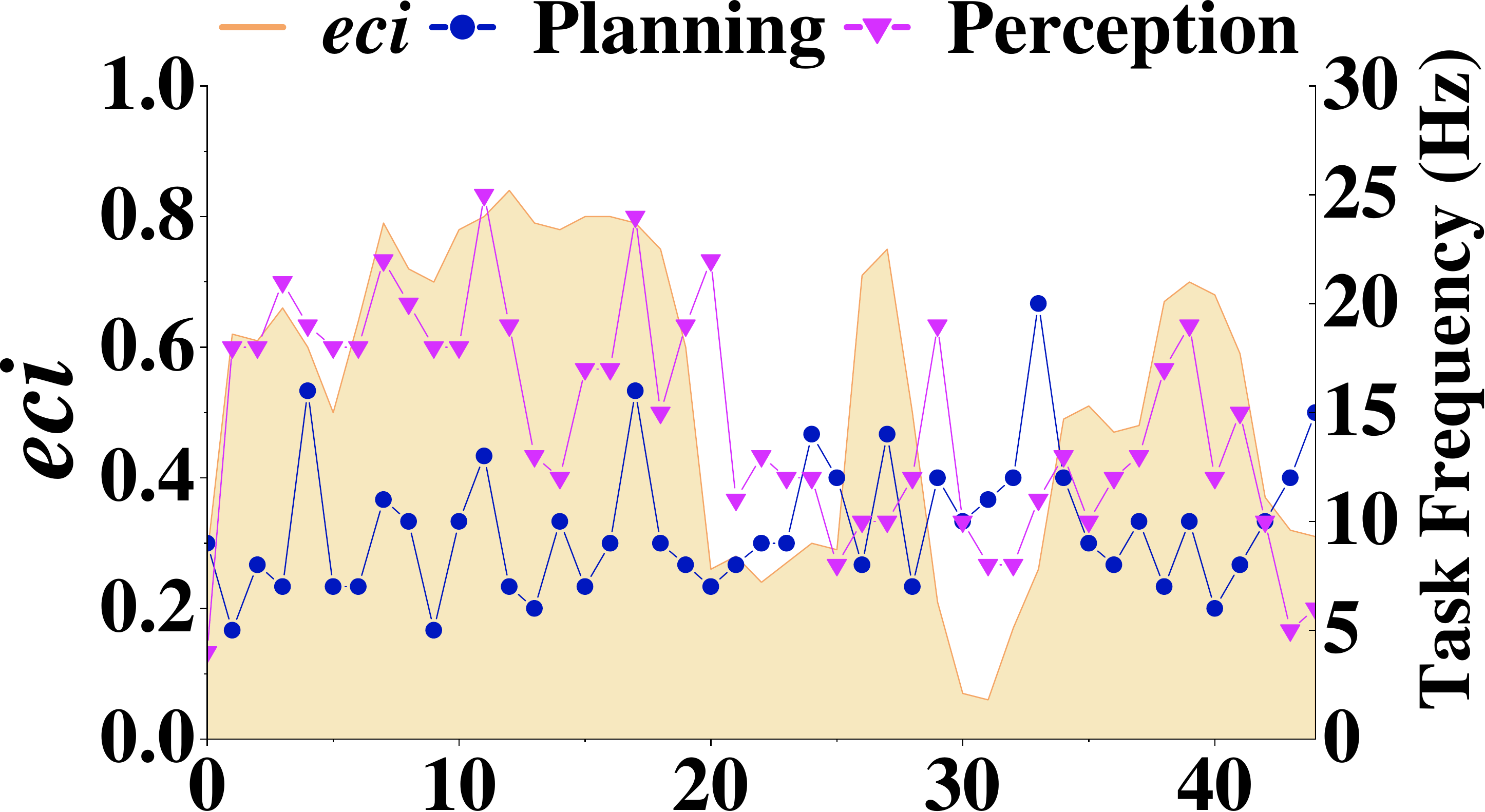}
            \label{pi5-a-s4}
        }
        \hspace{-4mm}
        \subfigure[Frequency variation in S5]{
            \includegraphics[width=.193\linewidth]{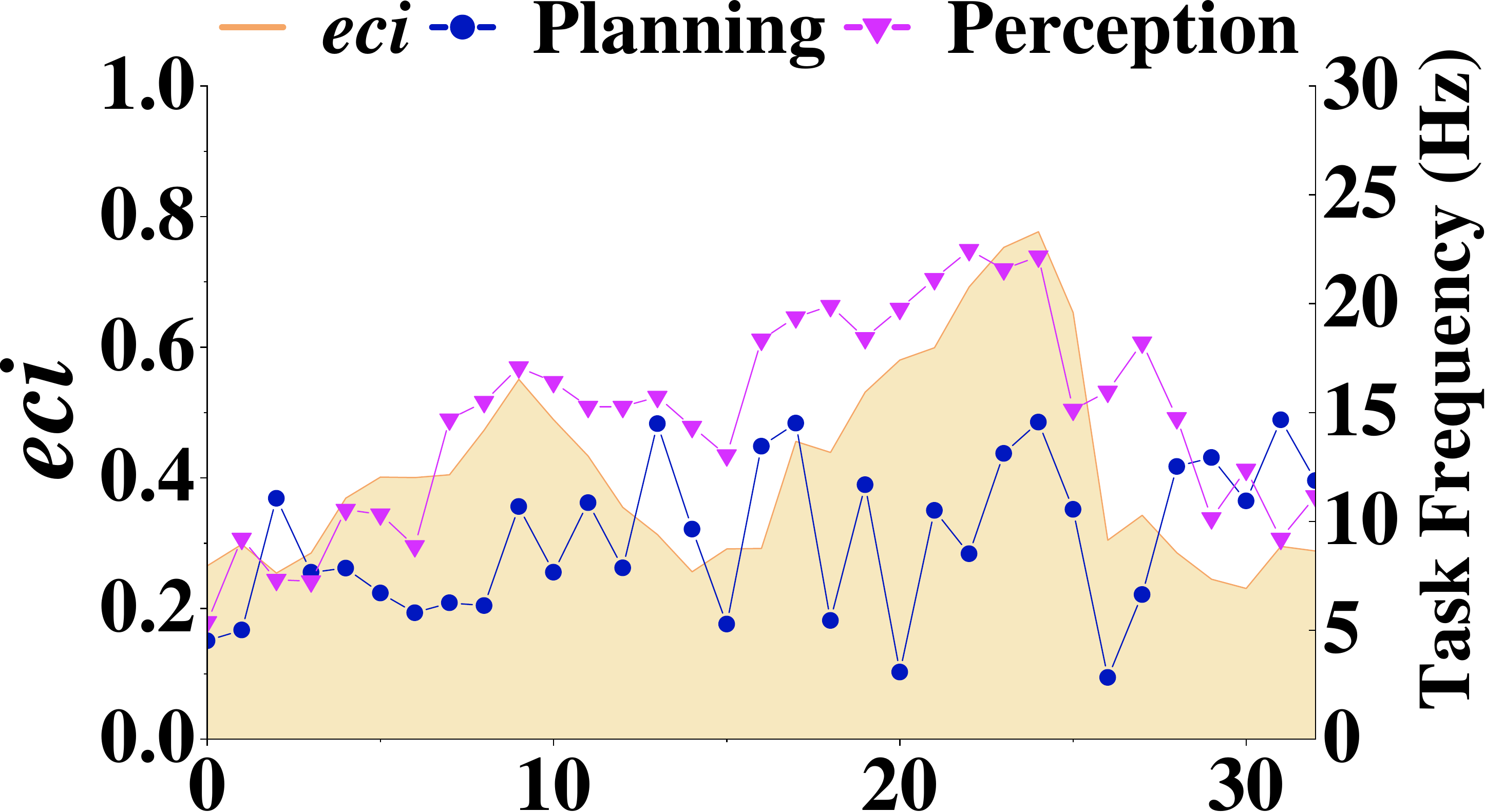}
            \label{pi5-a-s5}
        }
        \hspace{-4mm}
         \subfigure[Resolution variation in S1]{
            \includegraphics[width=.193\linewidth]{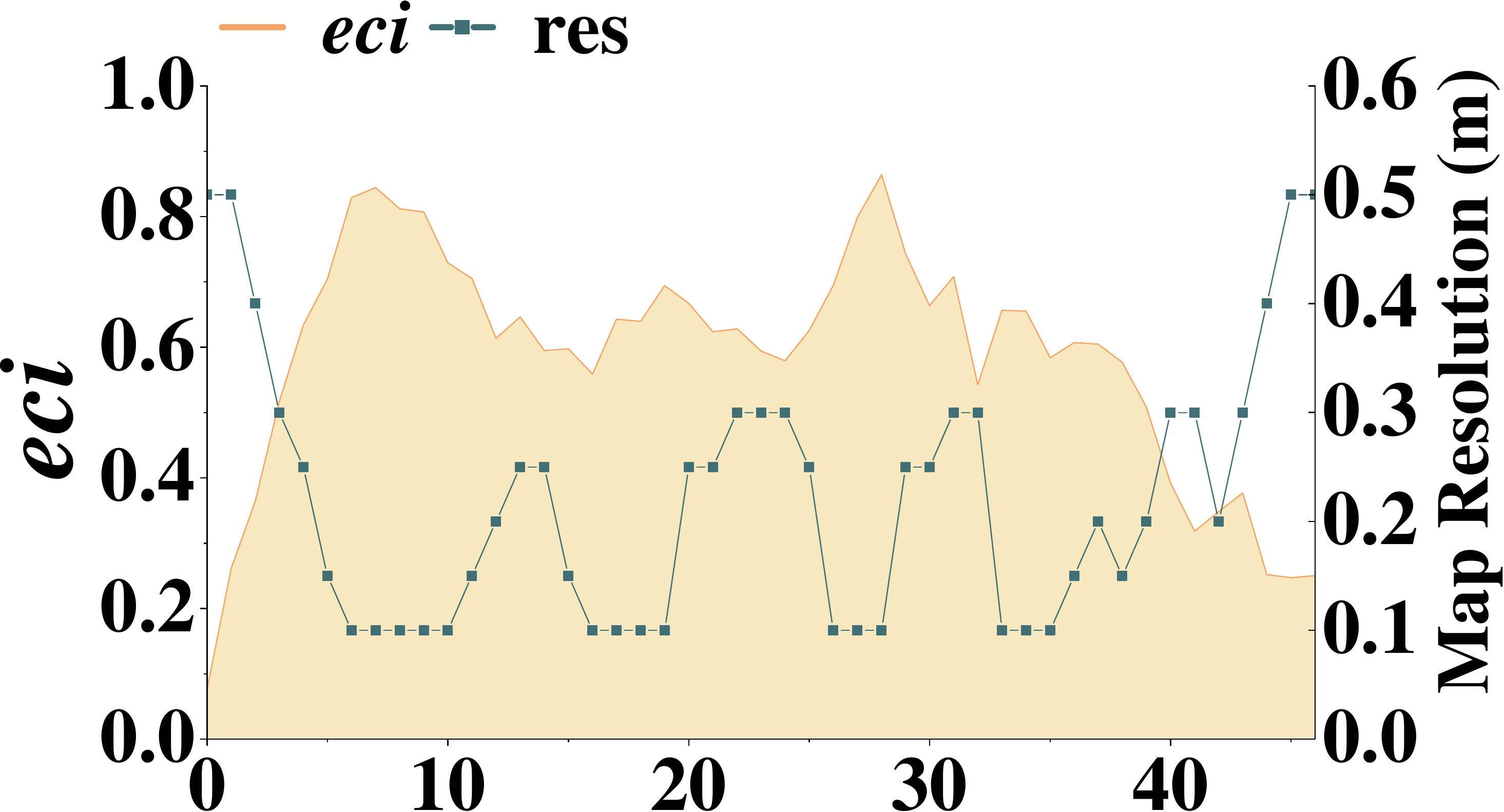}
            \label{pi5-a-res-s1}
        }
        \hspace{-4mm}
        \subfigure[Resolution variation in S2]{
            \includegraphics[width=.193\linewidth]{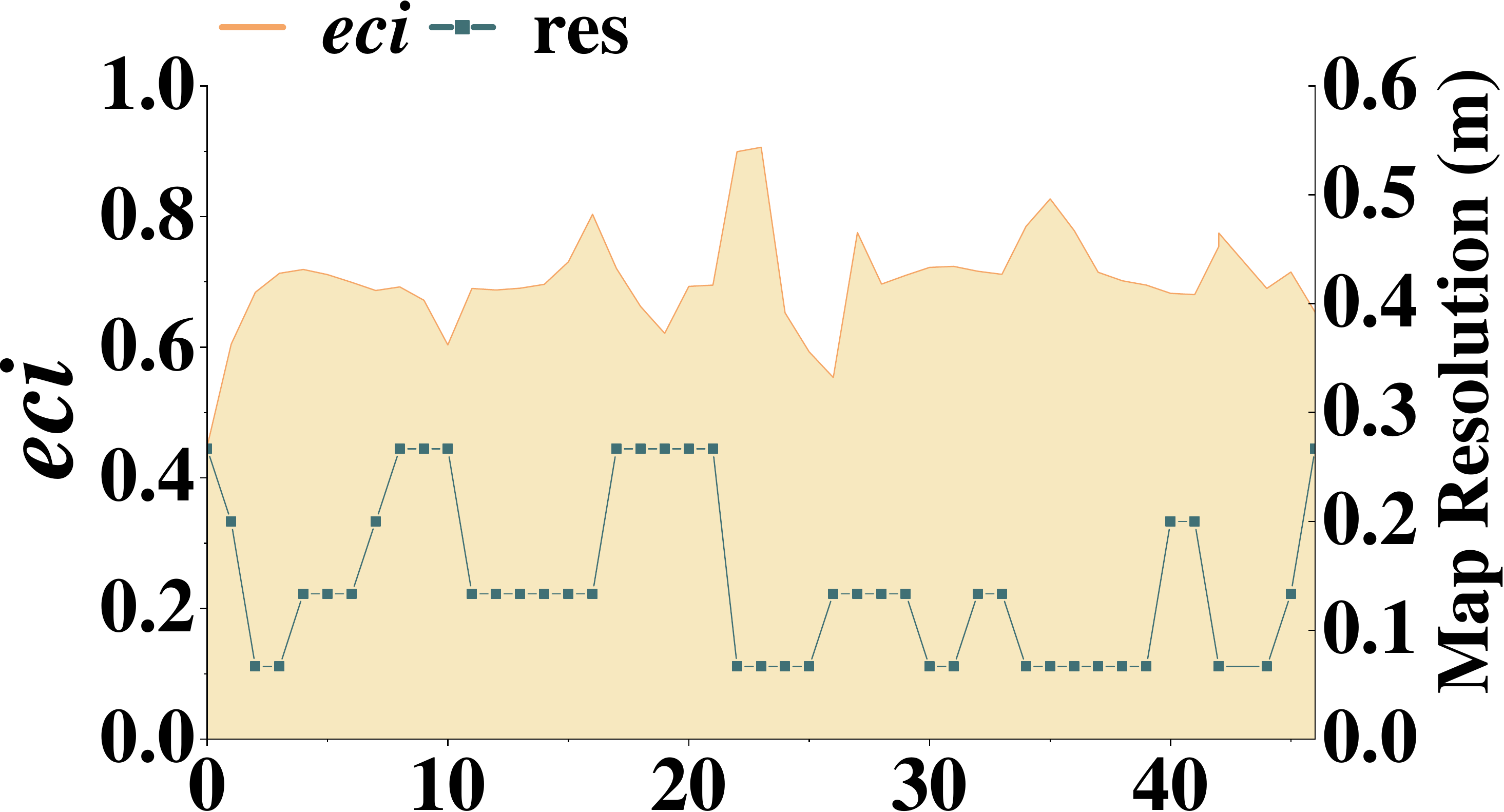}
            \label{pi5-a-res-s2}
        }
        \hspace{-4mm}
        \subfigure[Resolution variation in S3]{
            \includegraphics[width=.193\linewidth]{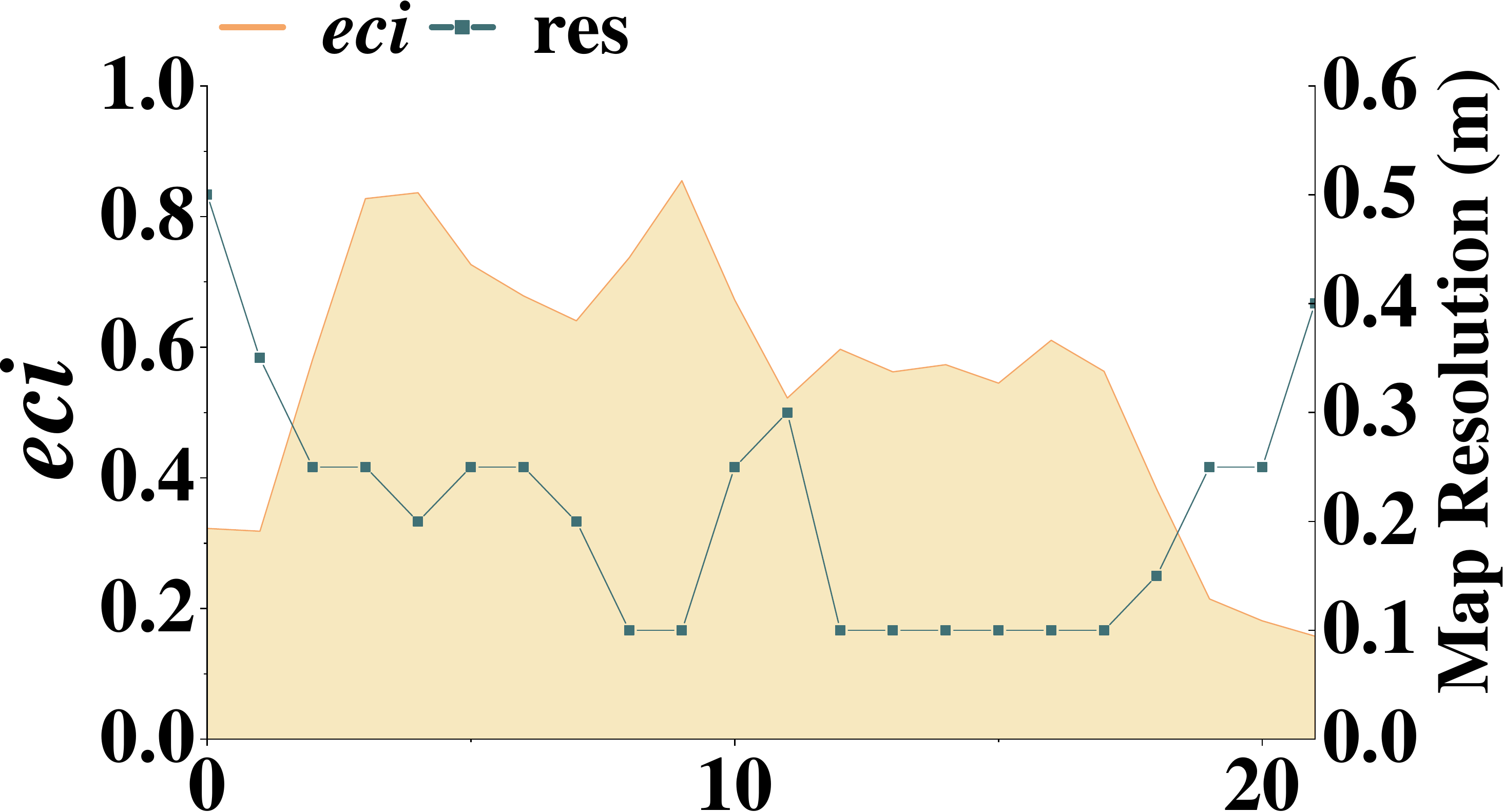}
            \label{pi5-a-res-s3}
        }
        \hspace{-4mm}
        \subfigure[Resolution variation in S4]{
            \includegraphics[width=.193\linewidth]{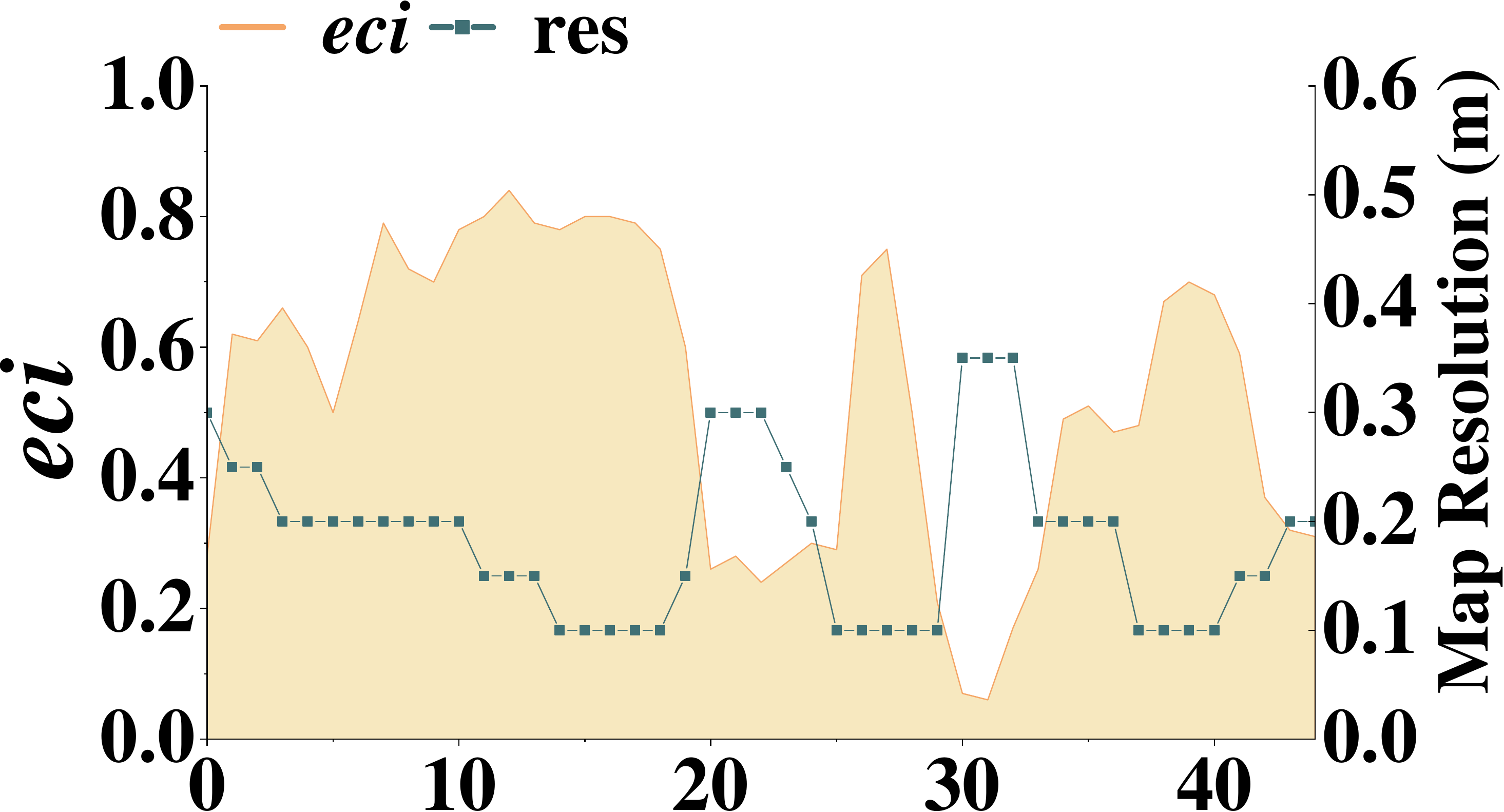}
            \label{pi5-a-res-s4}
        }
        \hspace{-4mm}
        \subfigure[Resolution variation in S5]{
            \includegraphics[width=.193\linewidth]{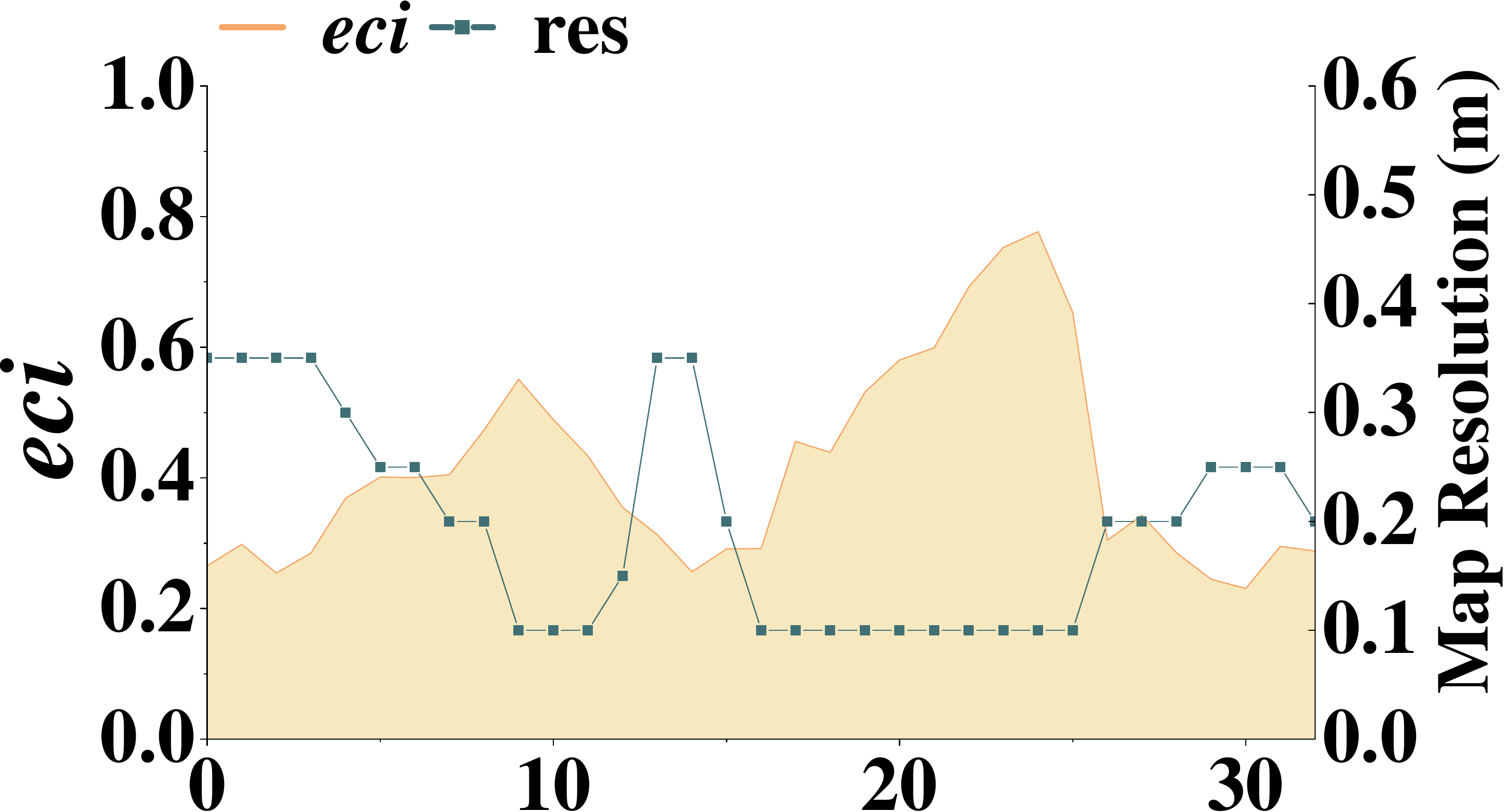}
            \label{pi5-a-res-s5}
        }
    }
    \setlength{\abovecaptionskip}{-2pt}
    \setlength{\belowcaptionskip}{-5pt}
    \caption{$eci$, perception, planning periods, resolution in different scenarios on OrangePi 5.(x-axis: time (s))}
    \vspace{-0.5em}
    \label{exp-hil-freq-eci}
\end{figure*}

The experimental results across the five scenarios and various hardware platforms are summarized in the Tab. \ref{exp-hil}, Fig. \ref{exp-hil-process}, and Fig. \ref{exp-hil-velocity}. 
The $eci$, perception, and planning task periods vary across different scenarios on OrangePi 5 are also displayed in Fig. \ref{exp-hil-freq-eci}. 

\begin{table*}[!htbp]
    \centering
    \caption{Average experimental results on HIL platform.}
    \begin{tabular}{ccccccc|ccccc|ccccccc}
      \cline{1-17}
      \multirow{2}{*}{\textbf{Device}} & \multirow{2}{*}{\textbf{Method}} &\multicolumn{5}{c}{\textbf{CPU Util.(\%)}} & \multicolumn{5}{c}{\textbf{Flight Time(s)}} & \multicolumn{5}{c}{\textbf{Flight Path(m)}}\\
      \cline{3-17}
      & & S1 & S2 & S3 & S4 & S5 & S1 & S2 & S3 & S4 & S5 & S1 & S2 & S3 & S4 & S5\\
      \cline{1-17}
      \multirow{2}{*}{\textbf{Pi 5}} & Ours & 27.7 & 36.5 & 39.0 & 36.9 & 39.7 & 46.0 & 53.3 & 23.3 & 44.1 & 33.9 & 44.8 & 50.0 & 17.6 & 42.1 & 31.8 \\
      & Baseline & 40.3 & 59.2 & 53.6 & 56.9 & 51.3 & 69.6 & 102.3 & 27.0 & 65.3 & 48.6 & 46.6 & 56.6 & 18.1 & 42.4 & 32.7 \\
      \cline{1-17}
      \multirow{2}{*}{\textbf{NX}} & Ours & 25.0 & 29.9 & 29.5 & 47.6 & 28.9 & 46.1 & 53.2 & 17.4 & 44.4 & 32.6 & 43.9 & 50.7 & 17.5 & 42.5 & 31.7 \\
      & Baseline & 36.7 & 64.9 & 45.8 & 59.1 & 44.1 & 89.9 & 128.5 & 27.1 & 122.6 & 52.0 & 56.2 & 69.0 & 20.6 & 76.0 & 34.0 \\
      \cline{1-17}
      \multirow{2}{*}{\textbf{x86}} & Ours & 38.2 & 46.9 & 38.2 & 46.3 & 41.2 & 45.8 & 53.0 & 27.2 & 44.5 & 32.0 & 44.3 & 52.9 & 21.6 & 42.9 & 32.0 \\
      & Baseline & 42.5 & 57.8 & 50.7 & 47.4 & 48.6 & 74.8 & 122.0 & 33.6 & 67.6 & 47.8 & 49.8 & 67.1 & 24.3 & 46.5 & 32.7 \\
      \cline{1-17}
      \multirow{2}{*}{\textbf{TX2}} & Ours & 50.8 & 59.2 & 51.7 & 49.1 & 45.9 & 46.7 & 52.3 & 18.3 & 50.2 & 33.2 & 44.3 & 53.5 & 16.2 & 44.6 & 31.9 \\
      & Baseline & 60.2 & 95.9 & 81.5 & 91.5 & 78.7 & 70.9 & 144.5 & 27.9 & 67.7 & 54.0 & 46.8 & 75.3 & 20.4 & 43.3 & 33.8 \\
      \cline{1-17}
      \multirow{2}{*}{\textbf{Pi 4B}} & Ours & 50.1 & 40.7 & 51.2 & 64.0 & 49.1 & 49.6 & 51.3 & 22.6 & 44.6 & 33.1 & 44.1 & 49.5 & 17.6 & 43.4 & 31.6 \\
      & Baseline & 47.9 & 85.0 & 68.2 & 78.6 & 73.5 & 75.1 & 119.3 & 26.0 & 65.8 & 66.2 & 50.4 & 65.5 & 18.4 & 44.0 & 36.7 \\
      \cline{1-17}
    \end{tabular}
    \label{exp-hil}
\end{table*}

\textbf{{Navigation Task Workload.}} E-Navi {reduces navigation task workload across all devices. On OrangePi 5, E-Navi reduces the average navigation task workload from 52.3\% to 35.9\%, with the largest improvement again observed in Scenario 2 (from 59.2\% to 36.5\%, a 38.3\% reduction). Improvements are also evident on NVIDIA Jetson NX, where E-Navi achieves an average reduction of 37.5\%, with the maximum in Scenario 2 (from 64.9\% to 29.9\%, a 53.9\% reduction). On x86 board, E-Navi reduces workload by 16.3\% on average, highlighting its ability to avoid redundant processing.}

{On the TX2, the average navigation task workload decreases from 81.5\% to 51.3\%, achieving a 37.1\% reduction. The most substantial gain occurs in Scenario 2, where the workload decreases from 95.9\% to 59.2\% (38.3\% reduction). 
On Raspberry Pi 4B, navigation task workload decreases by an average of 27.1\%, with Scenario 2 showing the most notable drop (from 85.0\% to 40.7\%, a 52.1\% reduction), demonstrating the system’s effectiveness in high-complexity conditions.}

\textbf{Flight Time and Path Improvements.} Despite reducing navigation task workload, E-Navi achieves shorter flight paths or comparable flight times by dynamically balancing perception updates and planning triggers, thereby avoiding delayed reactions and redundant re-planning. On the TX2 platform, the average flight time drops from 73.0 s to 40.1 s, achieving a 45.1\% improvement. The most significant gain is observed in Scenario 2, where the flight time is reduced from 144.5 s to 52.3 s, achieving a 63.8\% improvement. On NX, E-Navi reduces the average flight time from 84.0 s to 38.7 s (53.9\% reduction). On Raspberry Pi 4B, the flight time drops by 43.1\% on average. These results confirm that E-Navi’s efficient improves task responsiveness.

In terms of path length, both methods generate similar trajectories under low-complexity conditions. However, in cluttered scenes, E-Navi consistently yields slightly shorter paths. For instance, in Scenario 2 on NX, the path shortens from 69.0 m to 50.7 m. This reflects the tendency of the baseline to overreact or frequently replan when encountering dense obstacles, whereas E-Navi maintains smoother trajectories by proactively adapting task configurations.

\textbf{Velocity Stability Analysis.} Lower velocity standard deviation reflects more stable flight\cite{yu2022doma}. The velocity profiles in Fig.\ref{exp-hil-velocity} further demonstrate that UAVs guided by E-Navi experience smoother and more efficient flight phases while Tab. \ref{exp-hil} confirms consistent gains across scenarios. This stability reflects the planner’s capacity for foresighted and efficient decision-making, effectively avoiding abrupt stops or sharp acceleration changes that are characteristic of the systems. All configurations begin with an acceleration phase within the first 0–5 seconds, followed by a cruising period. However, clear differences are observed in acceleration dynamics, cruising velocity, and overall stability.

Scenario 2 highlights the contrast most distinctly. The baseline shows frequent speed fluctuations after 50 s due to delayed re-planning or emergency avoidance, especially in cluttered areas. These disruptions degrade planning performance. In contrast, E-Navi enables smoother acceleration and stable cruising, with similar trends across other scenarios. This is attributed to E-Navi’s complexity assessment, which supports early risk detection and proactive planning, reducing control oscillations and ensuring smoother trajectory execution.

\textbf{Performance Across Platforms.} Lastly, the performance gap across platforms shows an inverse correlation with hardware capability. On lower-end devices (e.g., Pi 4B), the baseline method struggles under high load, leading to delayed perception and frequent flight interruptions. E-Navi remains robust in these cases, demonstrating its practical value for UAVs on devices with limited resources.

\textbf{Task Adjustment With $eci$.} Fig.~\ref{exp-hil-freq-eci} illustrates that E-Navi generally selects higher perception and planning frequencies when $eci$ values are high, corresponding to complex environments. In open regions with low $eci$ values, the task frequencies are decreased to reduce computational overhead. In transitional areas where $eci$ fluctuates, E-Navi dynamically adjusts the task rates and avoids unnecessary overuse of resources. This behavior is observed across all evaluated scenarios and indicates that the model has learned a resource-aware adaptation strategy. For the map resolution, a consistent trend can be observed: higher $eci$ values lead to finer map resolutions, while lower $eci$ lead to coarser resolutions. Compared with the baseline that always uses fixed task frequencies and a fixed map resolution, E-Navi achieves better resource utilization by dynamically adapting both the task rates and the mapping granularity according to the environmental complexity.

\textbf{Scenario Behavior Analysis.} The experiments underscore E-Navi’s strength in different environments.

Scenario 1 features an open space with minimal obstacles and nearly direct paths. Both E-Navi and the baseline perform similarly, showing that E-Navi introduces no unnecessary overhead in simple environments.

Scenario 2 is the most challenging, with densely and irregularly spaced obstacles, creating narrow passages and frequent dead-ends. The baseline often falls into local minima or triggers repeated re-planning, leading to detours and high CPU usage from frequent collision checks. In contrast, E-Navi’s early complexity assessment identifies difficult regions in advance. Its dynamic task adjustment reduces effort in low-impact directions and focuses on promising paths, enabling smoother decisions. E-Navi avoids over-reacting by leveraging foresight-driven, rather than reactive planning.

Scenario 3 involves narrow corridors and static obstacles, requiring precise control over global flexibility. While both methods follow similar paths, E-Navi maintains better velocity stability through efficient local planning.

Scenarios 4 and 5 present mixed-complexity environments with transitions between open and cluttered areas. E-Navi adjusts configurations according to local $eci$ values. The baseline lacks this flexibility, applying uniform computation, which impairs performance during complexity transitions.

\subsubsection{Timing costs of core parts in E-Navi}
\begin{figure}[!htbp]
\vspace{-4pt}
    \centering{
        \subfigure[Timing cost of computing $eci$.]{
            \includegraphics[width=.9\linewidth]{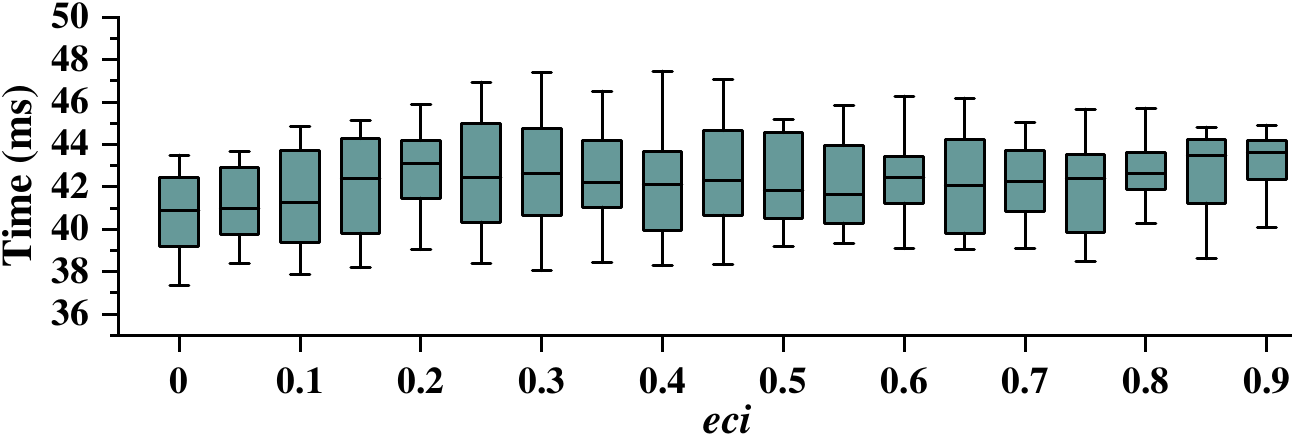}
            \label{eci_time}
        }
        \subfigure[Timing costs of adapters.]{
            \includegraphics[width=.9\linewidth]{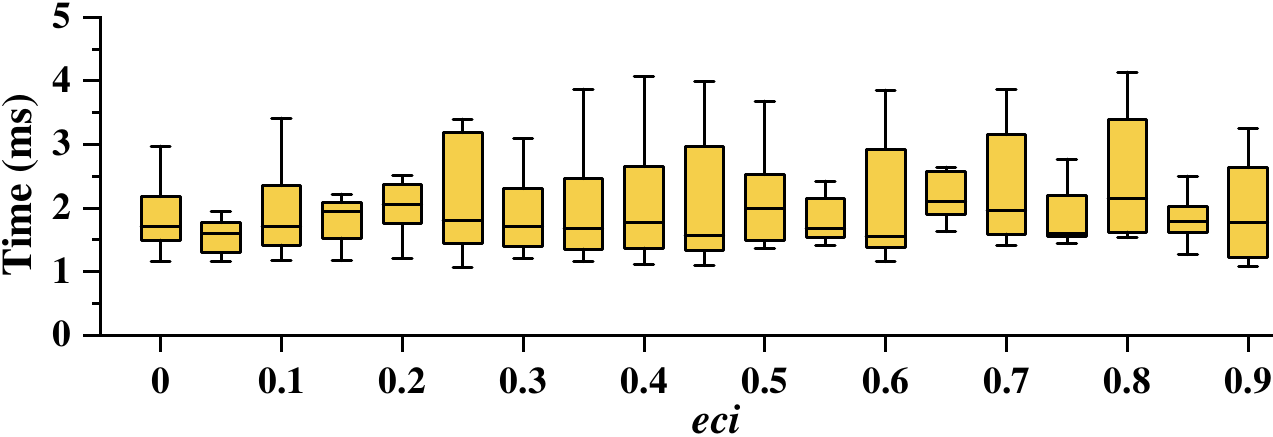}
            \label{adapter_time}
        }
    }
    \setlength{\belowcaptionskip}{-4pt}
    \caption{Timing costs on varying $eci$ scenarios.}
    \label{eci_timingcosts}
\end{figure}

In this section, we present the timing costs of the core parts of E-Navi under varying $eci$ ranging from 0 to 0.9. The execution times of $eci$ and the adapter are recorded in multiple trials on TX2 to evaluate the efficiency and stability of the module in Fig. \ref{eci_timingcosts}. The ECI is computed at a fixed frequency of 30Hz, which is consistent with the output frequency of the sensor.

\textbf{Timing Costs of $eci$ Computation.} As displayed in Fig. \ref{eci_time}, the $eci$ calculation times fluctuate between approximately 37 and 49 ms, with 41 ms on average on TX2. There is a slight upward trend in computation time as the $eci$ value increases, especially noticeable beyond 0.5. Specifically, the average times for lower $eci$ values (0 to 0.4) predominantly range around 38-42 ms, whereas higher $eci$ values (0.6 to 0.9) show increased timing spikes that reach up to 49 ms. The variance within each $eci$ interval remains moderate, indicating consistent performance. In addition, the average $eci$ computation time is 59 ms, 31 ms, and 29 ms on RaspberryPi, OrangePi, and x86 board, respectively, which is acceptable for UAVs \cite{8804776,tordesillas2019real}.

From the results, the E-Navi indeed brings a non-trivial cost for computing ECI. However, we note that implementation efforts can be conducted to further reduce this cost, e.g., using Voxel Grid downsampling and Octree indexing with a
pre-processing step to filter and normalize the sensor data. As shown by the results, the E-Navi remains effective by realizing adaptive navigation workload based on the ECI value, which achieves a more stable flight with a shorter path and time. In addition, both the HIL simulation and the real-world case study (see Sec. VI.C) justify the feasibility of deploying E-Navi on resource-constrained UAV platforms.


\textbf{Timing Performance of Adapters Computation.} As shown in Fig. \ref{adapter_time}, the E-Navi adapters demonstrate consistently low latency. The average time is approximately 2.08 ms, with 90\% of the samples below 3 ms. This is attributed to the lightweight structure of the adapters, which relies on precomputed constraints and a fast inferring mechanism rather than full optimization.
These results demonstrate that the output of the adapters introduces negligible latency compared to the navigation system\cite{8804776}. 

In addition, the deadline miss ratio of tasks is evaluated under all scenarios. Among the scenarios, only Scenario 2 exhibits deadline misses with a ratio of 0.34\%. This indicates E-Navi maintains stable and predictable task scheduling under varying environmental conditions, justifying the practicality of our adaptation mechanism for real-world deployment.

\subsection{Real-world Case Study}
\begin{figure}[!htbp]
    \centering
    \subfigure[Navigation workload comparison in case study.]{
        \includegraphics[width=\linewidth]{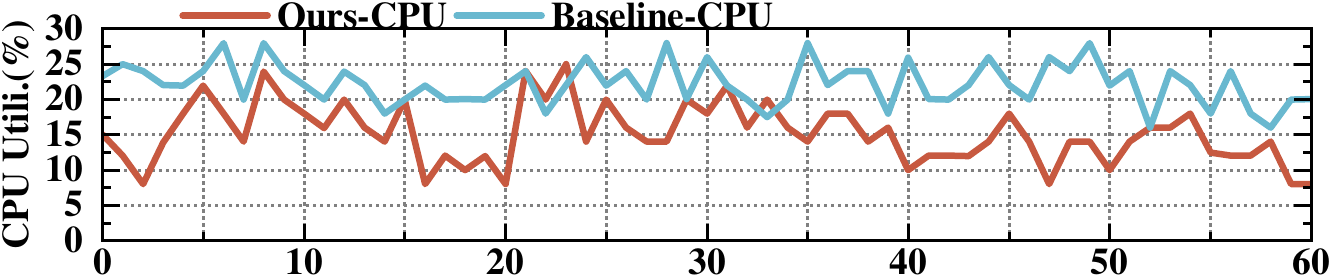}
        \label{rw-c}
    }
    \subfigure[Velocity comparison in case study.]{
        \includegraphics[width=\linewidth]{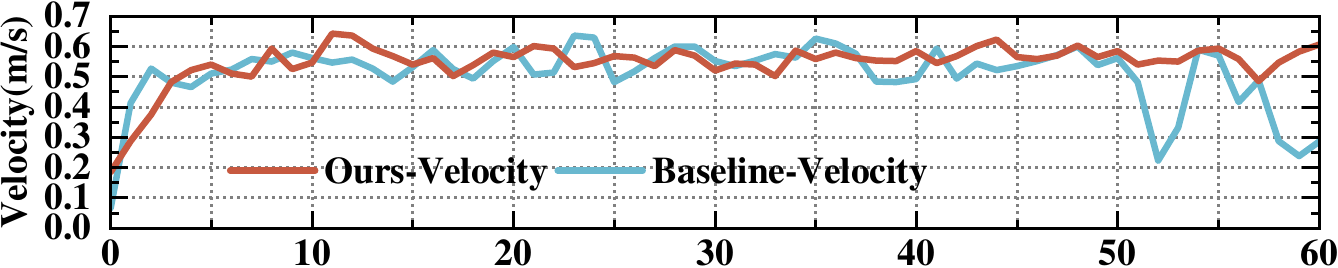}
        \label{rw-v}
    }
    \caption{Navigation workload and velocity. (x-axis: time (s))}
    
    \label{rw}
\end{figure}

The results of real-world case study in outdoor environments over a 60 s flight window are displayed in Fig. \ref{rw}.

\textbf{Navigation Workload.} As shown in Fig. \ref{rw-c}, E-Navi maintains lower utilization throughout the flight. The average navigation task workload in E-Navi is 15.28\%, compared to 22.64\% for the baseline, achieving a relative reduction of approximately 32.5\%. Additionally, peak utilization is reduced from 27.98\% to 23.96\%, indicating more efficient computing abilities. These results demonstrate that the dynamic configuration effectively decreases redundant computation, which is particularly important for platforms with limited resources.

\textbf{Velocity Stability.} Fig. \ref{rw-v} shows that E-Navi also achieves more stable and efficient flight behavior. The UAV maintains a more consistent velocity profile with fewer abrupt drops or spikes. The standard deviation of the velocity using E-Navi is 0.062 m/s, which is significantly lower than the baseline of 0.114 m/s. This demonstrates the E-Navi avoids conservative slowdowns in low-risk regions by dynamically adjusting the navigation task configurations according to the environmental complexity suggested by the ECI value.

Overall, the results demonstrate that E-Navi enables a tight coordination between perceptual workload and motion planning, leading to both reduced computational burden and improved flight smoothness. This validates its practical value in enabling efficient autonomy for UAV navigation systems.

\vspace{-0.5em}

\vspace{-1em}
\subsection{{Discussion}}
Based on the results, the E-Navi improves the UAV performance in the following aspects. 

\textbf{Resource Adaption.} E-Navi dynamically adjusts the frequencies of perception and planning as well as the map resolution in response to environmental complexity. Unlike static systems that operate at fixed high-frequency and high-resolution settings, {E-Navi effectively avoids the redundant re-planning in simple environments while being responsive to sudden environmental changes based on the ECI. This adaptive mechanism reduces unnecessary computation and ensures that resources are used more effectively across diverse scenarios.}

\textbf{Flight Performance.} Flight performance is significantly improved by developing more efficient and adaptive trajectories in complex scenarios. This improvement is achieved through the dynamic adjustment of perception and planning, which is determined by monitoring the surrounding environments. In simple scenarios, reducing redundant re-planning avoids frequent path changes and unnecessary accelerations or decelerations, leading to smoother and faster flights.

\textbf{Flight Stability.} Flight stability is enhanced by reducing sudden accelerations and emergency braking events. These instabilities typically occur when UAVs encounter abrupt environmental changes, such as sharp turns or newly appearing obstacles, without adequate time for reaction. In E-Navi, the UAV monitors the changes in $eci$, and proactively adjusts perception and planning configurations, thereby maintaining stable and responsive flight behaviors. 

%% file: tex/conclusion.tex
This paper presents E-Navi, an environmental-adaptive navigation system designed to enhance UAV performance in dynamic environments on resource-constrained platforms. Extensive HIL and real-world experiments across various platforms demonstrate that the E-Navi significantly outperforms the baseline method with up to {53.9\%} navigation task workload reduction, up to {63.8\%} flight time savings, and more efficient and smoother UAV operations in complex environments.

%% file: ref.bib
@inproceedings{shao2024design,
  title={Design and evaluation of motion planners for quadrotors in environments with varying complexities},
  author={Shao, Yifei Simon and Wu, Yuwei and Jarin-Lipschitz, Laura and Chaudhari, Pratik and Kumar, Vijay},
  booktitle={2024 IEEE International Conference on Robotics and Automation (ICRA)},
  pages={10033--10039},
  year={2024},
  organization={IEEE}
}

@inproceedings{boroujerdian2021roborun,
    title={Roborun: A robot runtime to exploit spatial heterogeneity},
    author={Boroujerdian, Behzad and Ghosal, Radhika and Cruz, Jonathan and Plancher, Brian and Reddi, Vijay Janapa},
    booktitle={Design Automation Conference (DAC)},
    pages={829--834},
    year={2021}
}

@inproceedings{boroujerdian2018mavbench,
    title={Mavbench: Micro aerial vehicle benchmarking},
    author={Boroujerdian, Behzad and Genc, Hasan and Krishnan, Srivatsan and Cui, Wenzhi and Faust, Aleksandra and Reddi, Vijay},
    booktitle={International symposium on microarchitecture (MICRO)},
    pages={894--907},
    year={2018}
}

@inproceedings{hsiao2022zhuyi,
    title={Zhuyi: perception processing rate estimation for safety in autonomous vehicles},
    author={Hsiao, Yu-Shun and Hari, Siva Kumar Sastry and Filipiuk, Micha{\l} and Tsai, Timothy and Sullivan, Michael B and Reddi, Vijay Janapa and Singh, Vasu and Keckler, Stephen W},
    booktitle={Design Automation Conference (DAC)},
    pages={289--294},
    year={2022}
}

@article{hsiao2023silent,
    title={Silent Data Corruption in Robot Operating System: A Case for End-to-End System-Level Fault Analysis Using Autonomous UAVs},
    author={Hsiao, Yu-Shun and Wan, Zishen and Jia, Tianyu and Ghosal, Radhika and Mahmoud, Abdulrahman and Raychowdhury, Arijit and Brooks, David and Wei, Gu-Yeon and Reddi, Vijay Janapa},
    journal={IEEE Transactions on Computer-Aided Design of Integrated Circuits and Systems},
    year={2023},
    publisher={IEEE}
}

@inproceedings{hadidi2021quantifying,
    title={Quantifying the design-space tradeoffs in autonomous drones},
    author={Hadidi, Ramyad and Asgari, Bahar and Jijina, Sam and Amyette, Adriana and Shoghi, Nima and Kim, Hyesoon},
    booktitle={International Conference on Architectural Support for Programming Languages and Operating Systems (ASPLOS)},
    pages={661--673},
    year={2021}
}

@inproceedings{hadidi2023context,
    title={Context-Aware Task Handling in Resource-Constrained Robots with Virtualization},
    author={Hadidi, Ramyad and Ghaleshahi, Nima Shoghi and Asgari, Bahar and Kim, Hyesoon},
    booktitle={International Conference on Edge Computing and Communications (EDGE)},
    pages={255--261},
    year={2023}
}

@article{zhou2020ego,
    title={Ego-planner: An esdf-free gradient-based local planner for quadrotors},
    author={Zhou, Xin and Wang, Zhepei and Ye, Hongkai and Xu, Chao and Gao, Fei},
    journal={IEEE Robotics and Automation Letters},
    volume={6},
    number={2},
    pages={478--485},
    year={2020},
    publisher={IEEE}
}

@article{wang2022geometrically,
    title={Geometrically constrained trajectory optimization for multicopters},
    author={Wang, Zhepei and Zhou, Xin and Xu, Chao and Gao, Fei},
    journal={IEEE Transactions on Robotics},
    volume={38},
    number={5},
    pages={3259--3278},
    year={2022},
    publisher={IEEE}
}

@inproceedings{chen2020computationally,
    title={Computationally efficient obstacle avoidance trajectory planner for uavs based on heuristic angular search method},
    author={Chen, Han and Lu, Peng},
    booktitle={International Conference on Intelligent Robots and Systems (IROS)},
    pages={5693--5699},
    year={2020}
}

@inproceedings{quan2021eva,
    title={EVA-planner: Environmental adaptive quadrotor planning},
    author={Quan, Lun and Zhang, Zhiwei and Zhong, Xingguang and Xu, Chao and Gao, Fei},
    booktitle={International Conference on Robotics and Automation (ICRA)},
    pages={398--404},
    year={2021}
}

@article{wang2022neither,
    title={Neither fast nor slow: How to fly through narrow tunnels},
    author={Wang, Luqi and Xu, Hao and Zhang, Yichen and Shen, Shaojie},
    journal={IEEE Robotics and Automation Letters},
    volume={7},
    number={2},
    pages={5489--5496},
    year={2022},
    publisher={IEEE}
}

@inproceedings{zhang2023asap,
    title={ASAP: A Spatial-Aware Perception Computing Method for Energy Reduction of Micro Aerial Vehicles A Spatial-Aware Perception Computing Method for MAV},
    author={Zhang, Yibo and Zhang, Yuanhai and Li, Boyang and Huang, Kai},
    booktitle={International Conference on Artificial Intelligence and Computer Engineering (ICAICE)},
    pages={143--149},
    year={2023}
}

@inproceedings{shah2023energy,
    title={Energy-efficient realtime motion planning},
    author={Shah, Deval and Yang, Ningfeng and Aamodt, Tor M},
    booktitle={International Symposium on Computer Architecture (ISCA)},
    pages={1--17},
    year={2023}
}

@article{liu2021pi,
    title={$\pi$-rt: A runtime framework to enable energy-efficient real-time robotic vision applications on heterogeneous architectures},
    author={Liu, Liu and Tang, Jie and Liu, Shaoshan and Yu, Bo and Xie, Yuan and Gaudiot, Jean-Luc},
    journal={Computer},
    volume={54},
    number={4},
    pages={14--25},
    year={2021},
    publisher={IEEE}
}

@ARTICLE{8951121,
    author={Aiello, Giuseppe and Hopps, Fabrizio and Santisi, Domenico and Venticinque, Mario},
    journal={IEEE Transactions on Engineering Management}, 
    title={The Employment of Unmanned Aerial Vehicles for Analyzing and Mitigating Disaster Risks in Industrial Sites}, 
    year={2020},
    volume={67},
    number={3},
    pages={519-530}
}

@article{alsamhi2022uav,
    title={UAV computing-assisted search and rescue mission framework for disaster and harsh environment mitigation},
    author={Alsamhi, Saeed Hamood and Shvetsov, Alexey V and Kumar, Santosh and Shvetsova, Svetlana V and Alhartomi, Mohammed A and Hawbani, Ammar and Rajput, Navin Singh and Srivastava, Sumit and Saif, Abdu and Nyangaresi, Vincent Omollo},
    journal={Drones},
    volume={6},
    number={7},
    pages={154},
    year={2022},
    publisher={MDPI}
}

@article{hazmy2023potential,
    title={Potential of Satellite-Airborne Sensing Technologies for Agriculture 4.0 and Climate-Resilient: A Review},
    author={Hazmy, Asa Ibnu and Hawbani, Ammar and Wang, Xingfu and Al-Dubai, Ahmed and Ghannami, Aiman and Yahya, Ali Abdullah and Zhao, Liang and Alsamhi, Saeed Hamood},
    journal={IEEE Sensors Journal},
    year={2023},
    publisher={IEEE}
}

@article{liu2021real,
    title={Real-time task scheduling for machine perception in intelligent cyber-physical systems},
    author={Liu, Shengzhong and Yao, Shuochao and Fu, Xinzhe and Shao, Huajie and Tabish, Rohan and Yu, Simon and Bansal, Ayoosh and Yun, Heechul and Sha, Lui and Abdelzaher, Tarek},
    journal={IEEE Transactions on Computers},
    volume={71},
    number={8},
    pages={1770--1783},
    year={2021},
    publisher={IEEE}
}

@article{son2023gradient,
    title={Gradient informed proximal policy optimization},
    author={Son, Sanghyun and Zheng, Laura and Sullivan, Ryan and Qiao, Yi-Ling and Lin, Ming},
    journal={Advances in Neural Information Processing Systems},
    volume={36},
    pages={8788--8814},
    year={2023}
}

@article{faessler2016autonomous,
    title={Autonomous, vision-based flight and live dense 3D mapping with a quadrotor micro aerial vehicle},
    author={Faessler, Matthias and Achtelik, Markus W and Kneip, Laurent and Scaramuzza, Davide},
    journal={Journal of Field Robotics},
    volume={33},
    number={4},
    pages={431--450},
    year={2016},
    publisher={Wiley Online Library}
}

@article{loquercio2021learning,
    title={Learning high-speed flight in the wild},
    author={Loquercio, Antonio and Kaufmann, Elia and Ranftl, Rene and Dosovitskiy, Alexey and Koltun, Vladlen and Scaramuzza, Davide},
    journal={Science Robotics},
    volume={6},
    number={59},
    pages={eabg5810},
    year={2021},
    publisher={American Association for the Advancement of Science}
}

@misc{nvidiaJetson,
    title = {NVIDIA Jetson Platform},
    url = {https://developer.nvidia.com/embedded-computing}
}

@misc{raspberryPi,
    title = {Raspberry Pi Foundation},
    url = {https://www.raspberrypi.org/}
}

@misc{orangepi5,
    title = {Orange Pi 5 Specifications},
    url = {http://www.orangepi.org/OrangePi5/}
}

@misc{x86,
    title = {X86 Products},
    url = {http://www.intel.com/}
}

@inproceedings{li2023red,
    title={Red: A systematic real-time scheduling approach for robotic environmental dynamics},
    author={Li, Zexin and Ren, Tao and He, Xiaoxi and Liu, Cong},
    booktitle={2023 IEEE Real-Time Systems Symposium (RTSS)},
    pages={210--223},
    year={2023},
    organization={IEEE}
}

@inproceedings{mcgowen2024scheduling,
    title={Scheduling for Cyber-Physical Systems with Heterogeneous Processing Units under Real-World Constraints},
    author={McGowen, Justin and Dagli, Ismet and Dantam, Neil T and Belviranli, Mehmet E},
    booktitle={Proceedings of the 38th ACM International Conference on Supercomputing},
    pages={298--311},
    year={2024}
}

@inproceedings{xu2024automatic,
    title={Automatic Hardware/Software Design for High-Speed Autonomous Unmanned Aerial Vehicles Guided by a Flight Model},
    author={Xu, Yuanfan and Yu, Jincheng and Zhang, Suquan and Xiang, Yunfei and Jia, Hongyang and Wang, Yu},
    booktitle={Proceedings of the 61st ACM/IEEE Design Automation Conference},
    pages={1--4},
    year={2024}
}

@article{wu2025towards,
    title={Towards Optimizing a Convex Cover of Collision-Free Space for Trajectory Generation},
    author={Wu, Yuwei and Spasojevic, Igor and Chaudhari, Pratik and Kumar, Vijay},
    journal={IEEE Robotics and Automation Letters},
    year={2025},
    publisher={IEEE}
}

@article{huang2024safe,
    title={Safe Interval Motion Planning for Quadrotors in Dynamic Environments},
    author={Huang, Songhao and Wu, Yuwei and Tao, Yuezhan and Kumar, Vijay},
    journal={arXiv preprint arXiv:2409.10647},
    year={2024}
}

@ARTICLE{9520259,
    author={Liu, Shengzhong and Yao, Shuochao and Fu, Xinzhe and Shao, Huajie and Tabish, Rohan and Yu, Simon and Bansal, Ayoosh and Yun, Heechul and Sha, Lui and Abdelzaher, Tarek},
    journal={IEEE Transactions on Computers}, 
    title={Real-Time Task Scheduling for Machine Perception in Intelligent Cyber-Physical Systems}, 
    year={2022},
    volume={71},
    number={8},
    pages={1770-1783}
}

@INPROCEEDINGS{9355507,
    author={Liu, Shengzhong and Yao, Shuochao and Fu, Xinzhe and Tabish, Rohan and Yu, Simon and Bansal, Ayoosh and Yun, Heechul and Sha, Lui and Abdelzaher, Tarek},
    booktitle={2020 IEEE Real-Time Systems Symposium (RTSS)}, 
    title={On Removing Algorithmic Priority Inversion from Mission-critical Machine Inference Pipelines}, 
    year={2020},
    pages={319-332}
}

@article{burri2016euroc,
    title={The EuRoC micro aerial vehicle datasets},
    author={Burri, Michael and Nikolic, Janosch and Gohl, Pascal and Schneider, Thomas and Rehder, Joern and Omari, Sammy and Achtelik, Markus W and Siegwart, Roland},
    journal={The International Journal of Robotics Research},
    volume={35},
    number={10},
    pages={1157--1163},
    year={2016}
}

@INPROCEEDINGS{8804776,
    author={Palossi, Daniele and Conti, Francesco and Benini, Luca},
    booktitle={2019 15th International Conference on Distributed Computing in Sensor Systems (DCOSS)}, 
    title={An Open Source and Open Hardware Deep Learning-Powered Visual Navigation Engine for Autonomous Nano-UAVs}, 
    year={2019},
    volume={},
    number={},
    pages={604-611}
}

@article{lopez2004utilization,
    title={Utilization bounds for EDF scheduling on real-time multiprocessor systems},
    author={L{\'o}pez, Jos{\'e} Mar{\'\i}a and D{\'\i}az, Jos{\'e} Luis and Garc{\'\i}a, Daniel F},
    journal={Real-Time Systems},
    volume={28},
    pages={39--68},
    year={2004},
    publisher={Springer}
}

@article{liu1973scheduling,
    title={Scheduling algorithms for multiprogramming in a hard-real-time environment},
    author={Liu, Chung Laung and Layland, James W},
    journal={Journal of the ACM (JACM)},
    volume={20},
    number={1},
    pages={46--61},
    year={1973},
    publisher={ACM New York, NY, USA}
}

@inproceedings{tordesillas2019real,
    title={Real-time planning with multi-fidelity models for agile flights in unknown environments},
    author={Tordesillas, Jesus and Lopez, Brett T and Carter, John and Ware, John and How, Jonathan P},
    booktitle={International Conference on Robotics and Automation (ICRA)},
    pages={725--731},
    year={2019}
}

@inproceedings{yu2022doma,
    title={Doma: Deep smooth trajectory generation learning for real-time uav motion planning},
    author={Yu, Jin and Piao, Haiyin and Hou, Yaqing and Mo, Li and Yang, Xin and Zhou, Deyun},
    booktitle={Proceedings of the International Conference on Automated Planning and Scheduling},
    volume={32},
    pages={662--666},
    year={2022}
}
